\documentclass[10pt,twocolumn,letterpaper]{article}

\usepackage[pagenumbers]{cvpr} %
\usepackage[dvipsnames]{xcolor}

\usepackage{times}
\usepackage{epsfig}
\usepackage{graphicx}
\usepackage{amsmath}
\usepackage{amssymb}
\usepackage{float}
\usepackage{booktabs}
\usepackage{multirow}
\usepackage{comment}

\newcommand{\myparagraph}[1]{\noindent\textbf{#1}~~}

\newcommand{\ci}{c_I}

\newcommand{\ce}{c_{E}}
\newcommand{\tc}{c}
\newcommand{\gx}{\pmb{x}}

\newcommand{\tgx}{\Tilde{\pmb{x}}}

\newcommand{\E}{\mathbb{E}}

\usepackage{caption}

\definecolor{cvprblue}{rgb}{0.21,0.49,0.74}
\usepackage[pagebackref,breaklinks,colorlinks,citecolor=cvprblue]{hyperref}

\usepackage[capitalize]{cleveref}

\def\paperID{12986} %
\def\confName{CVPR}
\def\confYear{2024}

\makeatletter
\def\blfootnote{\gdef\@thefnmark{}\@footnotetext}
\makeatother

\begin{document}

\title{HIVE: Harnessing Human Feedback for Instructional Visual Editing}

\author
{
Shu Zhang*$^{1}$,
Xinyi Yang*$^1$,
Yihao Feng*$^1$,
Can Qin$^1$,
Chia-Chih Chen$^1$,
Ning Yu$^1$,
Zeyuan Chen$^1$,\\
Huan Wang$^1$,
Silvio Savarese$^{1,2}$,
Stefano Ermon$^2$,
Caiming Xiong$^1$,
Ran Xu$^1$ \\
$^1$Salesforce AI Research, $^2$Stanford University\\
}

\twocolumn[{%
\renewcommand\twocolumn[1][]{#1}%
\maketitle

\vspace{-7.5mm}
\begin{center}
    \includegraphics[width=\linewidth]{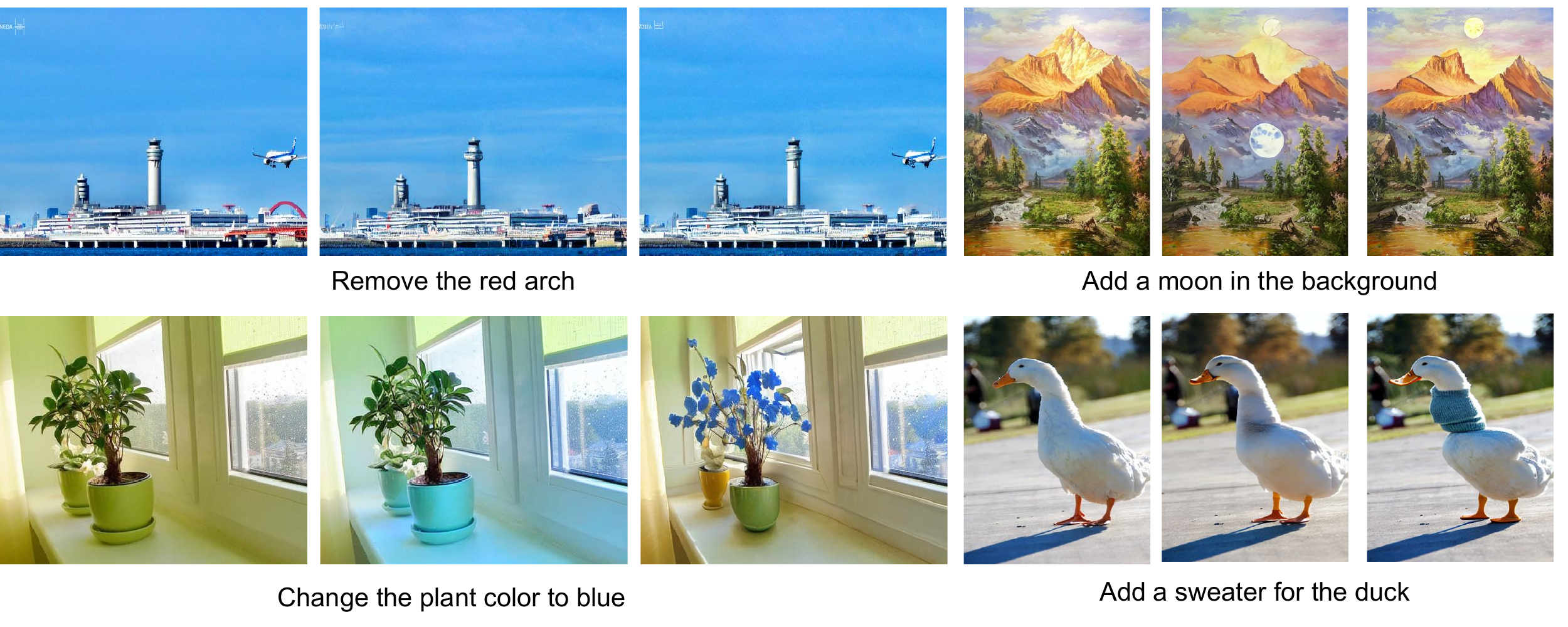}
    \vspace{-2.4em}
    \captionof{figure}{
    We show four groups of representative results. In each triplet, from left to right are: the original image,  InstructPix2Pix~\cite{brooks2022instructpix2pix} using our data (IP2P-Ours), and HIVE. We observe that HIVE leads to more acceptable results than the model without human feedback. For instance, in the left two examples, IP2P-Ours understands the editing instruction ``remove'' and ``change to blue'' individually, but fails to understand the corresponding objects.  Human feedback resolves this ambiguity, as shown in  other examples as well.}
    \label{fig:representative_results}
    \vspace*{-0.2mm}
\end{center}%
}]

\maketitle

\begin{abstract}
\vspace*{-2.9mm}
Incorporating human feedback has been shown to be crucial to align text generated by large language models to human preferences.
We hypothesize that state-of-the-art instructional image editing models, where outputs are generated based on an input image and an editing instruction, could similarly benefit from human feedback, as their 
outputs may not adhere to the correct instructions and preferences of users.
In this paper, we present a novel framework to harness human feedback for instructional visual editing (HIVE). Specifically, we collect human feedback on the edited images and learn a reward function to capture the underlying user preferences. We then introduce scalable diffusion model fine-tuning methods that can incorporate human preferences based on the estimated reward.  

Besides, to mitigate the bias brought by the limitation of data,
we contribute a new 1.1M \textbf{training dataset},  a 3.6K \textbf{reward dataset} for rewards learning, and a 1K \textbf{evaluation dataset} to boost the performance of instructional image editing. 
We conduct extensive empirical experiments quantitatively and qualitatively, showing that HIVE is favored over previous state-of-the-art instructional image editing approaches by a large margin.
\vspace*{-0.8mm}
\blfootnote{\hspace{-2em}*Denotes equal contribution. Primary contact: shu.zhang@salesforce.com. \\Our project page: https://shugerdou.github.io/hive/.}

\end{abstract}

\section{Introduction}

\begin{figure*}
    \includegraphics[width=\linewidth]{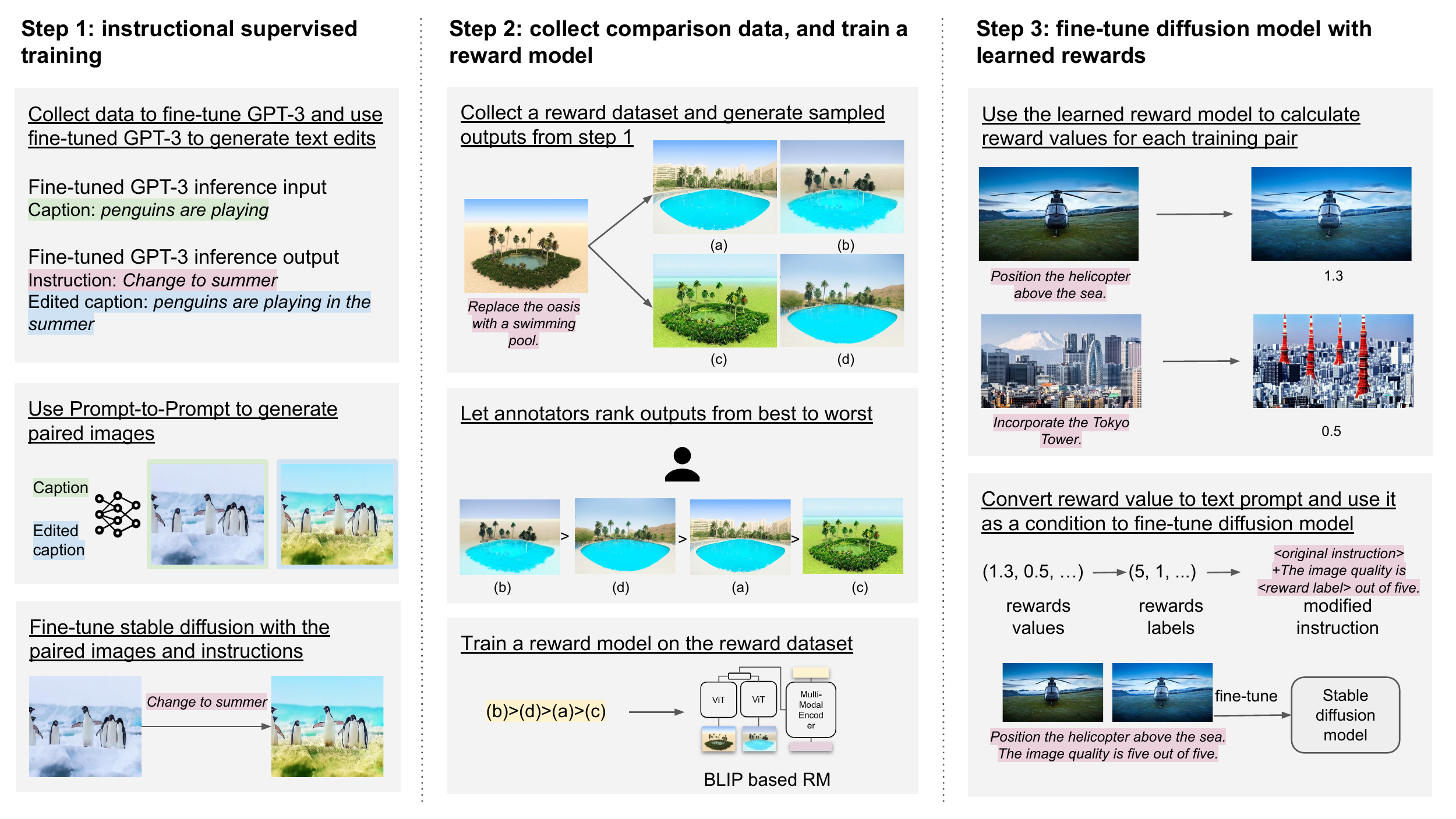}
    \vspace{-2.5em}
    \caption{Overall architecture of HIVE. The first step is to train a baseline HIVE without human feedback. In the second step, we collect human feedback to rank variant outputs for each image-instruction pair, and train a reward model to learn the rewards. In the third step, we fine-tune diffusion models by integrating the estimated rewards.}
    \label{fig:method}
    \vspace*{-3.9mm}
\end{figure*}

State-of-the-art (SOTA) text-to-image generative models have shown impressive performance in terms of both image quality and alignment between 
output images and captions
\cite{alayrac2022flamingo,rombach2022high,ramesh2022hierarchical}. Thanks to the impressive generation abilities of these models, \textit{instructional image editing} has emerged as one of the most promising application scenarios for content generation \cite{brooks2022instructpix2pix}. 
Different from traditional image editing \cite{avrahami2022blended,hertz2022prompt,wallace2022edict,liew2022magicmix,hertz2022prompt,wallace2022edict},
where both the input and the edited caption are needed, \textit{instructional image editing} only requires human-readable instructions. 
For instance, classic image editing approaches require an input caption ``a dog is playing a ball'', and an edited caption ``a cat is playing a ball''. In contrast, instructional image editing only needs editing instruction such as ``change the dog to a cat''. 
This experience mimics \textit{how humans naturally perform image editing}.

Instructional image editing was first proposed in InstructPix2Pix \cite{brooks2022instructpix2pix}, which fine-tunes a pre-trained stable diffusion \cite{rombach2022high} by curating a triplet of the original image, instruction, and edited image, 
with the help of GPT-3 \cite{Brown2020GPT3} and Prompt-to-Prompt image editing \cite{hertz2022prompt}. 
Though achieving promising results,
the training data generation process of InstructPix2Pix lacks explicit alignment between editing instructions and edited images.

Consequently, the modified images may only align to a certain extent with the editing instructions, as shown in the second column of Fig. \ref{fig:result_official_v1_rw}. Furthermore, since these editing instructions are provided by human users, it's crucial that the final edited images accurately reflect the users' true intentions and preferences. Typically, humans prefer to make selective changes to the original images, which are usually not factored into the training data or objectives of InstructPix2Pix \cite{brooks2022instructpix2pix}. Considering this observation and the recent successes of ChatGPT \cite{chatGPT}, we propose to refine the stable diffusion process with human feedback. This adjustment aims to ensure that the edited images more closely correspond to editing instructions provided by humans.

For large language models (LLMs) such as InstructGPT \cite{chatGPT,Ouyang2022instructgpt}, 
we often first learn a reward function to reflect what humans care about or prefer on the generated text output, and then leverage reinforcement learning (RL) algorithms such as proximal policy optimization (PPO) \cite{schulman2017ppo} to fine-tune the models.
This process is often referred to as \textit{reinforcement learning with human feedback} (RLHF).
Leveraging RLHF to fine-tune diffusion-based generative models, however,  remains challenging.
Applying on-policy algorithms (\eg,PPO) to maximize rewards during the fine-tuning process can be prohibitively expensive due to the hundreds or thousands of denoising steps required for each sampled image. 
Moreover, even with fast sampling methods \cite{song2020denoising,xiao2021tackling,karras2206elucidating,lu2022dpm}, 
it is still challenging to back-propagate the gradient signal to the parameters of the U-Net. \footnote{ We present a rigorous discussion on the difficulty in Appendix~\ref{sec:appendix_sampling_methods}.}

To address the technical issues described above, 
we propose Harnessing \textbf{H}uman Feedback for \textbf{I}nstructional \textbf{V}isual \textbf{E}diting (\textbf{HIVE}), which allows us to fine-tune diffusion-based generative models with human feedback.
As shown in Fig. \ref{fig:method}, HIVE consists of three steps:

\textbf{1)}~ We perform instructional supervised fine-tuning on the dataset that combines our newly collected 1.1M training data and the data from InstructPix2Pix. Since observing failure cases and suspecting the grounding visual components from image to instruction is still a challenging problem, we collect 1.1M training data.

\textbf{2)}~ 
For each input image and editing instruction pair, we ask human annotators to rank variant outputs of the fine-tuned model from step 1, which gives us a reward learning dataset. 
Using the collected dataset, we then train a reward model (RM) that reflects human preferences. 

\textbf{3)}~ We estimate the reward for each training data used in step 1, and integrate the reward to perform human feedback diffusion model finetuning using our proposed objectives presented in Sec.~\ref{sec:hfdf}.

Our main contributions are summarized as follows:

\noindent $\bullet$~ To tackle the technical challenge of fine-tuning diffusion models using human feedback, we introduce two scalable fine-tuning approaches in Sec.~\ref{sec:hfdf}, which are computationally efficient and offer similar costs compared with supervised fine-tuning. Moreover, we empirically show that human feedback is an essential component to boost the performance of instructional image editing models. 

\noindent $\bullet$~ To explore the fundamental ability of instructional editing, we create a new dataset for HIVE including three sub-datasets: a new 1.1M \textbf{training dataset},  a 3.6K \textbf{reward dataset} for rewards learning, and a 1K \textbf{evaluation dataset}. 

\noindent  $\bullet$~ To increase the diversity of the data for training, we introduce cycle consistency augmentation based on the inversion of editing instruction. Our dataset has been enriched with one pair of data for bi-directional editing.

 \section{Related Work}

\myparagraph{Text-To-Image Generation.} Text-to-image generative models have achieved tremendous success in the past decade. Generative adversarial nets (GANs) \cite{Goodfellow2014GAN} is one of the fundamental methods that dominated the early-stage works \cite{reed2016generative,han2017stackgan,xu2018Attngan}.
Recently, diffusion models \cite{sohl15deep,ho2020denoising,song2019generative,song2020denoising} have achieved state-of-the-art text-to-image generation performance. \cite{dhariwal2021diffusion,nichol2021glide,ramesh2021zero,ramesh2022hierarchical,saharia2022photorealistic,yu2022scaling,rombach2022high,li2023gligen}.  
As a result, instead of training a text-to-image model from scratch,  our work focuses on \textit{fine-tuning} existing stable diffusion model \cite{rombach2022high}, by leveraging additional human feedback.

\myparagraph{Image Editing.}
Similarly, 
diffusion models based image editing methods, \eg SDEdit \cite{meng2021sdedit}, BlendedDiffusion \cite{avrahami2022blended}, BlendedLatentDiffusion \cite{avrahami2022blendedlatent}, DiffusionClip \cite{Kim_2022_diffusionclip}, EDICT \cite{wallace2022edict} or MagicMix \cite{liew2022magicmix}, have garnered significant attention in recent years.
To leverage a pre-trained image-text representation (\eg, CLIP \cite{radford2021learning}, BLIP \cite{li2022blip}) and text-to-image diffusion based pre-trained models \cite{ramesh2022hierarchical,saharia2022photorealistic,rombach2022high},
most existing works focus on text-based localized editing \cite{bar2022text2live,meng2022sdedit,hertz2022prompt}.
Prompt-to-Prompt \cite{hertz2022prompt} edits the cross-attention layer in Imagen and stable diffusion to control the similarity of image and text prompt.
ControlNet~\cite{zhang2023adding} and UniControl~\cite{qin2023unicontrol} adopt controllable conditions to control image editings.
Recently, InstructPix2Pix \cite{brooks2022instructpix2pix} tackle the problem via a different approach, requiring only human-readable editing instruction to perform image editing. 
Our work follows the same direction as InstructPix2Pix\cite{brooks2022instructpix2pix} and leverages human feedback to address the misalignment between editing instructions and resulting edited images.

\myparagraph{Learning with Human Feedback.}
Incorporating human feedback into the learning process can be a highly effective way to enhance performance across various tasks
 such as fine-tuning LLMs \cite{chatGPT,Bai2022training,Scheurer2022training,Ouyang2022instructgpt,stiennon2020learning}, robotic simulation \cite{christiano2017deepreinforcement,Ibarz2018reward}, computer vision \cite{pinto2023tuning}, and to name a few.
Many existing works leverage PPO \cite{schulman2017ppo} to align to human feedback, however on-policy RL algorithms are not suitable for diffusion-based model fine-tuning (See more discussion in Appendix \ref{sec:appendix_sampling_methods}).

Simultaneously, several concurrent works \cite{wu2023better,xu2023imagereward,Lee2023aligning} study the text-to-image generation problem using human feedback.
 ReFL \cite{xu2023imagereward} investigates how to back-propagate the reward signal to random latter denoising step in the diffusion process, while \cite{wu2023better} explores how to design fine-grained
 human preference score to improve the generation quality. 
\cite{Lee2023aligning} leverages human feedback to align text-to-image generation, where they naively view reward as weights to perform maximum likelihood training. 
Different from the above works, our work tackles the problem of \textit{instructional image editing}, where there are \textit{little or even no ground truth data} for the alignment between \textit{human-readable} editing instructions and {edited images}. 
In addition, the conditions on both image input and instructions make the human feedback more valuable than standard text-to-image tasks, since the conditions make the training harder than standard text-to-image tasks.

\vspace{-.5em}
\section{Methodology}
\vspace{-.3em}
In this section, we introduce the new datasets we collected in Sec.~\ref{sec:dataset}, and explain the three major steps of HIVE in the rest of the section.  
Concretely, we introduce the instructional supervised training in Sec.~\ref{sec:instructional_supervised_training},
 and describe how to train a reward model to score edited images in Sec.~\ref{sec:rewardslearning}, then present two scalable fine-tuning methods to align diffusion models with human feedback in Sec.~\ref{sec:policy_update}.

\vspace{-.3em}
\subsection{Dataset}
\vspace{-.3em}
\label{sec:dataset}

\myparagraph{Instructional Edit Training Dataset.} We follow the same method of \cite{brooks2022instructpix2pix} to generate the training dataset. We collect 1K images and their corresponding captions. We ask three annotators to write three instructions and corresponding edited captions based on the collected input captions. Therefore, we obtain 9K prompt triplets: input caption, instruction, and edited caption. We fine-tune GPT-3 \cite{Brown2020GPT3} with OpenAI API v0.25.0 \cite{openaiapi} with them. 
We use the fine-tuned GPT-3 to generate five instructions and edited captions per input image-caption pair in Laion-Aesthetics V2 \cite{schuhmann2022laion5B}. We observe that the captions from Laion are not always visually descriptive, so we use BLIP \cite{li2022blip} to generate more diverse types of image captions. Later stable diffusion based Prompt-to-Prompt \cite{hertz2022prompt} is adopted to generate paired images. In addition, we design a cycle-consistent augmentation method (Sec.~\ref{sec:cycleconsistency}) to generate additional training data. We generate 1.17M training triplets in total. 
Combining the 281K training data from \cite{brooks2022instructpix2pix}, we obtain 1.45M training image pairs along with instructions.

\myparagraph{Reward Fine-tuning Dataset.} We collect 3.6K image-instruction pairs for the task of reward fine-tuning. Among them, 1.6K image-instruction pairs are manually collected, and the rest are from Laion-Aesthetics V2 with GPT-3 generated instructions. We use this dataset to ask annotators to rank various model outputs.

\myparagraph{Evaluation Dataset.} We use two evaluation datasets: the test dataset in \cite{brooks2022instructpix2pix} for quantitative evaluation and a new 1K dataset collected for the user study. The quantitative evaluation dataset is generated following the same method as the training dataset, which means that the dataset does not contain real images. Our collected 1K dataset contains 200 real images, and each image is annotated with five human-written instructions. More details of annotation tooling, guidelines, and analysis are in Appendix~\ref{sec:appendix_dataset}.

\subsection{Instructional Supervised Training}
\label{sec:instructional_supervised_training}
We follow the instructional fine-tuning method in \cite{brooks2022instructpix2pix} with two major upgrades on dataset curation (Sec.~\ref{sec:dataset}) and cycle consistency augmentation (Sec.~\ref{sec:cycleconsistency}). A pre-trained stable diffusion model \cite{rombach2022high} is adopted as the backbone architecture. 
 In instructional supervised training,  the stable diffusion model has two conditions $c = \left[ c_I, c_E \right]$, where $c_E$ is the editing instruction, and $c_I$ is the latent space of the original input image.
 In the training process, a pre-trained auto-encoder \cite{kingma2013autoencoding} with encoder $\mathcal{E}$ and decoder $\mathcal{D}$ is used to convert between edited image $\tgx$ and its latent representation $z=\mathcal{E}(\tgx)$. 
 The diffusion process is composed of an equally weighted sequence of denoising autoencoders $\epsilon_{\theta}(z_t, t, c)$, $t = 1, \cdots , T$, which are trained to predict a denoised variant of their input $z_t$, a noisy version of $z$. The objective of instructional supervised training is:
 \vspace{-.9em}
\begin{align*}
L = \mathbb{E}_{\mathcal{E}(\tgx), c, \epsilon \sim \mathcal{N}(0, 1), t }\Big[ \Vert \epsilon - \epsilon_\theta(z_{t}, t, c)) \Vert_{2}^{2}\Big]\,.
\vspace{-.5em}
\label{equ:obj_ip2p}
\end{align*}
\vspace{-1.5em}
\subsubsection{Cycle Consistency Augmentation}
\vspace{-.2em}
\label{sec:cycleconsistency}

Cycle consistency is a powerful technique that has been widely applied in image-to-image generation~\cite{zhu2017unpaired,isola2017image}. It involves coupling and inverting bi-directional mappings of two variables $X$ and $Y$, $G: X \rightarrow Y$ and $F: Y \rightarrow X$, such that $F(G(X)) \approx X$ and vice versa. This approach has been shown to enhance generative mapping in both directions.

While Instructpix2pix~\cite{brooks2022instructpix2pix}  considers instructional image editing as a single-direction mapping, we propose adding cycle consistency. Our approach involves a forward-pass editing step, $F: x \stackrel{inst}{\longrightarrow} \tgx$. We then introduce instruction reversion to enable a reverse-pass mapping, $R: \tgx \stackrel{\sim inst}{\longrightarrow} x$. In this way, we could close the loop of image editing as: $x \stackrel{inst}{\longrightarrow} \tgx \stackrel{\sim inst}{\longrightarrow} x$, \eg ``add a dog'' to ``remove the dog''.

To ensure the effectiveness of this technique, we need to separate invertible and non-invertible instructions from the dataset.
We devised a rule-based method that combines speech tagging and template matching. We found that most instructions adhere to a particular structure, with the verb appearing at the start, followed by objects and prepositions. Thus, we grammatically tagged all instructions using the Natural Language Toolkit (NLTK)~\footnote{https://www.nltk.org/}. We identified all invertible verbs and pairing verbs, and also analyzed the semantics of the objects and the prepositions used. By summarizing invertible instructions in predefined templates, we matched desired instructions. Our analysis revealed that 29.1\% of the instructions in the dataset were invertible. We augmented this data to create more comprehensive training data, which facilitated cycle consistency. For more information, see Appendix~\ref{sec:appendix_instructional_supervised_training}.

\subsection{Human Feedback Reward Learning}
\label{sec:rewardslearning}

The second step of HIVE is to learn a reward function $\mathcal{R}_{\phi}(\tilde{\gx}, \tc)$, which takes the original input image, the text instruction condition $c = \left[\ci, \ce \right]$, and the edited image $\tilde{\gx}$ that is generated by the fine-tuned stable diffusion as input, and outputs a scalar that reflects human preference.

\begin{figure}
\begin{center}
\includegraphics[ width=0.99\linewidth]{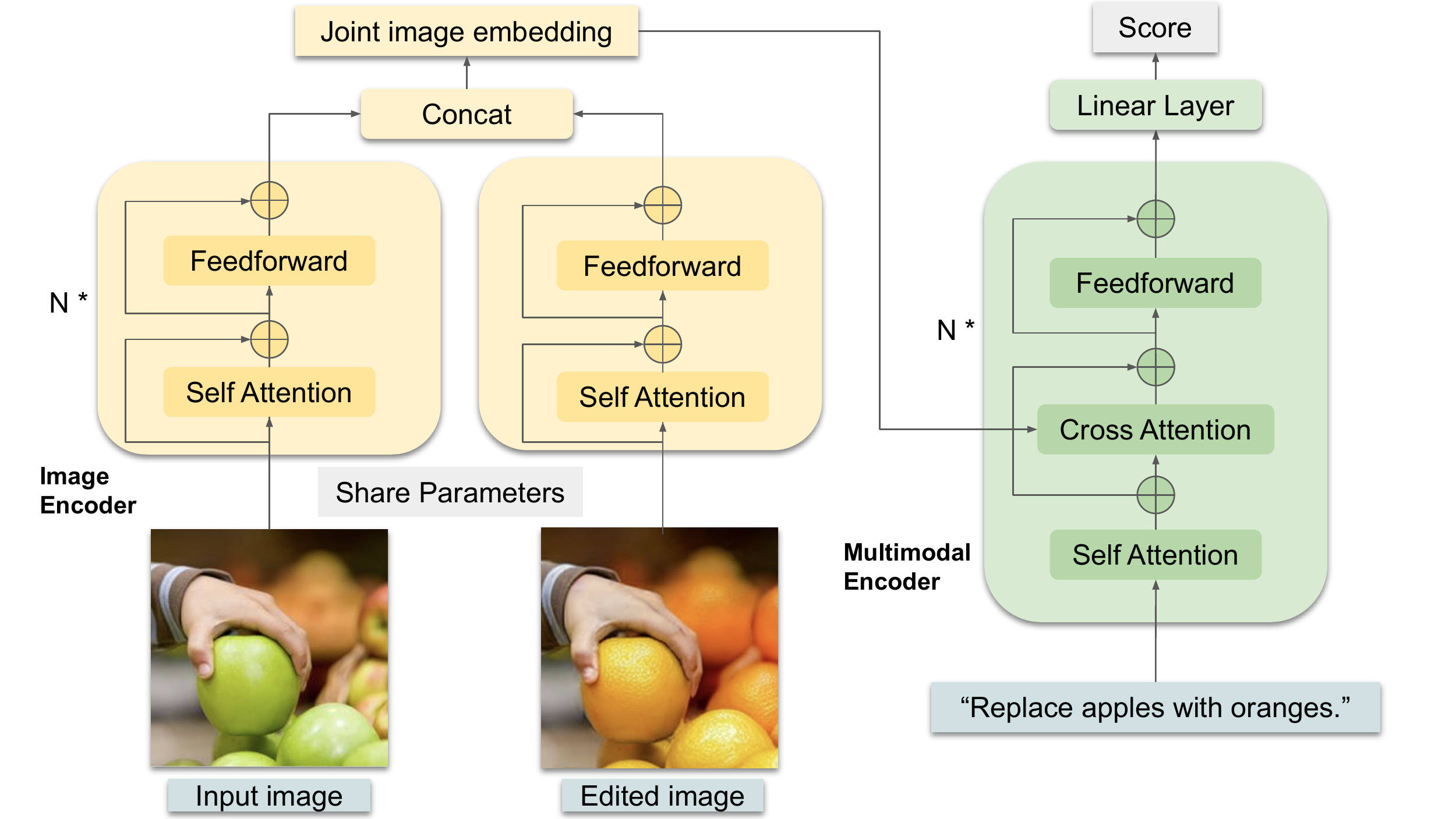}
\end{center}
\vspace{-1.7em}
   \caption{Model architecture for reward $R(\tgx, c)$. Here the reward model evaluates human preference for an edited image of a hand selecting an orange compared to the original input  image of the hand selecting an apple. The input to the reward model includes both images and a text instruction. The output is a score indicating the degree of preference for the edited image based on the input image and instruction.}
\label{fig:reward_model_architecture}
\end{figure}

Unlike InstructGPT which only takes text as input, our reward model $\mathcal{R}_{\phi}(\tilde{\gx}, \tc)$
needs to measure the alignment between instructions and the edited images.
To address the challenge, we present a reward model architecture in Fig.~\ref{fig:reward_model_architecture},
which leverages pre-trained vision-language models such as BLIP \cite{li2022blip}.
More specifically, the reward model employs an image-grounded text encoder as the multi-modal encoder to take the joint image embedding and the text instruction as input and produce a multi-modal embedding. A linear layer is then applied to the multi-modal embedding to map it to a scalar value. More details are in Appendix~\ref{sec:appendix_rm}.

With the specifically designed network architecture, 
we train the reward function $\mathcal{R}_{\phi}(\tilde{\gx}, \tc)$ with our collected reward fine-tuning dataset $\mathcal{D}_\mathrm{human}$ induced in Sec.~\ref{sec:dataset}. 
For each input image $\ci$ and instruction $\ce$ pair,
we have $K$ edited images $\{\tilde{\gx}\}_{k=1}^{K}$ ranked by human annotators,
and denote the human preference of edited image $\tgx_i$ over $\tgx_j$  by  $\tgx_i \succ \tgx_j $. 
Then we can follow the Bradley-Terry model
of preferences \cite{bradley1952rank,Ouyang2022instructgpt} to define the pairwise loss function:
\vspace{-.5em}
\begin{equation*}\textstyle
\resizebox{0.9\linewidth}{!}{%
$
 \ell_{\mathrm{RM}}(\phi) := - \sum_{\tgx_i \succ \tgx_j}\log \left[\frac{\exp(\mathcal{R}_{\phi}(\tgx_i, c))}{\sum_{k=i, j}\exp(\mathcal{R}_{\phi}(\tgx_k, c))}\right]\,,
$%
}
\end{equation*}
where $(i, j)\in [1\ldots K]$ and we can get $K \choose 2$ pairs of comparison for each condition $c$.
Similar to \cite{Ouyang2022instructgpt}, we put all the $K \choose 2$ pairs for each condition $c$ in a single batch to learn the reward functions. 
We provide a detailed reward model training discussion in Appendix~\ref{sec:appendix_rm}.

\subsection{Human Feedback based Model Fine-tuning}\label{sec:hfdf}
With the learned reward function $\mathcal{R}_{\phi}(c, \tgx)$, 
the next step is to improve the instructional supervised training model
by reward maximization. As a result, we can obtain an instructional diffusion model that aligns with human preferences.

The RL fine-tuning techniques we present are built upon recent offline RL techniques \cite{levine2020offline,peng2019advantage,chen2021decision,janner2021sequence}
With an input image and editing instruction condition $c=[\ci, \ce]$, 
we define the edited image data distribution generated by the instructional supervised diffusion model as $p(\tgx|~c)$,
and the edited image data distribution generated by the current diffusion model we want to optimize as $\rho(\tgx | ~c)$\,,
then under the pessimistic principle of offline RL, we can optimize $\rho$ by the following objectives:
\vspace{-.5em}
\begin{align}\textstyle
    J(\rho) := \max_{\rho} \E_{c}\big[&\E_{\tgx \sim \rho(\cdot | c)}[\mathcal{R}_{\phi}(\tgx, c)] -\notag \\
        & ~~~~~~~~~~~~\eta \mathrm{KL}(\rho(\tgx | c) ||p(\tgx | c) ) \big]\,, \label{equ:rl_obj}
\end{align}
where $\eta$ is a hyper-parameter.
The first term in Eq.~\eqref{equ:rl_obj} is the standard reward maximization in RL. The second term is a regularization to stabilize learning, which is a widely used 
technique in offline RL \cite{kumar2020conservative}, and is also adopted for PPO fine-tuning of InstructGPT (\textit{a.k.a} ``PPO-ptx'') \cite{Ouyang2022instructgpt}.

To avoid using sampling-based methods to optimize $\rho$, we can
differentiate $J(\rho)$ \textit{w.r.t} $\rho(\tgx | c )$ and solve for the optimal $\rho^{*}(\tgx | c)$, resulting the following expression for the optimal solution of Eq.~\eqref{equ:rl_obj}:
\vspace{-.5em}
\begin{align}
    \rho^{*}(\tgx | c) \propto p(\tgx |c)\exp\left(\mathcal{R}_{\phi}(\tgx ,c) / \eta\right)\,, \label{equ:optim_q}
\end{align}
or $\rho^{*}(\tgx | c) = \frac{1}{Z(c)}p(\tgx |c)\exp\left(\mathcal{R}_{\phi}(\tgx ,c) / \eta\right)$, with $Z(c) = \int p(\tgx |c)\exp\left(\mathcal{R}_{\phi}(\tgx ,c) / \eta\right) d\tgx$ being the partition function.  A detailed derivation is in Appendix~\ref{sec:appendix_derivation}.

\myparagraph{Weighted Reward Loss.} The optimal target distribution $\rho^{*}(\tgx |c)$ in Eq.~\eqref{equ:optim_q}  can be viewed as an exponential reward-weighted distribution for $p(\tgx | c)$. 
Moreover, we have already obtained the empirical edited image data drawn from $p(\tgx |c)$ when constructing the instructional editing dataset, 
and we can view the exponential reward weighted edited image $\tgx$ from the instructional editing dataset as an empirical approximation of samples drawn from $\rho^{*}(\tgx | c)$. Formally, we can fine-tune a diffusion model thus it generates data from $\rho^{*}(\tgx | c)$, resulting in the weighted reward loss:
\begin{equation*}\textstyle
\resizebox{0.95\linewidth}{!}{%
$
  \ell_\mathrm{WR}(\theta) := \E_{\mathcal{E}(\tgx), c, \epsilon \sim \mathcal{N}(0, 1), t }\left[\omega(\tgx, c)\cdot \left\Vert \epsilon - \epsilon_\theta(z_{t}, t, c) \right\Vert_{2}^{2}\right]\,,
$%
}
\end{equation*}
with $\omega(\tgx, c) = \exp\left(\mathcal{R}_{\phi}(\tgx, c) / \eta\right) $ being the \textit{exponential reward weight} for edited image $\tgx$ and condition $c$. 
Different from RL literature \cite{peters2010relative,peng2019advantage} using exponential reward or advantage weights to learn a policy function,
our weighted reward loss is derived for fine-tuning stable diffusion.

\myparagraph{Condition Reward Loss.} We can also leverage the control-as-inference perspective of RL \cite{levine2018reinforcement} 
to transform Eq.~\eqref{equ:optim_q} to a conditional reward expression, thus we can directly view the reward as a conditional label to fine-tune diffusion models. 
Similar to \cite{levine2018reinforcement}, we introduce a new binary variable $R^{*}$ indicating whether human prefers the edited image or not, where $R^{*} = 1$ denotes that human prefers the edited image, and $R^{*} = 0$ denotes that human does not prefer, thus we have $p(R^{*} = 1 ~| ~\tgx, c) \propto \exp\left(\mathcal{R}_{\phi}(\tgx ,c)\right)$.
Together with Eq.~\eqref{equ:optim_q}, and applying Bayes rules gives us the following derivation: 
\vspace{-1em}
\begin{align*}\textstyle
    p(\tgx |c)&\exp\left(\mathcal{R}_{\phi}(\tgx ,c) / \eta\right) := q(\tgx | c) \left(p(R^{*} = 1 ~| ~\tgx, c)\right)^{1 / \eta}  \\
    & = p(\tgx | c) \left(\frac{p(\tgx|~R^{*} = 1, c)p(R^{*} = 1|~c)}{p(\tgx | c)}\right)^{1 / \eta} \\
    & \propto  p(\tgx | c)^{1 - 1  / \eta}p(\tgx|~R^{*} = 1, c)^{1 / \eta}\,,
\end{align*}
where we drop $p(R^{*} = 1|~c)$ since it is a constant w.r.t $\tgx$. 
We can now view the reward for each edited image as an additional condition. 
Define the new condition $\tilde{c}=[\ci, \ce, c_{R}]$, with $c_{R}$ as the reward label, we can fine-tune the diffusion model with the condition reward loss:
\vspace{-.5em}
\begin{align*}\textstyle
    \ell_\mathrm{CR}(\theta) = \mathbb{E}_{\mathcal{E}(x), \tilde{c}, \epsilon \sim \mathcal{N}(0, 1), t }\Big[ \Vert \epsilon - \epsilon_\theta(z_{t}, t, \tilde{c}) \Vert_{2}^{2}\Big]\,.
\end{align*}

\begin{figure*}[t]
\begin{center}
\includegraphics[ width=0.99\linewidth]{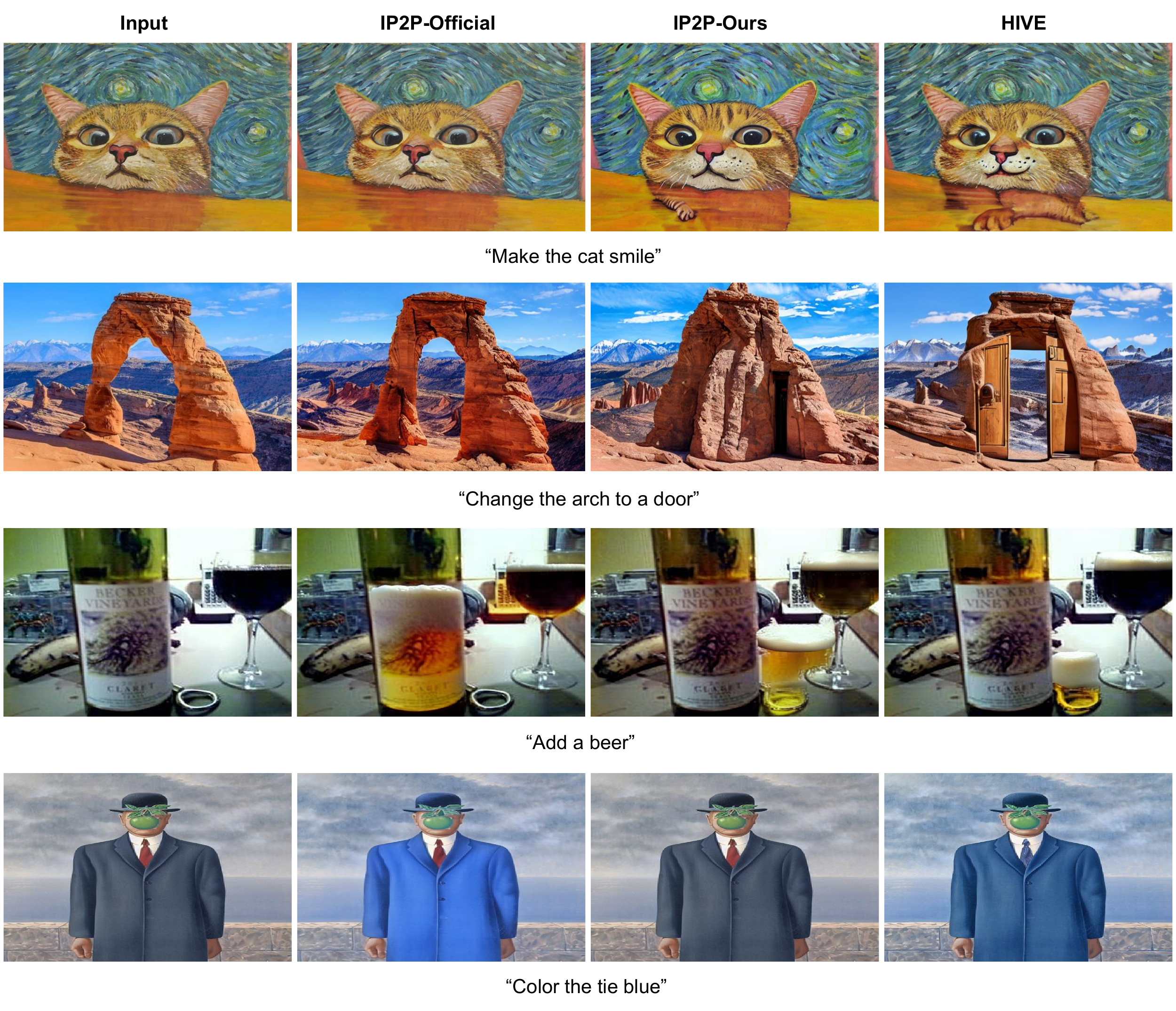}
\end{center}
\vspace{-2.5em}
   \caption{Comparisons between IP2P-Official (InstructPix2Pix official model), IP2P-Ours (InstructPix2Pix using our data) and HIVE. HIVE can boost performance by understanding the instruction correctly.}
\label{fig:result_official_v1_rw}
\vspace{-1em}
\end{figure*}

 We quantize the reward into five categories, based on the quantile of the empirical reward distribution of the training dataset, and convert the reward value into a text prompt. For instance, if the reward value of a training pair lies in the bottom 20\% of the reward distribution of the dataset, then we convert the reward value as a text prompt condition $c_{R}:=$\textit{``The image quality is one out of five''}. 
 And during the inference time to generate edited images, we fix the text prompt as $c_{R}:=$\textit{``The image quality is five out of five''}, indicating we want the generated edited images with the highest reward. 
 We empirically find this technique improves the stability of fine-tuning.

\label{sec:policy_update}

\vspace{-.2em}
\section{Experiments}\label{sec:exp}
\vspace{-.3em}

This section presents the experimental results and ablation studies of HIVE's technical choices, demonstrating the effectiveness of our method. We adopt the default guidance scale parameters in InstrcutPix2Pix for a fair comparison. Through our experiments, we discovered that the conditional reward loss performs slightly better than the weighted reward loss, and therefore, we present our results based on the conditional reward loss.  The detailed comparisons can be found in Sec.~\ref{sec:ablation} and Appendix~\ref{sec:appendix_rewards_exp}.

\begin{figure}[t]
\begin{center}
\includegraphics[ width=1\linewidth]{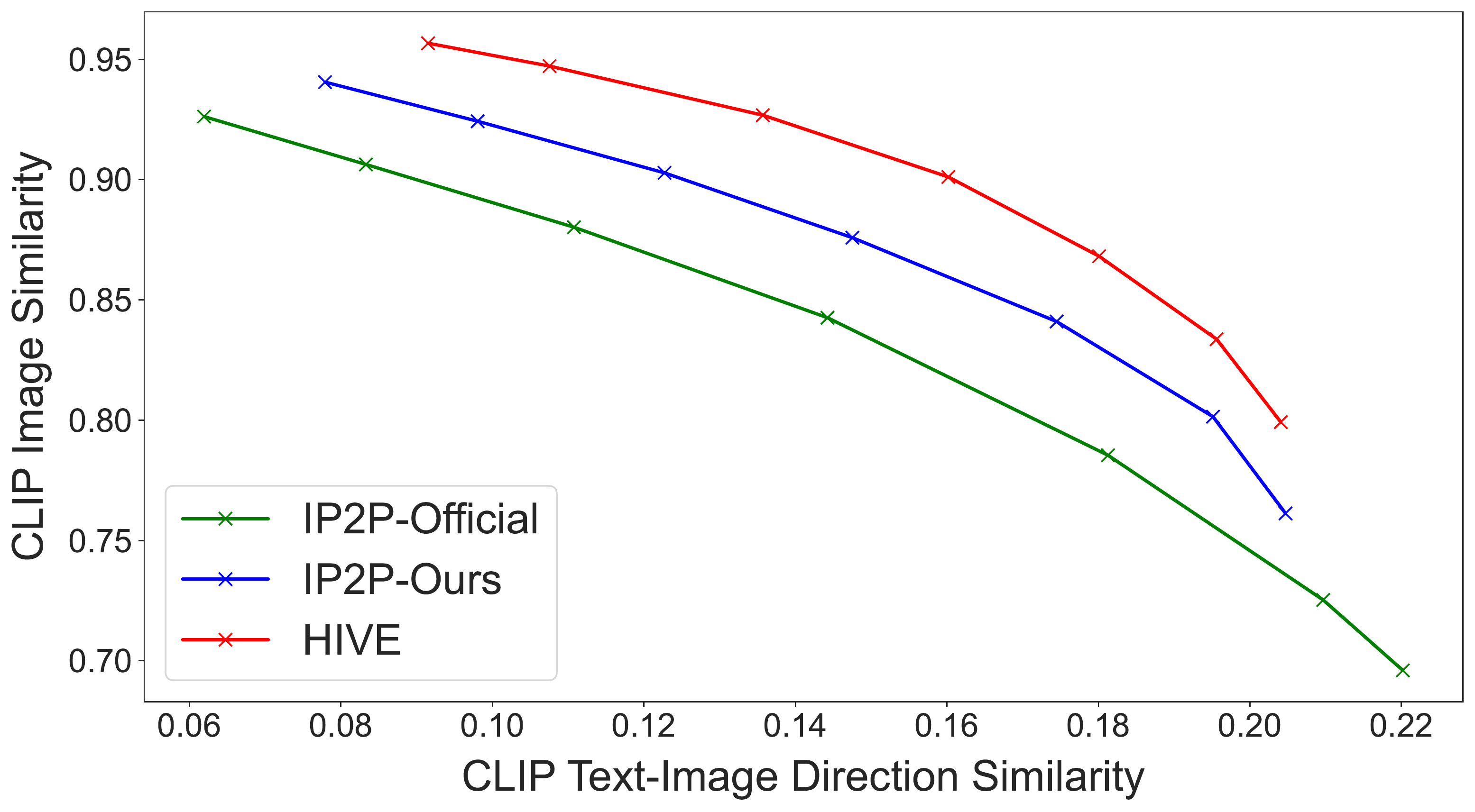}
\end{center}
\vspace{-1.8em}
   \caption{Comparisons between IP2P-Official, IP2P-Ours, and HIVE. It plots tradeoffs between consistency with the input image and consistency with the edit. The higher the better. For all methods, we adopt the same parameters as that in \cite{brooks2022instructpix2pix}.}
\label{fig:plot_instructpix2pix_vs_v1_wofeedback}
\vspace{-1em}
\end{figure}

We evaluate our method using two datasets: a synthetic evaluation dataset with 15,652 image pairs from \cite{brooks2022instructpix2pix} and a self-collected 1K evaluation dataset with real image-instruction pairs. For the synthetic dataset, we follow InstructPix2Pix's quantitative evaluation metric and plot the trade-offs between CLIP image similarity and directional CLIP similarity~\cite{Gal2022stylegannana}. For the 1K dataset, we conduct a user study where for each instruction, the images generated by competing methods are reviewed and voted by three human annotators, and the winner is determined by majority votes.

\begin{figure}[t]
    \centering
    \begin{tabular}{cc}
         \hspace{-0.5em}\includegraphics[width=.23\textwidth]{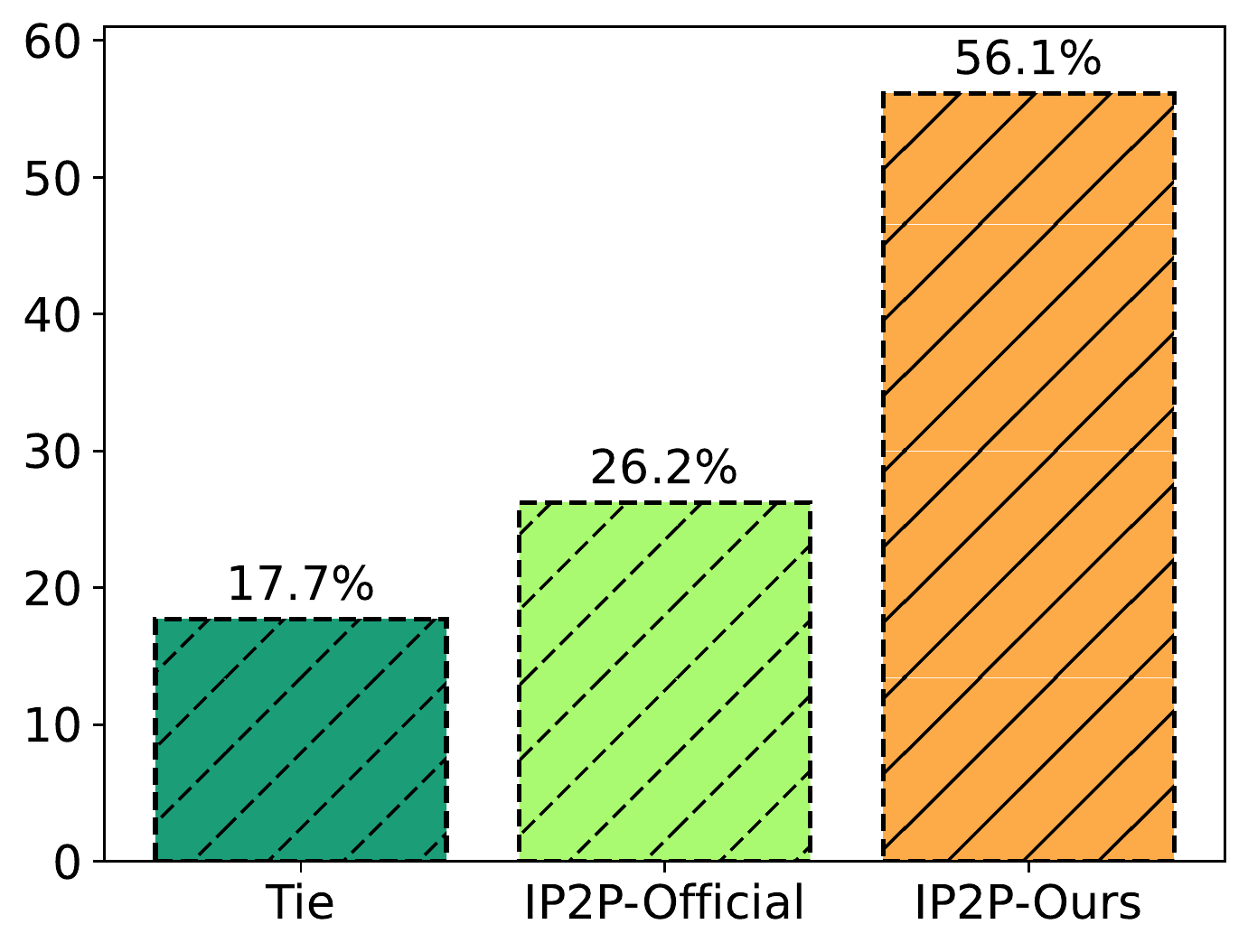}  &  
         \hspace{-1em}\includegraphics[width=.23\textwidth]{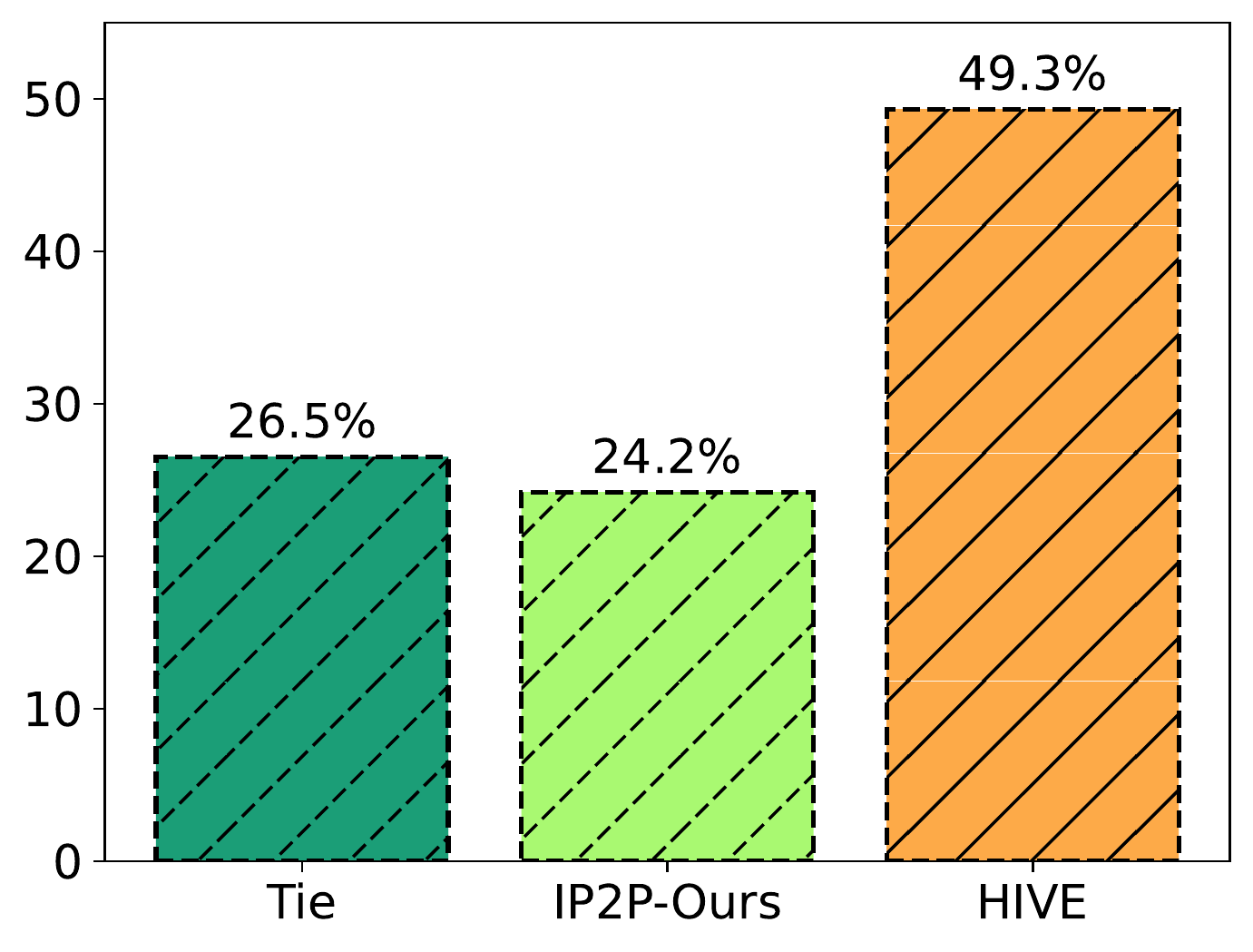}  \\
         \small{IP2P-Official vs IP2P-Ours} & \small{IP2P-Ours vs HIVE} \\
       
    \end{tabular}
    
    \vspace{-1em}
    \caption{User study of comparison between (a) IP2P-Official vs IP2P-Ours and (b) IP2P-Ours and HIVE. IP2P-Ours obtains 30\% more votes than IP2P-Official. HIVE obtains 25\% more votes than that IP2P-Ours.}
    \label{fig:userstudy_instructpix2pix_vs_hive}
    \vspace{-4mm}
\end{figure}

\subsection{Baseline Comparisons}
\label{sec:exp_baseline}

We perform experiments with the same setup as InstructPix2Pix, where stable diffusion (SD) v1.5 is adopted. We compare three models: InstructPix2Pix official model (IP2P-Official), InstructPix2Pix using our data (IP2P-Ours) \footnote{It is the same to HIVE without human feedback.}, and HIVE. We report the quantitative results on the synthetic evaluation dataset in Fig.~\ref{fig:plot_instructpix2pix_vs_v1_wofeedback}. We observe that IP2P-Ours improves notably over IP2P-Official (blue curve vs. green curve). Moreover, human feedback further boosts the performance of HIVE (red curve vs blue curve) over IP2P-Ours by a large margin. In other words, with the same directional similarity value, HIVE obtains better image consistency than InstructPix2Pix. 

 To test the effectiveness of HIVE on real-world images, we report the user study results on the 1K evaluation dataset. We use ``Tie'' to represent that users think results are equally good or equally bad. As shown in Fig.~\ref{fig:userstudy_instructpix2pix_vs_hive}(a), IP2P-Ours gets around 30\% more votes than the IP2P-Official. The result is consistent with the user study on the synthetic dataset. We also demonstrate the user study outcome between HIVE and IP2P-Ours in Fig.~\ref{fig:userstudy_instructpix2pix_vs_hive}(b). The user study indicates similar conclusions to the consistency plot, where HIVE gets around 25\% more favorites than IP2P-Ours.

 In Fig.~\ref{fig:result_official_v1_rw}, we present representative edits that demonstrate the effectiveness of HIVE. The results show that while using more data can partially improve editing instructions without human feedback, the reward model leads to better alignment between instruction and the edited image. For example, in the second row, IP2P-Ours generates a door-like object, but with the guidance of human feedback, the generated door matches human perception better. In the fourth row, the example of which is from the failure examples in \cite{brooks2022instructpix2pix}, HIVE can locate the tie and change its color correctly.

\begin{figure}[t]
\begin{center}
\includegraphics[ width=0.99\linewidth]{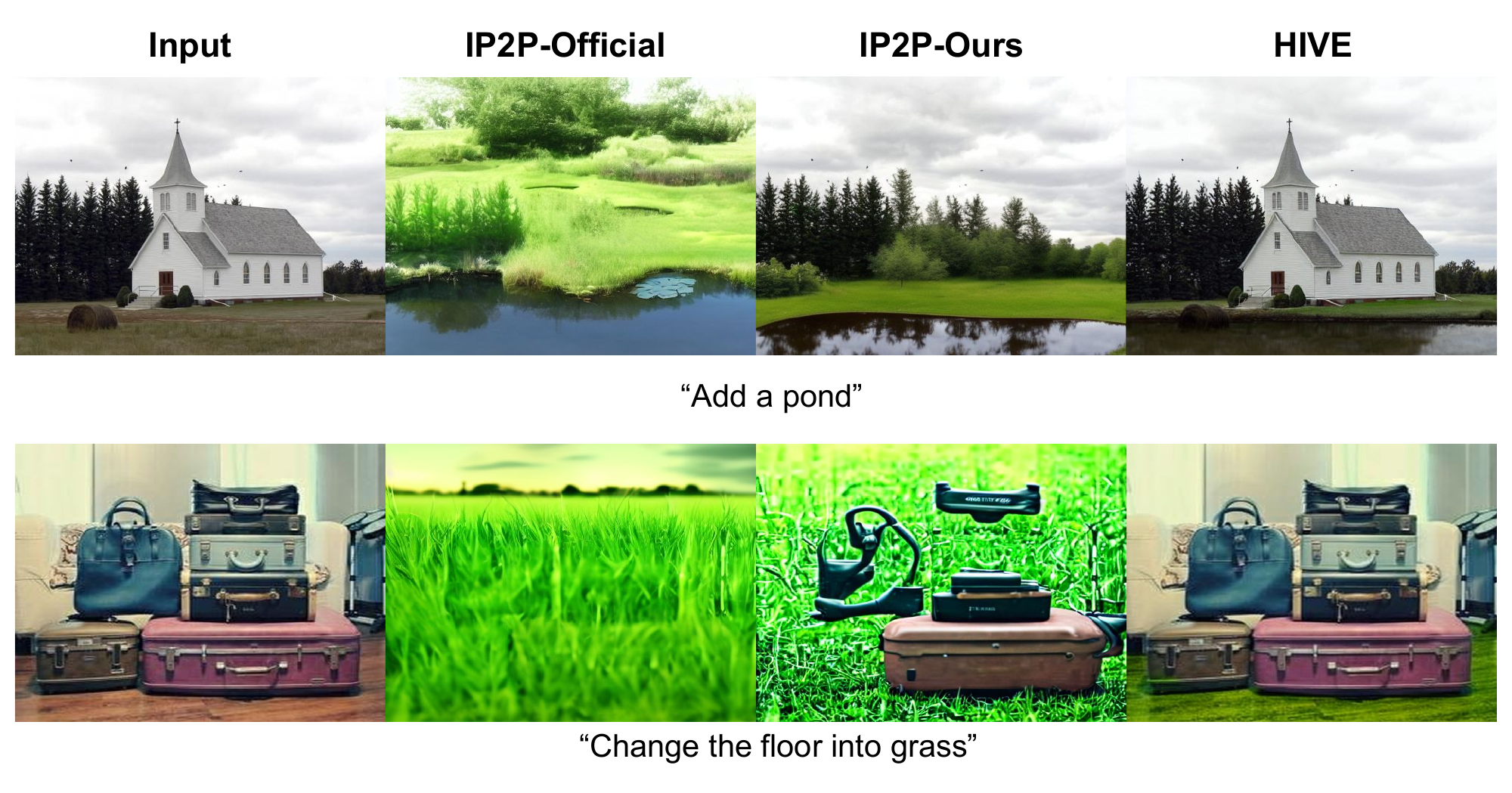}
\end{center}
    \vspace{-2.5em}
   \caption{Human feedback tends to help HIVE avoid unwanted excessive image modifications.}
\label{fig:result_preserve}
\vspace{-1em}
\end{figure}

Additionally, our visual analysis of the results (Fig.~\ref{fig:result_preserve}) indicates that the HIVE model tends to preserve the remaining part of the original image that is not instructed to be edited, while IP2P-Ours leads to excessive image editing more often. For instance, in the first example of Fig.~\ref{fig:result_preserve}, HIVE blends a pond naturally into the original image. The two InstructPix2Pix models fulfill the same instruction, however, at the same time, alter the uninstructed part of the original background.

\vspace{-.3em}
\subsection{Ablation Study}
\label{sec:ablation}

\begin{figure}[t]
    \centering
    \begin{tabular}{cc}
         \hspace{-.5em}\includegraphics[width=.23\textwidth]{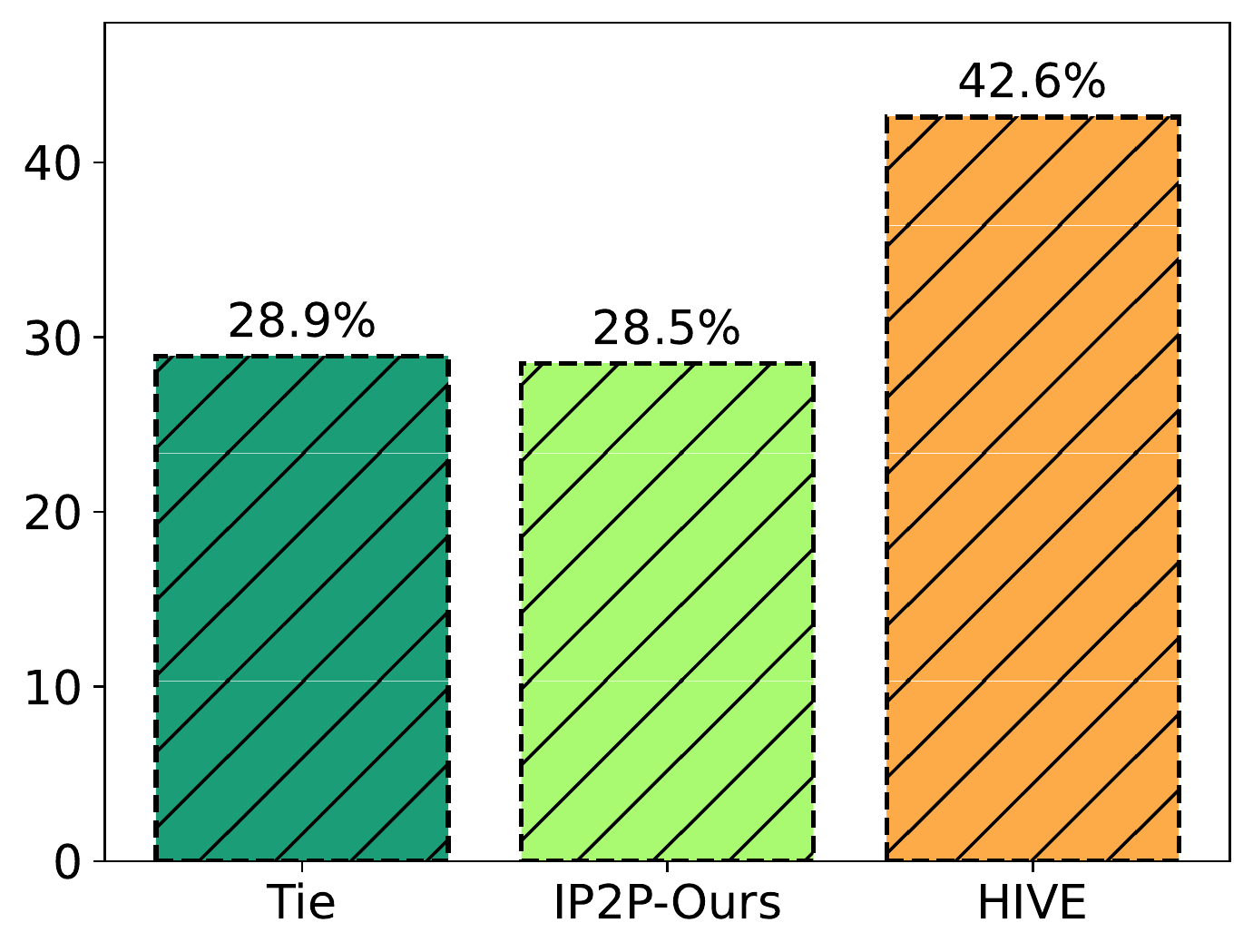}&  
         \hspace{-1em}\includegraphics[width=.23\textwidth]{Figures/user_study_hive_w_wo_feedback.pdf}   \\
         \footnotesize{HIVE with weighted reward loss} & \hspace{-2em}  \footnotesize{HIVE with condition reward loss} \\
       
    \end{tabular}
    
    \vspace{-1em}
    \caption{User study of pairwise comparison between (a) HIVE with weighted reward loss and (b) HIVE with condition reward loss. The human preferences are close to each other.}
    \label{fig:userstudy_rewardloss}
    \vspace{-1.25em}
\end{figure}

\myparagraph{Weighted Reward and Condition Reward Loss.} We perform user study on HIVE with these two losses individually. As shown in Fig.~\ref{fig:userstudy_rewardloss}, these two losses obtain similar human preferences on the evaluation dataset. More comparisons are in Appendix~\ref{sec:appendix_ablation}.

\begin{figure}[t]
    \centering
    \begin{tabular}{cc}
         \hspace{-.5em}\includegraphics[width=.23\textwidth]{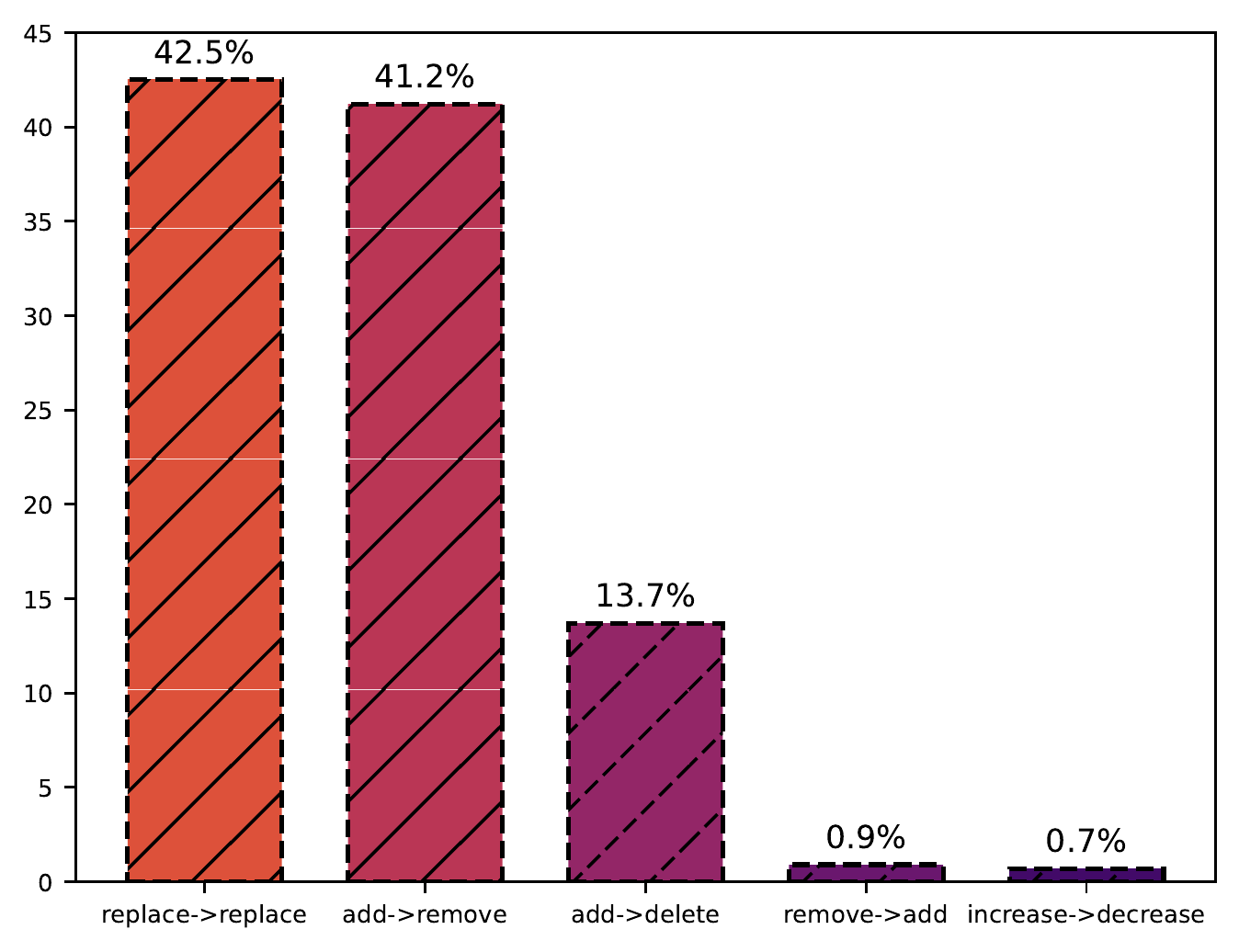}&  
         \hspace{-1em}\includegraphics[width=.23\textwidth]{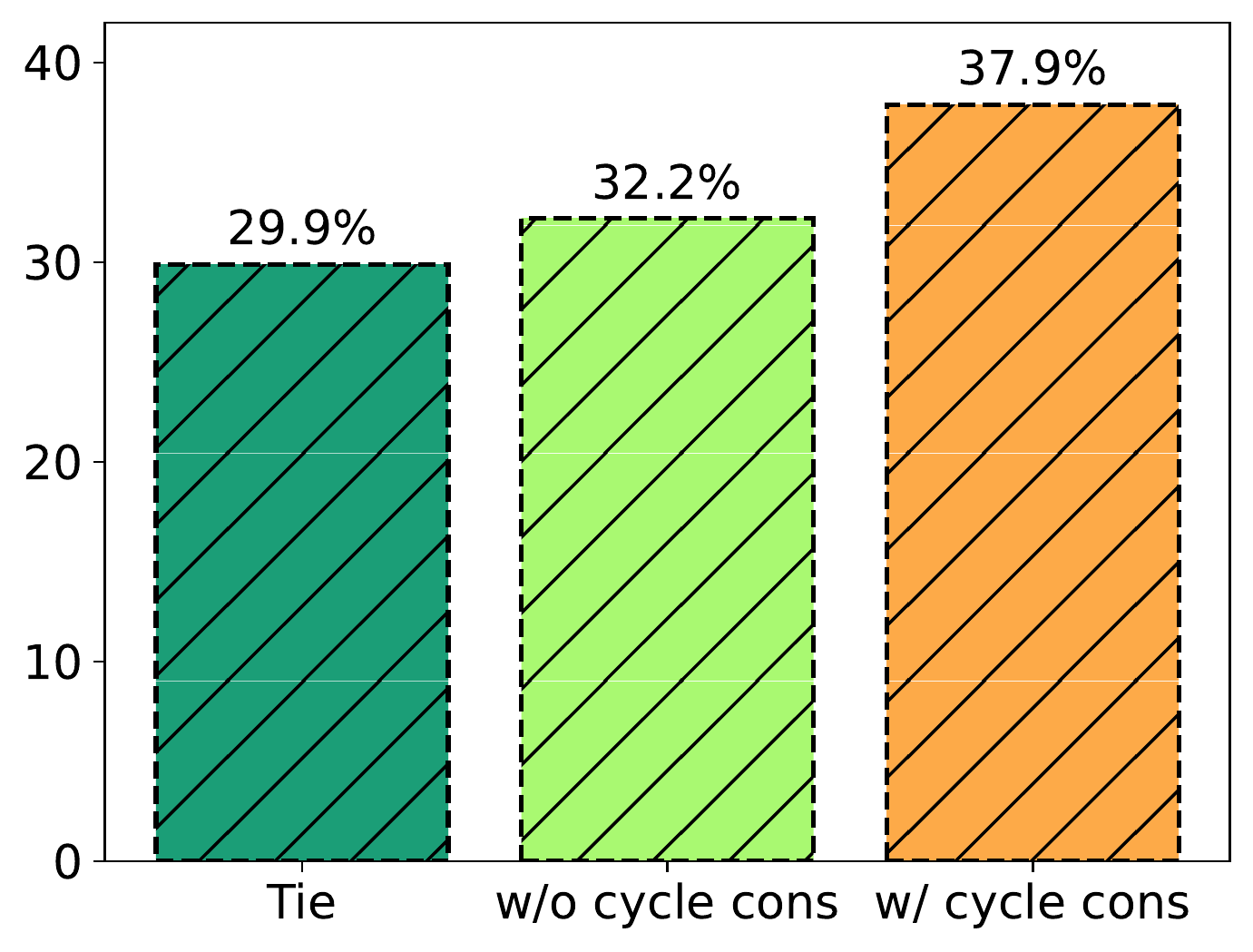}   \\
         \footnotesize{Top five augmentations.} & \hspace{-2em}  \footnotesize{User study of cycle consistency.} \\
       
    \end{tabular}
    
    \vspace{-1em}
    \caption{Cycle consistency analysis.}
    \label{fig:cycle_consistency}
    \vspace{-1.25em}
\end{figure}

\myparagraph{Cycle Consistency} We analyze the impact of it which is introduced in Sec.~\ref{sec:cycleconsistency}. The top five augmentations in the cycle consistency are demonstrated in Fig.~\ref{fig:cycle_consistency}(a). We perform evaluation on both synthetic dataset and the 1K evaluation dataset. The user study in Fig.~\ref{fig:cycle_consistency}(b) shows that the cycle consistency augmentation improves the performance of HIVE by a notable margin.

\begin{figure}[t]
    \centering
    \begin{tabular}{cc}
         \hspace{-0.5em}\includegraphics[width=.23\textwidth]{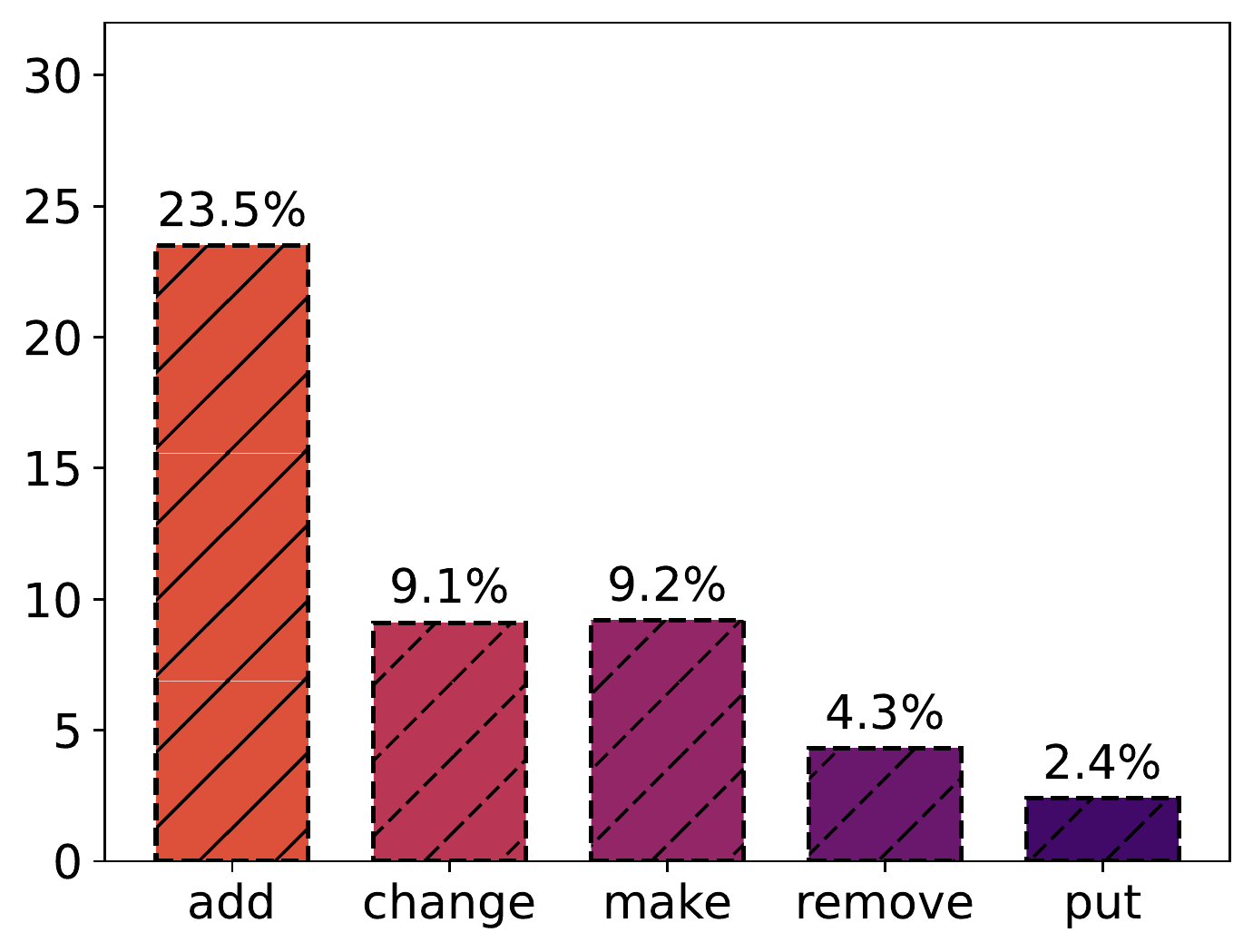}&  
         \hspace{-1em}\includegraphics[width=.23\textwidth]{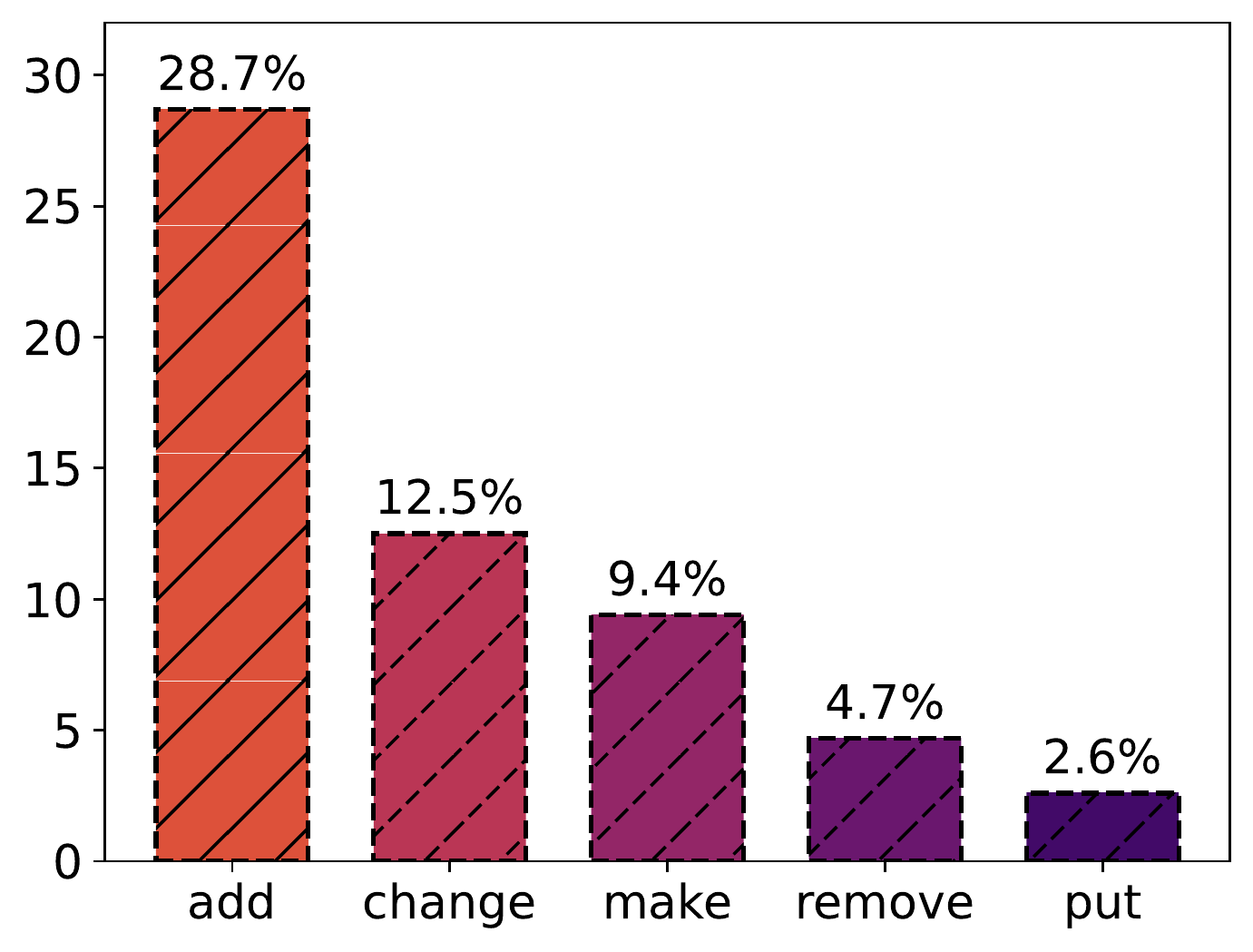}   \\
         \small{IP2P-Ours} & \small{HIVE} \\
       
    \end{tabular}
    
    \caption{Success rate of IP2P-Ours and HIVE on top five verbs.}
    \label{fig:stat_verb_success}
\end{figure}

\myparagraph{Success Rate on Verbs} It is observed that five verbs take around 85\% of all verbs, where details can be found in Sec.~\ref{sec:appendix_dataset}. We compare HIVE with IP2P-Ours on these five verbs, and report the success rate of these two methods on these verbs. It is seen in Fig.~\ref{fig:stat_verb_success} that HIVE improves the most on ``add'' from 23.5\% to 28.7\%.

\begin{figure}[t]
    \centering
    \begin{tabular}{cc}
         \hspace{-.5em}\includegraphics[width=.23\textwidth]{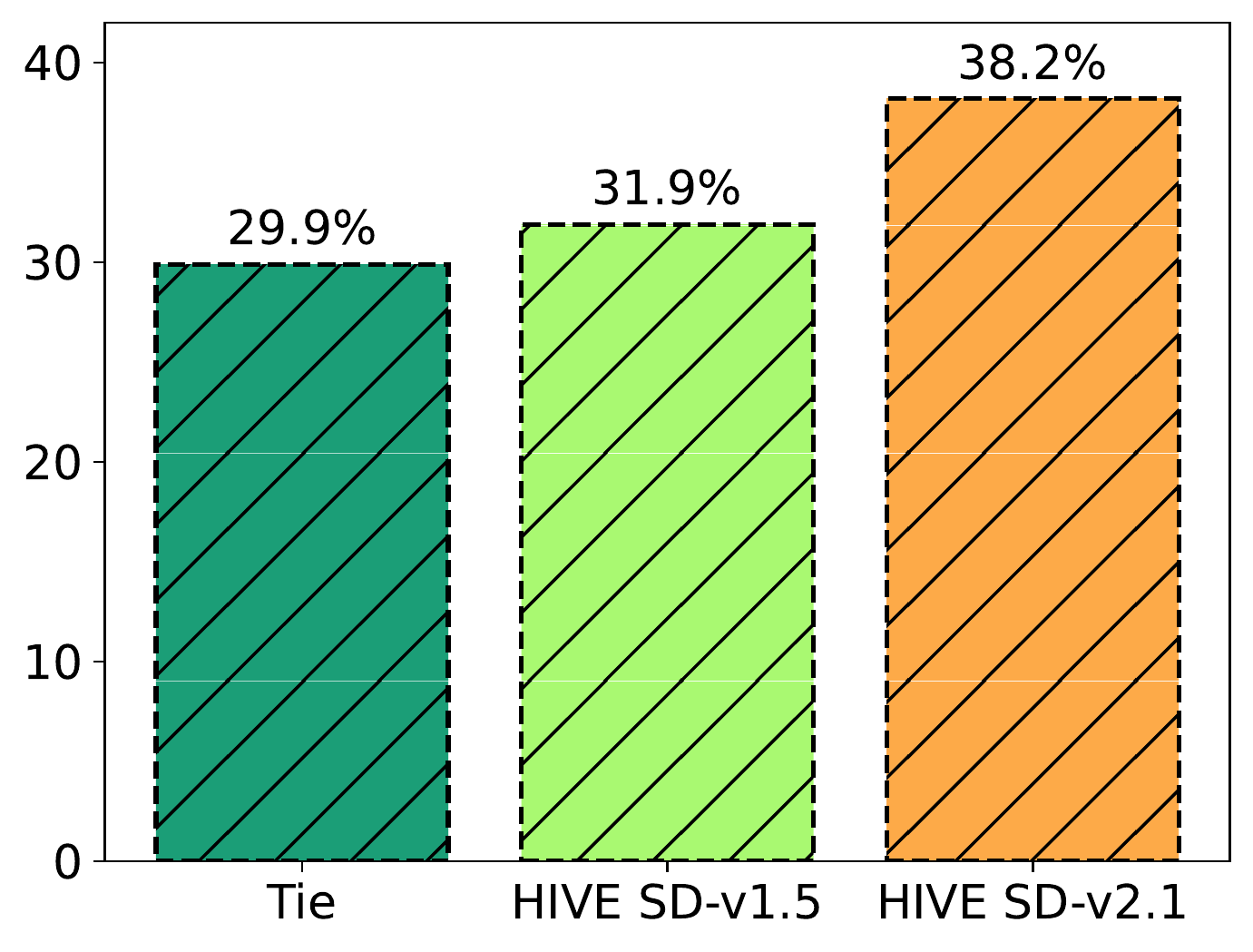}&  
         \hspace{-1em}\includegraphics[width=.23\textwidth]{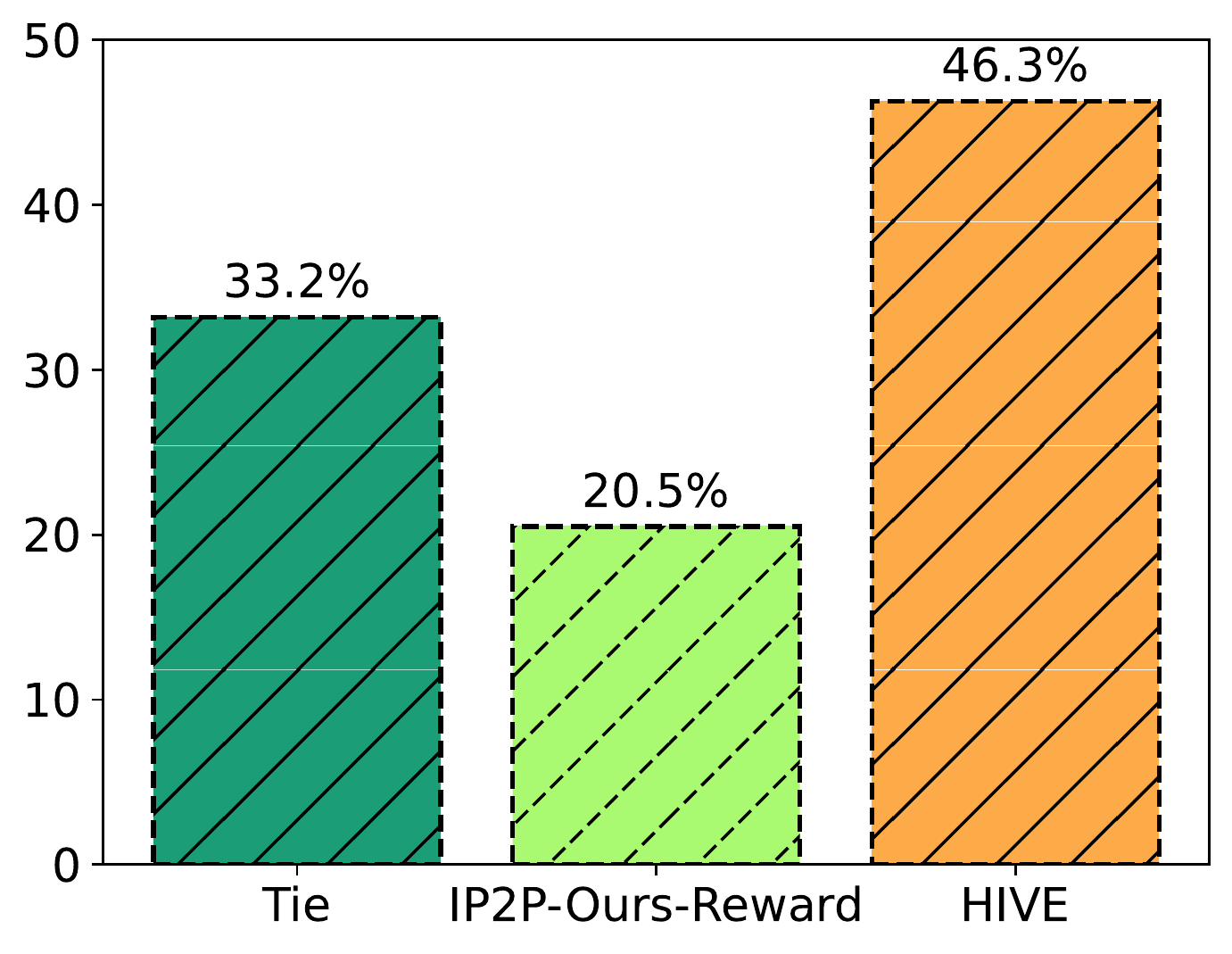}   \\
         \footnotesize{HIVE SD v1.5 and v2.1} & \hspace{-2em}  \footnotesize{HIVE-Reward vs. HIVE} \\
       
    \end{tabular}
    
    \vspace{-1em}
    \caption{User study of pairwise comparison between (a) HIVE with SD v1.5 and v2.1 and (b) HIVE conditioning on reward score and HIVE. The human preferences are very close to each other.}
    \label{fig:userstudy_rewardscore_SDversions}
    \vspace{-1.25em}
\end{figure}

\myparagraph{Other Baselines.} To test the effectiveness of HIVE, we experiment two additional baselines. In Fig. \ref{fig:userstudy_rewardscore_SDversions}(a), we upgrade the backbone of stable diffusion from v1.5 to v2.1. We observe that the upgraded backbone slightly improves the results. 
In Fig. \ref{fig:userstudy_rewardscore_SDversions}(b), we directly use the reward scalar instead of the reward prompt as the condition for training, and the condition on the highest reward scalar for generating the image.
We adopt the user study to compare it (named HIVE-reward) with HIVE.  HIVE obtains 25.8 \% more votes than the baseline model conditioned on the reward score. This is mainly because directly conditioning on the highest reward might cause overfiting.

\begin{figure}[t]
\begin{center}
\includegraphics[ width=0.97\linewidth]{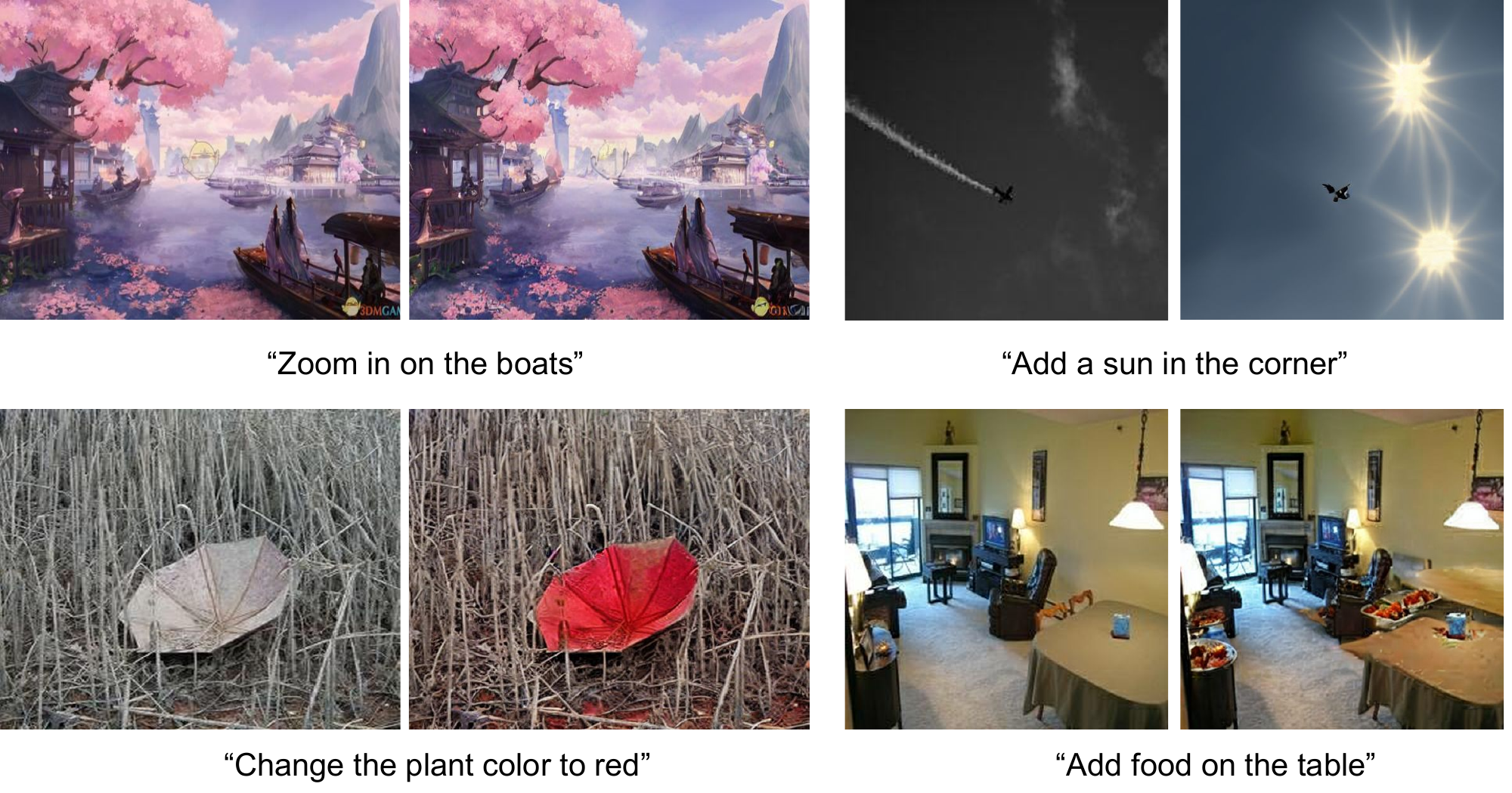}
\end{center}
 \vspace{-2em}
   \caption{Failure examples.}
\label{fig:result_fail}
\vspace{-1.2em}
\end{figure}

\myparagraph{Failure Cases and Limitations.} We summarize representative failure cases in Fig. \ref{fig:result_fail}.
First, some instructions cannot be understood. In the upper left example in Fig. \ref{fig:result_fail}, the prompt ``zoom in'' or similar instructions can rarely be successful. We believe the root cause is current training data generation method fails to generate image pairs with this type of instruction. Second, counting and spatial reasoning are common failure cases (see the upper right example in Fig. \ref{fig:result_fail}). We find that the instruction ``one'', ``two'', or ``on the right'' can lead to many undesired results.
Third, the object understanding sometimes is wrong. In the bottom left example, the red color is changed on the wrong object. This is a common error in HIVE, where instructed edited objects are wrongly recognized. 

We find some other limitations as well. One limitation of HIVE is that it cannot bring benefits to the cases where all outputs by the model without human feedback obtain the same wrong results. In such cases, user preferences cannot always be beneficial to the results. We believe that improving the data as well as the base model is an important step in the future. Another limitation is that compared to Prompt-to-Prompt \cite{hertz2022prompt}, which is used to generate our training data, HIVE sometimes leads to some unstructured change in the image. We think that it is because of the limitation of the current training data. Instructed editing can have more diverse and ambiguous scenarios than traditional image editing problems. Using GPT-3 to finetune prompts to generate the training data is limited by the model and the labeled data.
More ablation studies are in Appendix~\ref{sec:appendix_ablation}.

 \section{Conclusion and Discussion}
In our paper, we introduce a novel framework called HIVE that enables instructional image editing with human feedback. Our framework integrates human feedback, which is quantified as reward values, into the diffusion model fine-tuning process. We design two variants of the approach and both of them improve performance over previous state-of-the-art instructional image editing methods.
Our work demonstrates instructional image editing with human feedback is a variable approach to align image generation with human preference, thus unlocking new opportunities and potential to scale up the model capabilities towards more powerful applications such as conversational image editing. 
While our method demonstrates impressive performance, we have also identified failure scenarios, as discussed in Sec.~\ref{sec:ablation}. In addition, it is possible that our trained model inherits bias and suffers from harmful content from pre-trained foundation models such as Stable Diffusion, GPT3 and BLIP. These limitations would be considered when interpreting our results, and we expect red teaming with human feedback to mitigate some of the risks in future work.

\clearpage

{\small
\bibliographystyle{ieeenat_fullname}
\bibliography{egbib}

\begin{thebibliography}{63}
\providecommand{\natexlab}[1]{#1}
\providecommand{\url}[1]{\texttt{#1}}
\expandafter\ifx\csname urlstyle\endcsname\relax
  \providecommand{\doi}[1]{doi: #1}\else
  \providecommand{\doi}{doi: \begingroup \urlstyle{rm}\Url}\fi

\bibitem[Alayrac et~al.(2022)Alayrac, Donahue, Luc, Miech, Barr, Hasson, Lenc, Mensch, Millican, Reynolds, et~al.]{alayrac2022flamingo}
Jean-Baptiste Alayrac, Jeff Donahue, Pauline Luc, Antoine Miech, Iain Barr, Yana Hasson, Karel Lenc, Arthur Mensch, Katie Millican, Malcolm Reynolds, et~al.
\newblock Flamingo: a visual language model for few-shot learning.
\newblock \emph{arXiv preprint arXiv:2204.14198}, 2022.

\bibitem[Avrahami et~al.(2022{\natexlab{a}})Avrahami, Fried, and Lischinski]{avrahami2022blendedlatent}
Omri Avrahami, Ohad Fried, and Dani Lischinski.
\newblock Blended latent diffusion.
\newblock \emph{arXiv preprint arXiv:2206.02779}, 2022{\natexlab{a}}.

\bibitem[Avrahami et~al.(2022{\natexlab{b}})Avrahami, Lischinski, and Fried]{avrahami2022blended}
Omri Avrahami, Dani Lischinski, and Ohad Fried.
\newblock Blended diffusion for text-driven editing of natural images.
\newblock In \emph{Proceedings of the IEEE/CVF Conference on Computer Vision and Pattern Recognition}, pages 18208--18218, 2022{\natexlab{b}}.

\bibitem[Bai et~al.(2022)Bai, Jones, Ndousse, Askell, et~al.]{Bai2022training}
Yuntao Bai, Andy Jones, Kamal Ndousse, Amanda Askell, et~al.
\newblock Training a helpful and harmless assistant with reinforcement learning from human feedback.
\newblock \emph{arXiv preprint arXiv:2204.05862}, 2022.

\bibitem[Bar-Tal et~al.(2022)Bar-Tal, Ofri-Amar, Fridman, Kasten, and Dekel]{bar2022text2live}
Omer Bar-Tal, Dolev Ofri-Amar, Rafail Fridman, Yoni Kasten, and Tali Dekel.
\newblock Text2live: Text-driven layered image and video editing.
\newblock In \emph{European Conference on Computer Vision}, pages 707--723. Springer, 2022.

\bibitem[Bradley and Terry(1952)]{bradley1952rank}
Ralph~Allan Bradley and Milton~E Terry.
\newblock Rank analysis of incomplete block designs: I. the method of paired comparisons.
\newblock \emph{Biometrika}, 39\penalty0 (3/4):\penalty0 324--345, 1952.

\bibitem[Brooks et~al.(2022)Brooks, Holynski, and Efros]{brooks2022instructpix2pix}
Tim Brooks, Aleksander Holynski, and Alexei~A Efros.
\newblock Instructpix2pix: Learning to follow image editing instructions.
\newblock \emph{arXiv preprint arXiv:2211.09800}, 2022.

\bibitem[Brown et~al.(2020)Brown, Mann, Ryder, Subbiah, Kaplan, Dhariwal, Neelakantan, Shyam, Sastry, Askell, Agarwal, Herbert-Voss, Krueger, Henighan, Child, Ramesh, Ziegler, Wu, Winter, Hesse, Chen, Sigler, Litwin, Gray, Chess, Clark, Berner, McCandlish, Radford, Sutskever, and Amodei]{Brown2020GPT3}
Tom Brown, Benjamin Mann, Nick Ryder, Melanie Subbiah, Jared Kaplan, Prafulla Dhariwal, Arvind Neelakantan, Pranav Shyam, Girish Sastry, Amanda Askell, Sandhini Agarwal, Ariel Herbert-Voss, Gretchen Krueger, Tom Henighan, Rewon Child, Aditya Ramesh, Daniel Ziegler, Jeffrey Wu, Clemens Winter, Christopher Hesse, Mark Chen, Eric Sigler, Mateusz Litwin, Scott Gray, Benjamin Chess, Jack Clark, Christopher Berner, Sam McCandlish, Alec Radford, Ilyva Sutskever, and Dario Amodei.
\newblock Language models are few-shot learners.
\newblock \emph{arXiv preprint arXiv:2005.14165}, 2020.

\bibitem[Chen et~al.(2021)Chen, Lu, Rajeswaran, Lee, Grover, Laskin, Abbeel, Srinivas, and Mordatch]{chen2021decision}
Lili Chen, Kevin Lu, Aravind Rajeswaran, Kimin Lee, Aditya Grover, Misha Laskin, Pieter Abbeel, Aravind Srinivas, and Igor Mordatch.
\newblock Decision transformer: Reinforcement learning via sequence modeling.
\newblock \emph{Advances in neural information processing systems}, 34:\penalty0 15084--15097, 2021.

\bibitem[Christiano et~al.(2017)Christiano, Leike, Brown, Martic, Legg, and Amodei]{christiano2017deepreinforcement}
Paul Christiano, Jan Leike, Tom Brown, Miljan Martic, Shane Legg, and Dario Amodei.
\newblock Deep reinforcement learning from human preferences.
\newblock \emph{NeurIPS}, 2017.

\bibitem[Devlin et~al.(2018)Devlin, Chang, Lee, and Toutanova]{devlin2018bert}
Jacob Devlin, Ming-Wei Chang, Kenton Lee, and Kristina Toutanova.
\newblock Bert: Pre-training of deep bidirectional transformers for language understanding.
\newblock \emph{arXiv preprint arXiv:1810.04805}, 2018.

\bibitem[Dhariwal and Nichol(2021)]{dhariwal2021diffusion}
Prafulla Dhariwal and Alexander Nichol.
\newblock Diffusion models beat gans on image synthesis.
\newblock \emph{NeurIPS}, 34:\penalty0 8780--8794, 2021.

\bibitem[Dosovitskiy et~al.(2020)Dosovitskiy, Beyer, Kolesnikov, Weissenborn, Zhai, Unterthiner, Dehghani, Minderer, Heigold, Gelly, Uszkoreit, and Houlsby]{dosovitskiy2020vit}
Alexey Dosovitskiy, Lucas Beyer, Alexander Kolesnikov, Dirk Weissenborn, Xiaohua Zhai, Thomas Unterthiner, Mostafa Dehghani, Matthias Minderer, Georg Heigold, Sylvain Gelly, Jakob Uszkoreit, and Neil Houlsby.
\newblock An image is worth 16x16 words: Transformers for image recognition at scale.
\newblock \emph{arXiv preprint arXiv:2010.11929}, 2020.

\bibitem[Gal et~al.(2022)Gal, Patashnik, Maron, Bermano, Chechik, and Cohen-Or]{Gal2022stylegannana}
Rinon Gal, Or Patashnik, Haggai Maron, Amit Bermano, Gal Chechik, and Daniel Cohen-Or.
\newblock Stylegan-nada: Clip-guided domain adaptation of image generators.
\newblock \emph{ACM Transactions on Graphics}, 41\penalty0 (4):\penalty0 1--13, 2022.

\bibitem[Goodfellow et~al.(2014)Goodfellow, Pouget-Abadie, Mirza, Xu, Warde-Farley, Ozair, Courville, and Bengio]{Goodfellow2014GAN}
Ian Goodfellow, Jean Pouget-Abadie, Mehdi Mirza, Bing Xu, David Warde-Farley, Sherjil Ozair, Aaron Courville, and Yoshua Bengio.
\newblock Generative adversarial nets.
\newblock \emph{NeurIPS}, 2014.

\bibitem[Hertz et~al.(2022)Hertz, Mokady, Tenenbaum, Aberman, Pritch, and Cohen-Or]{hertz2022prompt}
Amir Hertz, Ron Mokady, Jay Tenenbaum, Kfir Aberman, Yael Pritch, and Daniel Cohen-Or.
\newblock Prompt-to-prompt image editing with cross attention control.
\newblock \emph{arXiv preprint arXiv:2208.01626}, 2022.

\bibitem[Ho et~al.(2020)Ho, Jain, and Abbeel]{ho2020denoising}
Jonathan Ho, Ajay Jain, and Pieter Abbeel.
\newblock Denoising diffusion probabilistic models.
\newblock \emph{NeurIPS}, 33:\penalty0 6840--6851, 2020.

\bibitem[Ibarz et~al.(2018)Ibarz, Leike, Pohlen, Irving, Legg, and Amodei]{Ibarz2018reward}
Borja Ibarz, Jan Leike, Tobias Pohlen, Geoffrey Irving, Shane Legg, and Dario Amodei.
\newblock Reward learning from human preferences and demonstrations in atari.
\newblock \emph{NeurIPS}, 2018.

\bibitem[Isola et~al.(2017)Isola, Zhu, Zhou, and Efros]{isola2017image}
Phillip Isola, Jun-Yan Zhu, Tinghui Zhou, and Alexei~A Efros.
\newblock Image-to-image translation with conditional adversarial networks.
\newblock In \emph{CVPR}, 2017.

\bibitem[Janner et~al.(2021)Janner, Li, and Levine]{janner2021sequence}
Michael Janner, Qiyang Li, and Sergey Levine.
\newblock Offline reinforcement learning as one big sequence modeling problem.
\newblock In \emph{Advances in Neural Information Processing Systems}, 2021.

\bibitem[Karras et~al.(2022)Karras, Aittala, Aila, and Laine]{karras2206elucidating}
Tero Karras, Miika Aittala, Timo Aila, and Samuli Laine.
\newblock Elucidating the design space of diffusion-based generative models, 2022.
\newblock \emph{URL https://arxiv. org/abs/2206.00364}, 2022.

\bibitem[Kim et~al.(2022)Kim, Kwon, and Ye]{Kim_2022_diffusionclip}
Gwanghyun Kim, Taesung Kwon, and Jong~Chul Ye.
\newblock Diffusionclip: Text-guided diffusion models for robust image manipulation.
\newblock In \emph{Proceedings of the IEEE/CVF Conference on Computer Vision and Pattern Recognition (CVPR)}, pages 2426--2435, 2022.

\bibitem[Kingma and Welling(2013)]{kingma2013autoencoding}
Diederik Kingma and Max Welling.
\newblock Auto-encoding variational bayes.
\newblock \emph{arXiv preprint arXiv:1312.6114}, 2013.

\bibitem[Kumar et~al.(2020)Kumar, Zhou, Tucker, and Levine]{kumar2020conservative}
Aviral Kumar, Aurick Zhou, George Tucker, and Sergey Levine.
\newblock Conservative q-learning for offline reinforcement learning.
\newblock \emph{Advances in Neural Information Processing Systems}, 33:\penalty0 1179--1191, 2020.

\bibitem[Lee et~al.(2023)Lee, Liu, Ryu, Watkins, Du, Boutilier, Abbeel, Ghavamzadeh, and Gu]{Lee2023aligning}
Kimin Lee, Hao Liu, Moonkyung Ryu, Olivia Watkins, Yuqing Du, Craig Boutilier, Pieter Abbeel, Mohammad Ghavamzadeh, and Shixiang~Shane Gu.
\newblock Aligning text-to-image models using human feedback.
\newblock \emph{arXiv preprint arXiv:2302.12192}, 2023.

\bibitem[Levine(2018)]{levine2018reinforcement}
Sergey Levine.
\newblock Reinforcement learning and control as probabilistic inference: Tutorial and review.
\newblock \emph{arXiv preprint arXiv:1805.00909}, 2018.

\bibitem[Levine et~al.(2020)Levine, Kumar, Tucker, and Fu]{levine2020offline}
Sergey Levine, Aviral Kumar, George Tucker, and Justin Fu.
\newblock Offline reinforcement learning: Tutorial, review, and perspectives on open problems.
\newblock \emph{arXiv preprint arXiv:2005.01643}, 2020.

\bibitem[Li et~al.(2022)Li, Li, Xiong, and Hoi]{li2022blip}
Junnan Li, Dongxu Li, Caiming Xiong, and Steven Hoi.
\newblock Blip: Bootstrapping language-image pre-training for unified vision-language understanding and generation.
\newblock \emph{arXiv preprint arXiv:2201.12086}, 2022.

\bibitem[Li et~al.(2023)Li, Liu, Wu, Mu, Yang, Gao, Li, and Lee]{li2023gligen}
Yuheng Li, Haotian Liu, Qingyang Wu, Fangzhou Mu, Jianwei Yang, Jianfeng Gao, Chunyuan Li, and Yong~Jae Lee.
\newblock Gligen: Open-set grounded text-to-image generation.
\newblock \emph{arXiv:2301.07093}, 2023.

\bibitem[Liew et~al.(2022)Liew, Yan, Zhou, and Feng]{liew2022magicmix}
Jun~Hao Liew, Hanshu Yan, Daquan Zhou, and Jiashi Feng.
\newblock Magicmix: Semantic mixing with diffusion models.
\newblock \emph{arXiv preprint arXiv:2210.16056}, 2022.

\bibitem[Lu et~al.(2022)Lu, Zhou, Bao, Chen, Li, and Zhu]{lu2022dpm}
Cheng Lu, Yuhao Zhou, Fan Bao, Jianfei Chen, Chongxuan Li, and Jun Zhu.
\newblock Dpm-solver: A fast ode solver for diffusion probabilistic model sampling in around 10 steps.
\newblock \emph{arXiv preprint arXiv:2206.00927}, 2022.

\bibitem[Meng et~al.(2021)Meng, Song, Song, Wu, Zhu, and Ermon]{meng2021sdedit}
Chenlin Meng, Yang Song, Jiaming Song, Jiajun Wu, Jun-Yan Zhu, and Stefano Ermon.
\newblock Sdedit: Image synthesis and editing with stochastic differential equations.
\newblock \emph{arXiv preprint arXiv:2108.01073}, 2021.

\bibitem[Meng et~al.(2022)Meng, He, Song, Song, Wu, Zhu, and Ermon]{meng2022sdedit}
Chenlin Meng, Yutong He, Yang Song, Jiaming Song, Jiajun Wu, Jun-Yan Zhu, and Stefano Ermon.
\newblock {SDE}dit: Guided image synthesis and editing with stochastic differential equations.
\newblock In \emph{International Conference on Learning Representations}, 2022.

\bibitem[Nichol et~al.(2021)Nichol, Dhariwal, Ramesh, Shyam, Mishkin, McGrew, Sutskever, and Chen]{nichol2021glide}
Alex Nichol, Prafulla Dhariwal, Aditya Ramesh, Pranav Shyam, Pamela Mishkin, Bob McGrew, Ilya Sutskever, and Mark Chen.
\newblock Glide: Towards photorealistic image generation and editing with text-guided diffusion models.
\newblock \emph{arXiv preprint arXiv:2112.10741}, 2021.

\bibitem[OpenAI({\natexlab{a}})]{chatGPT}
OpenAI.
\newblock Chatgpt.
\newblock \url{https://openai.com/blog/chatgpt/}, {\natexlab{a}}.

\bibitem[OpenAI({\natexlab{b}})]{openaiapi}
OpenAI.
\newblock Openaiapi.
\newblock \url{https://platform.openai.com/docs/guides/fine-tuning}, {\natexlab{b}}.

\bibitem[Ouyang et~al.(2022)Ouyang, Wu, Jiang, Almeida, Wainwright, Mishkin, Zhang, Agarwal, Slama, Ray, Schulman, Hilton, Kelton, Miller, Simens, Askell, Welinder, Christiano, Leike, and Lowe]{Ouyang2022instructgpt}
Long Ouyang, Jeff Wu, Xu Jiang, Diogo Almeida, Carroll Wainwright, Pamela Mishkin, Chong Zhang, Sandhini Agarwal, Katarina Slama, Alex Ray, John Schulman, Jacob Hilton, Fraser Kelton, Luke Miller, Maddie Simens, Amanda Askell, Peter Welinder, Paul Christiano, Jan Leike, and Ryan Lowe.
\newblock Training language models to follow instructions with human feedback.
\newblock \emph{arXiv preprint arXiv:2203.02155}, 2022.

\bibitem[Peng et~al.(2019)Peng, Kumar, Zhang, and Levine]{peng2019advantage}
Xue~Bin Peng, Aviral Kumar, Grace Zhang, and Sergey Levine.
\newblock Advantage-weighted regression: Simple and scalable off-policy reinforcement learning.
\newblock \emph{arXiv preprint arXiv:1910.00177}, 2019.

\bibitem[Peters et~al.(2010)Peters, Mulling, and Altun]{peters2010relative}
Jan Peters, Katharina Mulling, and Yasemin Altun.
\newblock Relative entropy policy search.
\newblock In \emph{Proceedings of the AAAI Conference on Artificial Intelligence}, pages 1607--1612, 2010.

\bibitem[Pinto et~al.(2023)Pinto, Kolesnikov, Shi, Beyer, and Zhai]{pinto2023tuning}
Andre~Susano Pinto, Alexander Kolesnikov, Yuge Shi, Lucas Beyer, and Xiaohua Zhai.
\newblock Tuning computer vision models with task rewards.
\newblock \emph{arXiv preprint arXiv:2302.08242}, 2023.

\bibitem[Qin et~al.(2023)Qin, Zhang, Yu, Feng, Yang, Zhou, Wang, Niebles, Xiong, Savarese, et~al.]{qin2023unicontrol}
Can Qin, Shu Zhang, Ning Yu, Yihao Feng, Xinyi Yang, Yingbo Zhou, Huan Wang, Juan~Carlos Niebles, Caiming Xiong, Silvio Savarese, et~al.
\newblock Unicontrol: A unified diffusion model for controllable visual generation in the wild.
\newblock In \emph{NeurIPS}, 2023.

\bibitem[Radford et~al.(2021)Radford, Kim, Hallacy, Ramesh, Goh, Agarwal, Sastry, Askell, Mishkin, Clark, et~al.]{radford2021learning}
Alec Radford, Jong~Wook Kim, Chris Hallacy, Aditya Ramesh, Gabriel Goh, Sandhini Agarwal, Girish Sastry, Amanda Askell, Pamela Mishkin, Jack Clark, et~al.
\newblock Learning transferable visual models from natural language supervision.
\newblock \emph{arXiv preprint arXiv:2103.00020}, 2021.

\bibitem[Ramesh et~al.(2021)Ramesh, Pavlov, Goh, Gray, Voss, Radford, Chen, and Sutskever]{ramesh2021zero}
Aditya Ramesh, Mikhail Pavlov, Gabriel Goh, Scott Gray, Chelsea Voss, Alec Radford, Mark Chen, and Ilya Sutskever.
\newblock Zero-shot text-to-image generation.
\newblock In \emph{ICML}, pages 8821--8831. PMLR, 2021.

\bibitem[Ramesh et~al.(2022)Ramesh, Dhariwal, Nichol, Chu, and Chen]{ramesh2022hierarchical}
Aditya Ramesh, Prafulla Dhariwal, Alex Nichol, Casey Chu, and Mark Chen.
\newblock Hierarchical text-conditional image generation with clip latents.
\newblock \emph{arXiv preprint arXiv:2204.06125}, 2022.

\bibitem[Reed et~al.(2016)Reed, Akata, Yan, Logeswaran, Schiele, and Lee]{reed2016generative}
Scott Reed, Zeynep Akata, Xinchen Yan, Lajanugen Logeswaran, Bernt Schiele, and Honglak Lee.
\newblock Generative adversarial text-to-image synthesis.
\newblock In \emph{ICML}, 2016.

\bibitem[Rombach et~al.(2022)Rombach, Blattmann, Lorenz, Esser, and Ommer]{rombach2022high}
Robin Rombach, Andreas Blattmann, Dominik Lorenz, Patrick Esser, and Bj{\"o}rn Ommer.
\newblock High-resolution image synthesis with latent diffusion models.
\newblock In \emph{CVPR}, pages 10684--10695, 2022.

\bibitem[Saharia et~al.(2022)Saharia, Chan, Saxena, Li, Whang, Denton, Ghasemipour, Ayan, Mahdavi, Lopes, et~al.]{saharia2022photorealistic}
Chitwan Saharia, William Chan, Saurabh Saxena, Lala Li, Jay Whang, Emily Denton, Seyed Kamyar~Seyed Ghasemipour, Burcu~Karagol Ayan, S~Sara Mahdavi, Rapha~Gontijo Lopes, et~al.
\newblock Photorealistic text-to-image diffusion models with deep language understanding.
\newblock \emph{arXiv preprint arXiv:2205.11487}, 2022.

\bibitem[Scheurer et~al.(2022)Scheurer, Campos, Chan, Chen, Cho, and Perez]{Scheurer2022training}
Jeremy Scheurer, Jon~Ander Campos, Jun~Shern Chan, Angelica Chen, Kyunghyun Cho, and Ethan Perez.
\newblock Training language models with language feedback.
\newblock \emph{arXiv preprint arXiv:2204.14146}, 2022.

\bibitem[Schuhmann et~al.(2021)Schuhmann, Vencu, Beaumont, Kaczmarczyk, Mullis, Katta, Coombes, Jitsev, and Komatsuzaki]{schuhmann2022laion5B}
Christoph Schuhmann, Richard Vencu, Romain Beaumont, Robert Kaczmarczyk, Clayton Mullis, Aarush Katta, Theo Coombes, Jenia Jitsev, and Aran Komatsuzaki.
\newblock Laion-5b: An open large-scale dataset for training next generation image-text models.
\newblock \emph{arXiv preprint arXiv:2111.02114}, 2021.

\bibitem[Schulman et~al.(2017)Schulman, Wolski, Dhariwal, Radford, and Klimov]{schulman2017ppo}
John Schulman, Filip Wolski, Prafulla Dhariwal, Alec Radford, and Oleg Klimov.
\newblock Proximal policy optimization algorithms.
\newblock \emph{arXiv preprint arXiv:1707.06347}, 2017.

\bibitem[Sohl-Dickstein et~al.(2015)Sohl-Dickstein, Weiss, Maheswaranathan, and Ganguli]{sohl15deep}
Jascha Sohl-Dickstein, Eric Weiss, Niru Maheswaranathan, and Surya Ganguli.
\newblock Deep unsupervised learning using nonequilibrium thermodynamics.
\newblock In \emph{Proceedings of International Conference on Machine Learning}, pages 2256--2265, 2015.

\bibitem[Song et~al.(2020)Song, Meng, and Ermon]{song2020denoising}
Jiaming Song, Chenlin Meng, and Stefano Ermon.
\newblock Denoising diffusion implicit models.
\newblock \emph{arXiv:2010.02502}, 2020.

\bibitem[Song and Ermon(2019)]{song2019generative}
Yang Song and Stefano Ermon.
\newblock Generative modeling by estimating gradients of the data distribution.
\newblock \emph{Advances in Neural Information Processing Systems}, 32, 2019.

\bibitem[Stiennon et~al.(2020)Stiennon, Ouyang, Wu, Ziegler, Lowe, Voss, Radford, Amodei, and Christiano]{stiennon2020learning}
Nisan Stiennon, Long Ouyang, Jeff Wu, Daniel Ziegler, Ryan Lowe, Chelsea Voss, Alec Radford, Dario Amodei, and Paul Christiano.
\newblock Learning to summarize from human feedback.
\newblock \emph{NeurIPS}, 2020.

\bibitem[Wallace et~al.(2022)Wallace, Gokul, and Naik]{wallace2022edict}
Bram Wallace, Akash Gokul, and Nikhil Naik.
\newblock Edict: Exact diffusion inversion via coupled transformations.
\newblock \emph{arXiv preprint arXiv:2211.12446}, 2022.

\bibitem[Wu et~al.(2023)Wu, Sun, Zhu, Zhao, and Li]{wu2023better}
Xiaoshi Wu, Keqiang Sun, Feng Zhu, Rui Zhao, and Hongsheng Li.
\newblock Better aligning text-to-image models with human preference.
\newblock \emph{arXiv preprint arXiv:2303.14420}, 2023.

\bibitem[Xiao et~al.(2021)Xiao, Kreis, and Vahdat]{xiao2021tackling}
Zhisheng Xiao, Karsten Kreis, and Arash Vahdat.
\newblock Tackling the generative learning trilemma with denoising diffusion gans.
\newblock \emph{arXiv preprint arXiv:2112.07804}, 2021.

\bibitem[Xu et~al.(2023)Xu, Liu, Wu, Tong, Li, Ding, Tang, and Dong]{xu2023imagereward}
Jiazheng Xu, Xiao Liu, Yuchen Wu, Yuxuan Tong, Qinkai Li, Ming Ding, Jie Tang, and Yuxiao Dong.
\newblock Imagereward: Learning and evaluating human preferences for text-to-image generation.
\newblock \emph{arXiv preprint arXiv:2304.05977}, 2023.

\bibitem[Xu et~al.(2018)Xu, Zhang, Huang, Zhang, Gan, Huang, and He]{xu2018Attngan}
Tao Xu, Pengchuan Zhang, Qiuyuan Huang, Han Zhang, Zhe Gan, Xiaolei Huang, and Xiaodong He.
\newblock Attngan: Fine-grained text to image generation with attentional generative adversarial networks.
\newblock In \emph{CVPR}, 2018.

\bibitem[Yu et~al.(2022)Yu, Xu, Koh, Luong, Baid, Wang, Vasudevan, Ku, Yang, Ayan, et~al.]{yu2022scaling}
Jiahui Yu, Yuanzhong Xu, Jing~Yu Koh, Thang Luong, Gunjan Baid, Zirui Wang, Vijay Vasudevan, Alexander Ku, Yinfei Yang, Burcu~Karagol Ayan, et~al.
\newblock Scaling autoregressive models for content-rich text-to-image generation.
\newblock \emph{arXiv preprint arXiv:2206.10789}, 2022.

\bibitem[Zhang et~al.(2017)Zhang, Xu, Li, Zhang, Wang, Huang, and Metaxas]{han2017stackgan}
Han Zhang, Tao Xu, Hongsheng Li, Shaoting Zhang, Xiaogang Wang, Xiaolei Huang, and Dimitris Metaxas.
\newblock Stackgan: Text to photo-realistic image synthesis with stacked generative adversarial networks.
\newblock In \emph{{ICCV}}, 2017.

\bibitem[Zhang et~al.(2023)Zhang, Rao, and Agrawala]{zhang2023adding}
Lvmin Zhang, Anyi Rao, and Maneesh Agrawala.
\newblock Adding conditional control to text-to-image diffusion models.
\newblock In \emph{IEEE International Conference on Computer Vision (ICCV)}, 2023.

\bibitem[Zhu et~al.(2017)Zhu, Park, Isola, and Efros]{zhu2017unpaired}
Jun-Yan Zhu, Taesung Park, Phillip Isola, and Alexei~A Efros.
\newblock Unpaired image-to-image translation using cycle-consistent adversarial networks.
\newblock In \emph{ICCV}, 2017.

\end{thebibliography}
}

\clearpage
\appendix
\onecolumn

\begin{center}
\Large
\textbf{Appendix}
\end{center}

\section{Data Collection and User Study}
\label{sec:appendix_dataset}

\begin{figure}[b]
\begin{center}
\includegraphics[ width=0.99\linewidth]{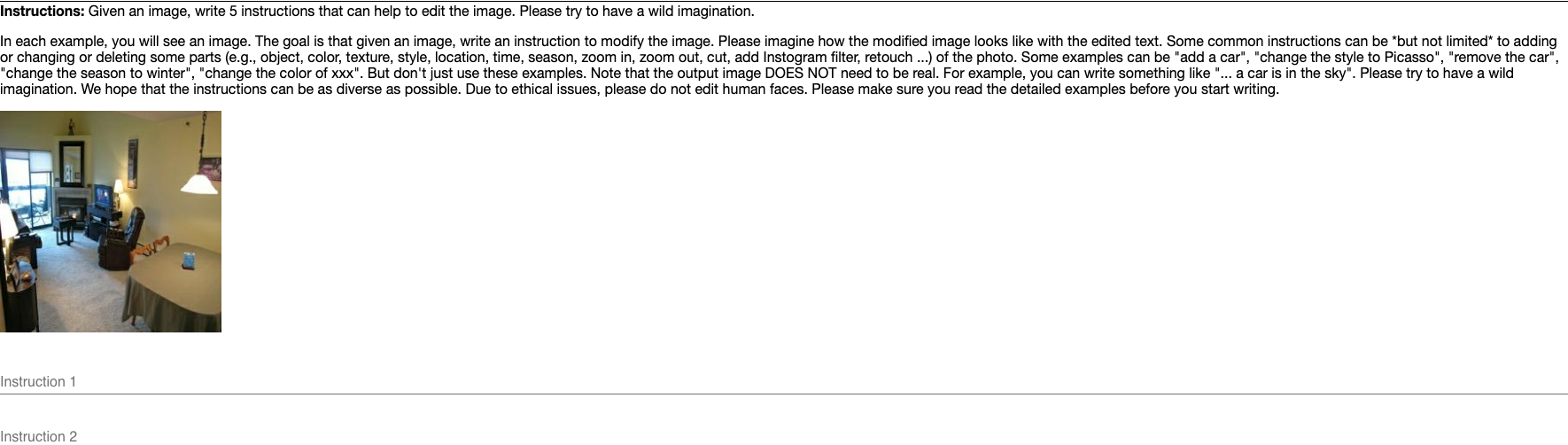}
\end{center}
    \vspace{-2em}
   \caption{Mturk writing editing instructions interface: write five instructions per image.}
\label{fig:mturk_write_instruction}
\end{figure}

\begin{figure}[t]
    \centering
    \begin{tabular}{cc}
         \hspace{-0.5em}\includegraphics[width=.4\textwidth]{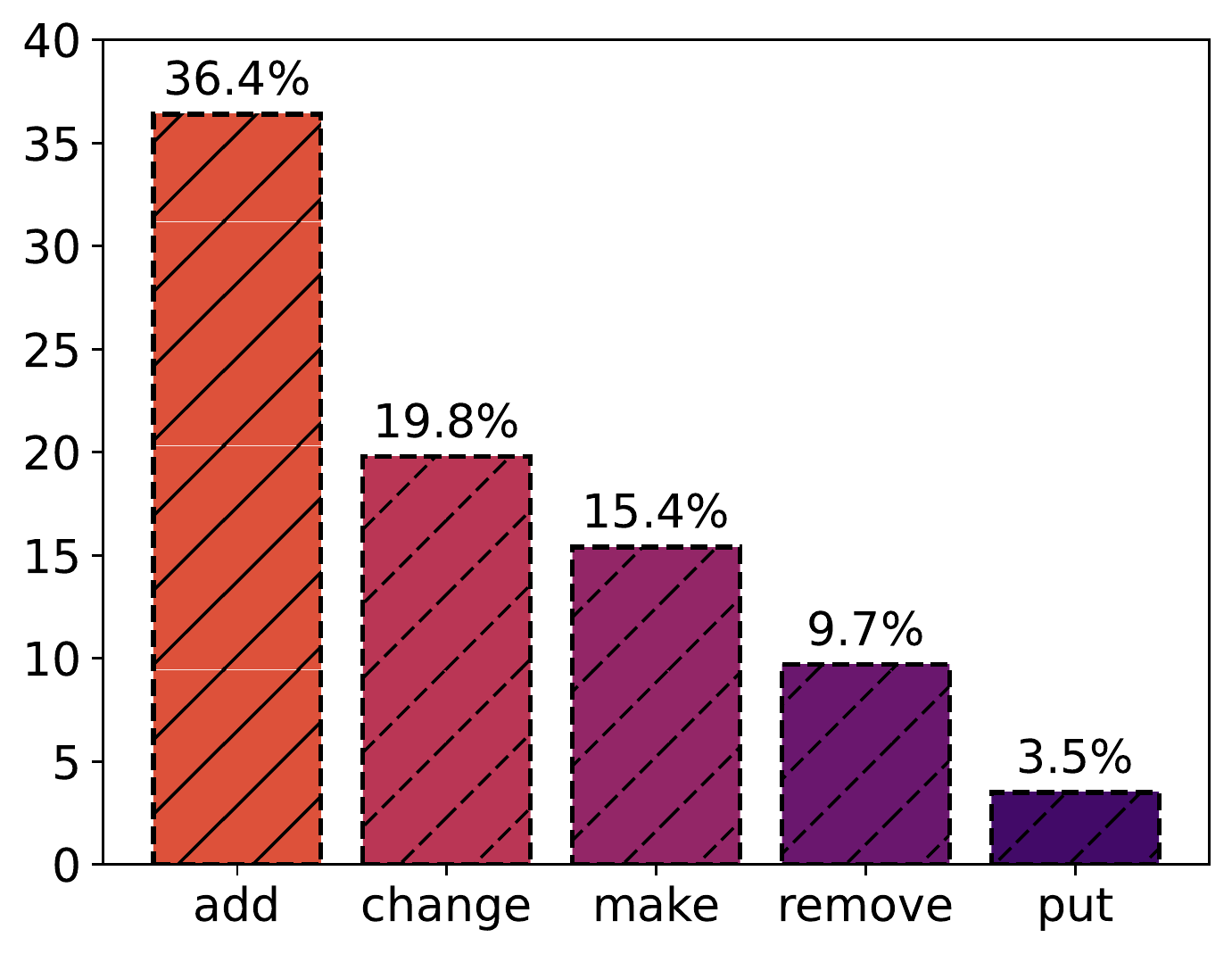}&  
         \hspace{-1em}\includegraphics[width=.4\textwidth]{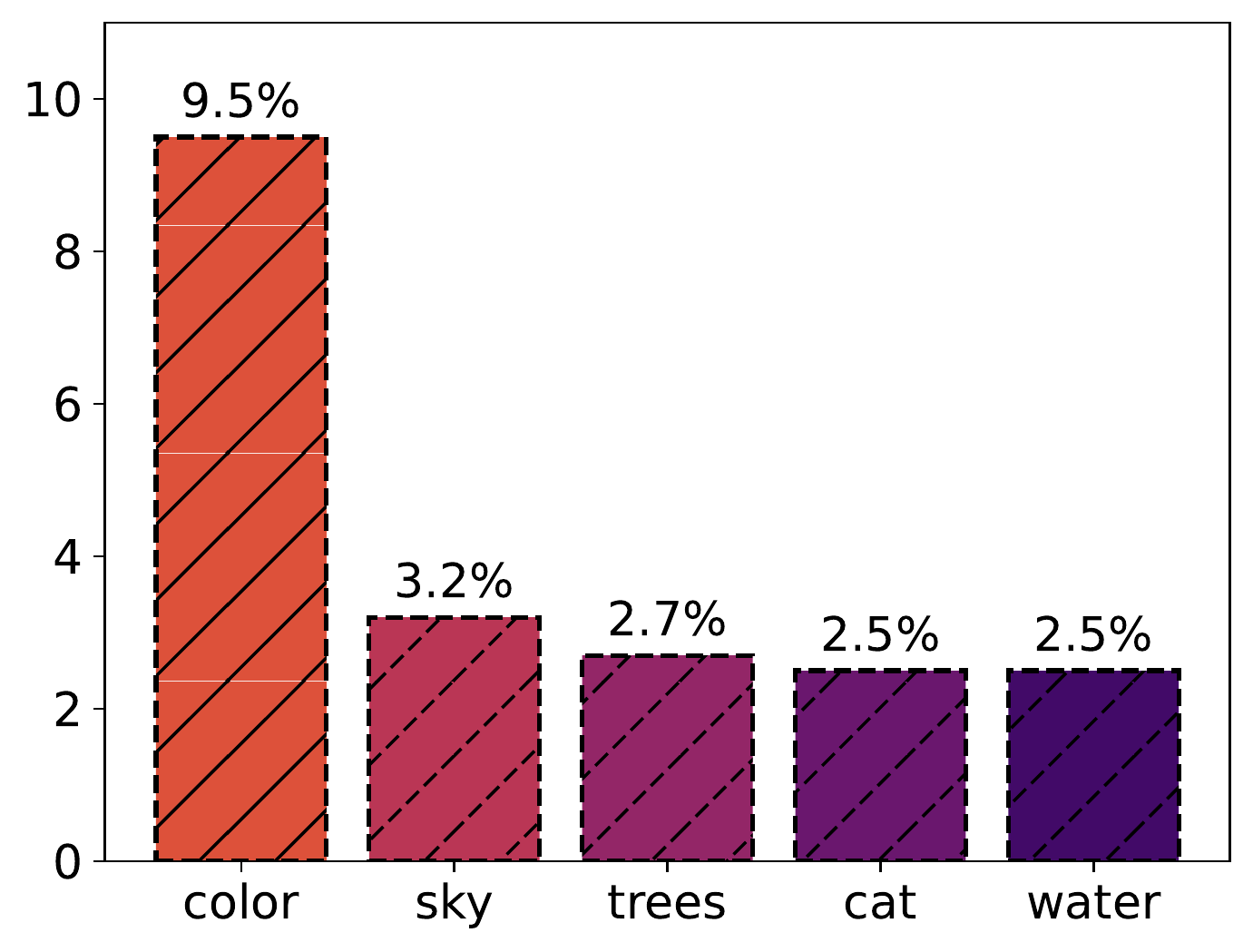}   \\
         \small{Top five verbs} & \small{Top five nouns} \\
       
    \end{tabular}
    
    \caption{Top five verbs and nouns in the evaluation dataset.}
    \label{fig:stat_noun_verb_evaluation}
\end{figure}

In the evaluation steps, we collect real-world images with instructions using Amazon Mechanical Turk (Mturk) \footnote{https://www.mturk.com}. We randomly collect 200 real-world images. Then we ask Mturk annotators to write five instructions for each image, and encourage them to have wild imaginations and diversify the instruction types. We encourage annotators to not be limited to making the image realistic. For example, annotators can write ``add a horse in the sky''. A screenshot of the interface is illustrated in Fig.~\ref{fig:mturk_write_instruction}. We analyze the top five verbs and nouns in the evaluation dataset. It is shown in Fig.~\ref{fig:stat_noun_verb_evaluation}(a) that the verbs ``add'', ``change'', ``make'', ``remove'' and ``put'' make up around 85\% of all verbs, which means that the editing instruction verbs have a long-tail distribution. In contrast, the distribution of nouns in Fig.~\ref{fig:stat_noun_verb_evaluation}(b) is close to uniform, where the top five nouns represent only around 20\% of all nouns.

\begin{figure}[b]
\begin{center}
\includegraphics[ width=0.99\linewidth]{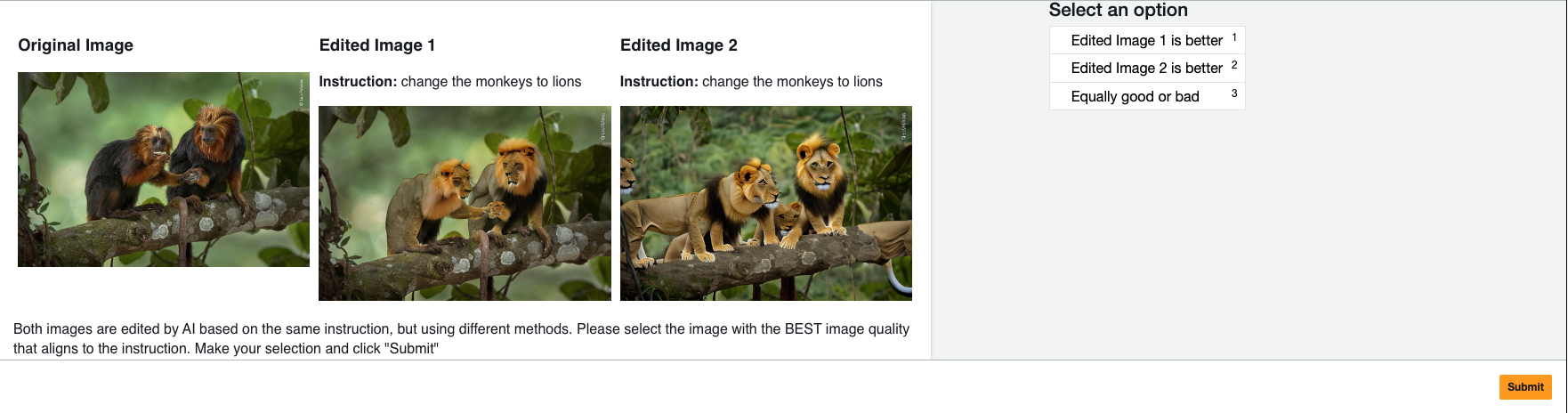}
\end{center}
    \vspace{-2em}
   \caption{Mturk labeling interface: select the better edited image.}
\label{fig:mturk_select_image}
\end{figure}

In user studies, we use Mturk to ask annotators to evaluate edited images. A screenshot of the interface is shown in Fig.~\ref{fig:mturk_select_image}. The annotators are provided with the original image, two edited images, and the editing instruction. They are asked to select the better edited image. The third option indicates that the edited images are equally good or equally bad. We ask three annotators to label one data sample, and use the majority votes to determine the results. We shuffle the edited images to avoid choosing the left image over the right and vice versa.

\section{Implementation Details}
\subsection{Instructional Supervised Training}
\label{sec:appendix_instructional_supervised_training}

We use pre-trained stable diffusion models as the initial checkpoint to start instructional supervised training. We train HIVE on 40GB NVIDIA A100 GPUs for 500 epochs. We use the learning rate of $10^{-4}$ and the image size of 256. In the inference, we use 512 as the default image resolution. 

\subsection{Human Feedback Rewards Learning}
\label{sec:appendix_rm}
\begin{figure}[b]
\begin{center}
\includegraphics[ width=0.99\linewidth]{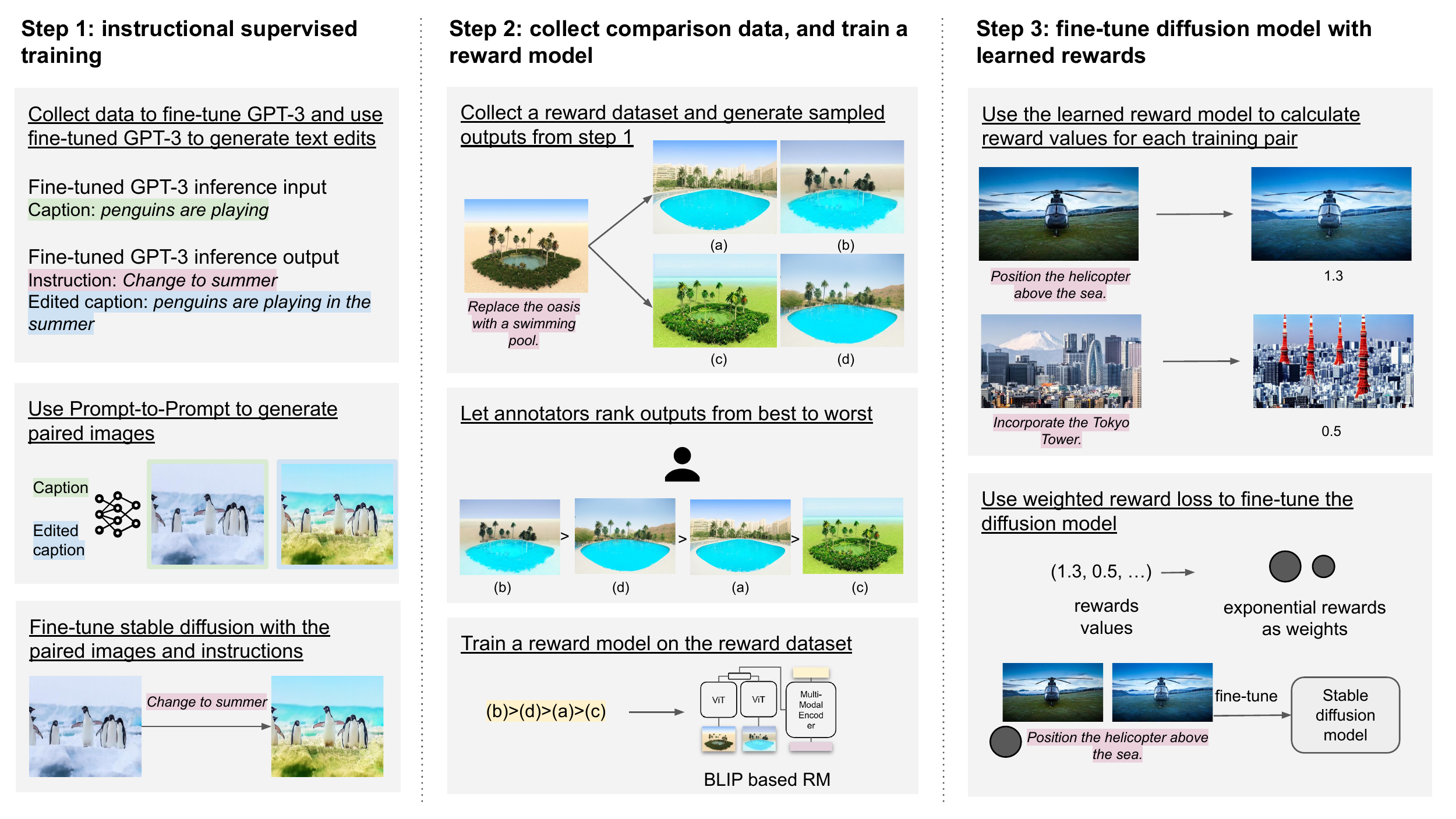}
\end{center}
    \vspace{-2em}
   \caption{Overall architecture of HIVE. Different from Fig. \ref{fig:method}, in the third step, we use weighted reward loss instead of condition reward loss to fine-tune the diffusion model.}
\label{fig:method_weighted}
\end{figure}

As shown in Fig.~\ref{fig:reward_model_architecture}, the reward model takes in an input image $c_I$, a text instruction $c_E$, and an edited image $\tilde{x}$ and outputs a scalar value. Inspired by the recent work on the vision-language model, especially BLIP 
 \cite{li2022blip}, we employ a visual transformer \cite{dosovitskiy2020vit} as our image encoder and an image-grounded text encoder as the multimodal encoder for images and text. Finally, we set a linear layer on top of the image-grounded text encoder to map the multimodal embedding to a scalar value.

(1) \textbf{Visual transformer}. We encode both the input image $c_I$ and edited image $\tilde{x}$ with the same visual transformer. Then we obtain the joint image embedding by concatenating the two image embeddings $vit(c_I)$, $vit(\tilde{x})$.

(2) \textbf{Image-grounded text encoder}. The image-grounded text encoder is a multimodal encoder that inserts one additional cross-attention layer between the self-attention layer and the feed-forward network for each transformer block of BERT \cite{devlin2018bert}. The additional cross-attention layer incorporates visual information into the text model. The output embedding of the image-grounded text encoder is used
as the multimodal representation of the ($c_I$, $c_E$, $\tilde{x}$) triplet.

We gather a dataset comprising 3,634 images for the purpose of ranking. For each image, we generate five variant edited images, and ask an annotator to rank images from best to worst. Additionally, we ask annotators to indicate if any of the following scenarios apply: (1) all edited images are edited but none of them follow the instruction; (2) all edited images are visually the same as the original image; (3) all images are edited beyond the scope of instruction; (4) edited images have harmful content containing sex, violence, porn, etc; and (5) all edited images look similar to each other. We compare training reward models by filtering some/all of these options.

We note that a considerable portion of the collected data falls under at least one of the aforementioned categories, indicating that even for humans, ranking these images is challenging. As a result, we only use the data that did not include any non-rankable options in the reward model training. From a pool of 1,412 images, we select 1,285 for the training set, while the remaining images were used for the validation set. The reward model is trained on a dataset of comparisons between multiple model outputs on the same input. Each comparison sample contains an input image, an instruction, five edited versions of the image, and the corresponding rankings. We divide the dataset into training and validation sets based on the distribution of the corresponding instructions.

We apply the method in Sec. \ref{sec:rewardslearning} on the reward data to develop a reward model. We initialize the reward model from the pre-trained BLIP, which was trained on paired images and captions using three objectives: image-text contrastive learning, image-text matching, and masked language modeling. Although there is a domain gap between BLIP's pre-training data and our reward data, where the captions in BLIP's data describe a single image, and the instructions in our data refer to the difference between image pairs. We hypothesized that leveraging the learned alignment between text and image in BLIP could enhance the reward model's ability to comprehend the relationship between the instruction and the image pairs.

The reward model is trained using 4 A100 GPUs for 10 epochs, employing a learning rate of $10^{-4}$ and weight decay of 0.05. The image encoder's and multimodal encoder's last layer outputs are utilized as image and multimodal representations, respectively. The encoders' final layer is the only  fine-tuned component.

We use the trained reward model to generate a reward score on our training data. We perform two experiments. The first experiment takes the exponential rewards as weights and fine-tunes the diffusion model with \textbf{weighted reward loss} as described in Sec. \ref{sec:hfdf}. See Fig. \ref{fig:method_weighted} for the visualization of the method. The second experiment transforms the rewards to text prompts and fine-tunes the diffusion model with the \textbf{condition reward loss} as described in Sec. \ref{sec:hfdf}. The method is introduced in Fig.~\ref{fig:method}. We compare those two experiment settings, and results can be found in Sec.~\ref{sec:appendix_rewards_exp}.

\section{Reward Maximization for Diffusion-Based Generative Models}
\subsection{Discussion on On-Policy based Reward Maximization for Diffusion Models}
\label{sec:appendix_sampling_methods}
Directly adapting on-policy RL methods to the current training pipeline might be computationally expensive, but we do not conclude that sampling-based approaches are not doable for diffusion models. 
We consider developing more scalable sampling-based methods as future work. 

We start the sampling methods derivation with the following objective:
\begin{align}
    J(\theta):= \max_{\pi_\theta}\E_{c \sim p_c}\Big[\E_{\tgx \sim \pi_{\theta}(\cdot | c)}\left[\mathcal{R}_{\phi}(\tgx, c)\right] - \eta \mathrm{KL}[p_{\mathcal{D}}(\tgx | c) ||  \pi_{\theta}(\tgx | c)]\Big]\,, \label{eq:reverse_obj}
\end{align}
where $p_c(c)p_{\mathcal{D}}(\tgx | c)$ is the joint distribution of the condition and edited images pair, 
and $\pi_{\theta}$ denotes the policy or the diffusion model we want to optimize.
Note that $p_{\mathcal{D}}(\tgx | c)$ and  $\pi(\tgx | c)$ are swaped compared with the objective in Eq.~\eqref{equ:rl_obj}. 
The second term in Eq.~\eqref{eq:reverse_obj}, is the \textit{KL Minimization} formula for maximum likelihood estimation, equivalent to the loss of diffusion models. 
We represent the policy $\pi_{\theta}$ via the \textit{reverse process} of a conditional diffusion model:
\begin{align*}
    \pi_\theta(\tgx | c):= p_{\theta}(\tgx^{0:T} | ~c) = p_{0}(\tgx^{T})\prod_{t=1}^{T}p_{\theta}(\tgx^{t-1}| \tgx^{t}; c)\,,
\end{align*}
where $p_{0}(\tgx^{T}):=\mathcal{N}(\tgx^{T}, \bf{0}; \bf{{I}}) $, and $p_{\theta}(\tgx^{t-1}| \tgx^{t}; c):= \mathcal{N}(\tgx^{t} | \mu_{\theta}(\tgx_{t}, t), \sigma_t^2) \bf{\mathrm{I}}$ is a Gaussian distribution, whose parameters are defined by score function $\epsilon_\theta$ and stepsize of noise scalings. 
So we can get a edited image sample $\tgx^{0}$ by running a reverse diffusion chain: 
\begin{align*}
    \tgx^{t-1} | \tgx^{t} = \frac{1}{\sqrt{\alpha_t}}\left(\tgx^t - \frac{1 - \alpha_t}{\sqrt{1 - \bar{\alpha}_t}}\epsilon_{\theta}(\tgx^t,c, t)\right) + \sigma_t \pmb{z}_t, ~~\pmb{z}\sim \mathcal{N}(\textbf{0}, \textbf{I}), \text{for}~~ t=T,\ldots,1\,,
\end{align*}
and $\tgx^{T} \sim \mathcal{N}(\textbf{0}, \textbf{I})$. 

As a result, the reverse diffusion process can be viewed as a black box function defined by $\epsilon_\theta$ and noises $\pmb{\epsilon}:= (\pmb{z}_T, \ldots, \pmb{z}_1, \tgx^T)$, which we can view as a \textit{shared parameter network with noises}. And for each layer, we can view the parameter is the score function $\epsilon_\theta$. 
Define the network as 
\begin{align*}
    \tgx^{0} := f(c, \pmb{\epsilon}; \theta)\,,~~ \pmb{\epsilon} \sim p_{\mathrm{noise}}(\cdot), c\sim p_{c}(\cdot)\,,
\end{align*}
where we can rewrite the first term as 
\begin{align*}
    \E_{c \sim \mathcal{D}, \pmb{\epsilon} \sim p_{\mathrm{noise}}(\cdot)}[\mathcal{R}_{\phi}(f(c, \pmb{\epsilon}; \theta), c)]\,,
\end{align*}
and we can optimize the parameter $\theta$ with path gradient if $\mathcal{R}_{\cdot}$ is differentiable with path gradient.
Similarly, suppose we want to optimize the first term via PPO. In that case, the main technical difficulty is to estimate $\nabla_{\theta}\log\pi_{\theta}(\tgx| c)$, which can be estimated with the following derivation:
\begin{align*}
\nabla_{\theta}\log \pi_{\theta}(\tgx| c) = \nabla_{\theta}\log p_{\theta}(\tgx^{0:T} | ~c) = \sum_{t=1}^{T}\nabla_{\theta}\log p_{\theta}(\tgx^{t-1}| \tgx^{t}; c)\,.
\end{align*}
Note that for both the end-to-end path gradient method and PPO we require to sample the reverse chain from $\tgx^{T}$ to $\tgx^{0}$, thus we can estimate $\nabla_{\theta}\log \pi(\tgx | c)$ using the empirical samples $\tgx^{0:T}$.

For the above two methods, to perform one step policy gradient update, we need to run the whole reverse chain to get an edited image sample $\tgx^{0}$ to estimate the parameter gradient for the first term. 
As a result, the computational cost is the number of diffusion steps more extensive than the supervised fine-tuning 
cost. Now we need more than two days to fine-tune the stable diffusion model, so for standard LDM, where the number of steps is 1000, we can not finish the training within an acceptable training time. 
Even if we can use some fast sampling methods such as DDIM or variance preserve (VP) based noise scaling, the diffusion steps are still more than 5 or 10. 
Further, we haven't seen any previous work using such noise scaling to fine-tune stable diffusion. As a result, we think naive sampling methods might have high risk to obtain similar performance, compared with our current offline RL based approaches.

\subsection{Derivation for Eq.~\eqref{equ:optim_q}}
\label{sec:appendix_derivation}
Take a functional view of Eq.~\eqref{equ:optim_q}, and differentiate $J(\rho)$ w.r.t $\rho$, we get 
\begin{align*}
    \frac{\partial J(\rho)}{\partial \rho}= \mathcal{R}_{\phi}(\tgx |c) - \eta \left(\log\rho(\tgx|c)  + 1 - \log p(\tgx | c)\right)\,.
\end{align*}
Setting $\frac{\partial J(\rho)}{\partial \rho} = 0$ gives us
\begin{align*}
    \log \rho(\tgx | c) &= \frac{1}{\eta}\mathcal{R}_{\phi}(\tgx |c) + \log p(\tgx |c)  - 1\,,\\
    \rho(\tgx | c) & \propto p(\tgx | c)\exp\left(\mathcal{R}_{\phi}(\tgx , c) / \eta\right)\,.
\end{align*}
Thus we can get the optimal $\rho^{*}(\tgx | c)$.

\section{Additional Ablation Study}
\label{sec:appendix_ablation}

\subsection{SD v1.5 and v2.1.} 

\begin{figure}[t]
    \centering
    \begin{tabular}{cc}
         \hspace{-0.5em}\includegraphics[width=.4\textwidth]{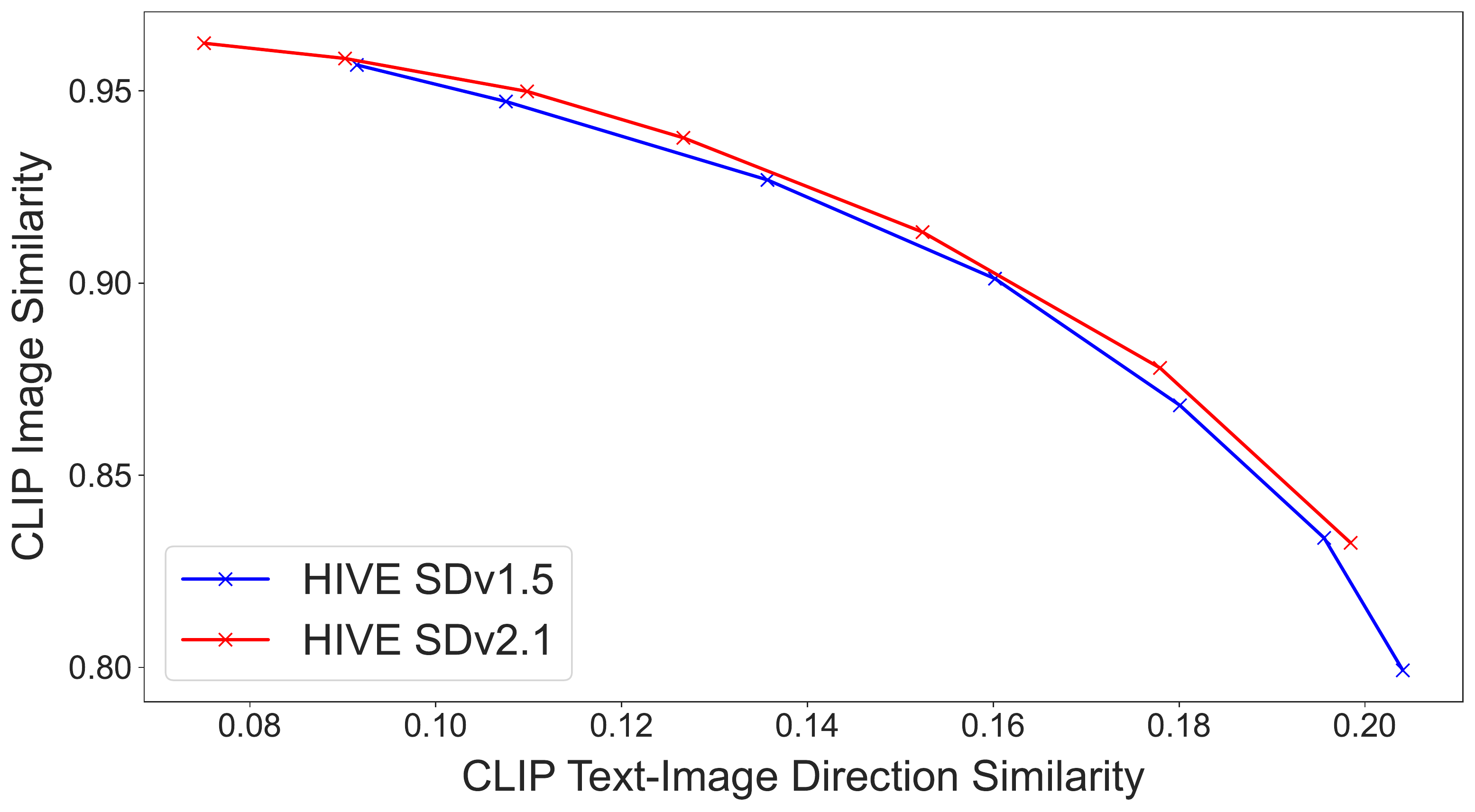}&  
         \hspace{-1em}\includegraphics[width=.4\textwidth]{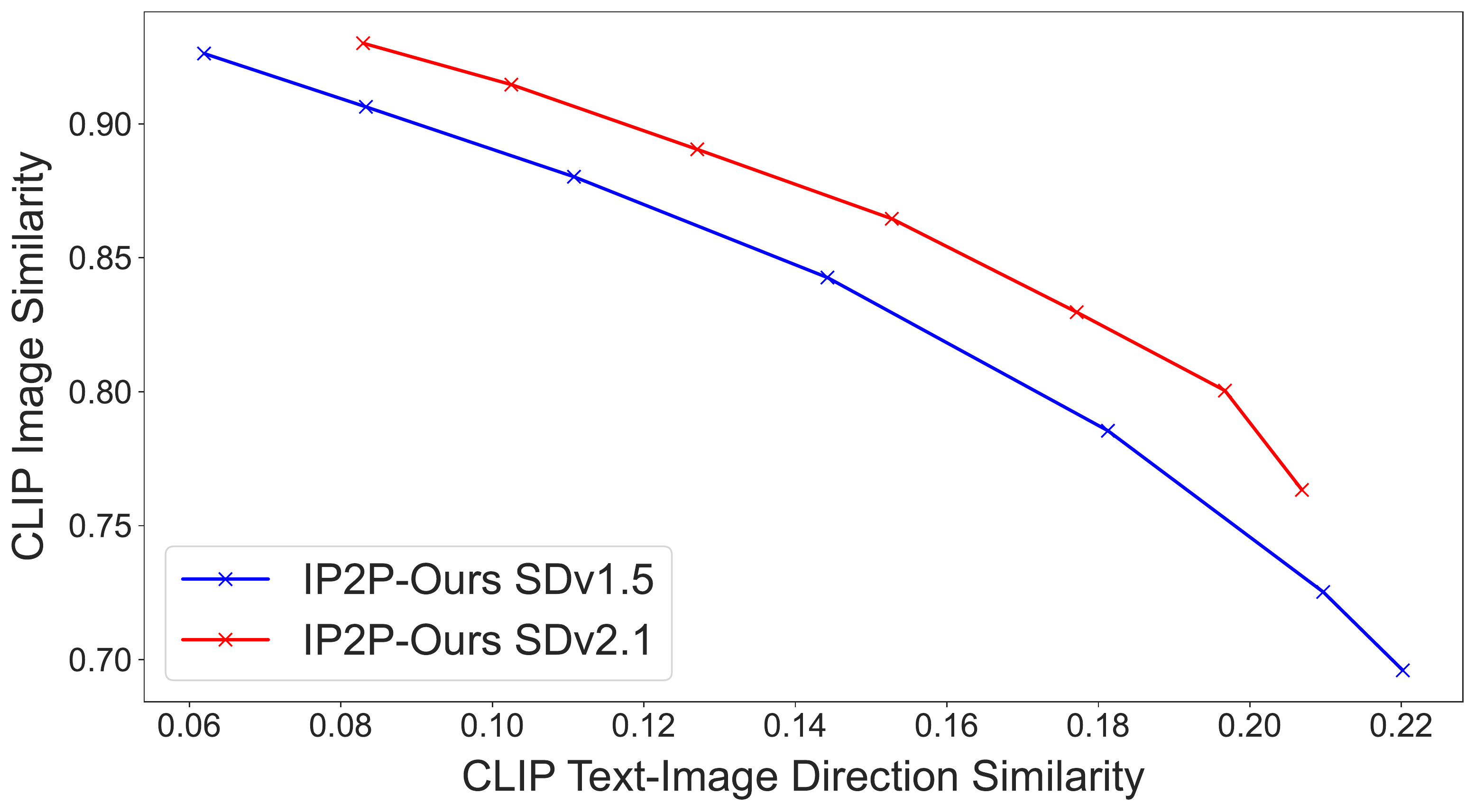}   \\
         \small{IP2P-Ours with SD v1.5 and v2.1} & \small{InstructPix2Pix with SD v1.5 and v2.1} \\
       
    \end{tabular}
    
    \caption{HIVE and IP2P-Ours with SD v1.5 and v2.1.}
    \label{fig:hive_ip2p_sdv1.5_2.1}
\end{figure}

In Sec. \ref{sec:ablation}, we upgrade the backbone of stable diffusion from v1.5 to v2.1, where OpenCLIP text encoder \cite{schuhmann2022laion5B} replaces the CLIP text encoder \cite{radford2021learning}. In this section, we demonstrate the quantitative consistency plot in Fig.~\ref{fig:hive_ip2p_sdv1.5_2.1}(a) on the synthetic evaluation dataset, which shows similar conclusions to the user study in Fig. \ref{fig:userstudy_rewardscore_SDversions}(a). We compare IP2P-Ours v1.5 with v2.1 as well. An interesting observation is that we train IP2P-Ours with SD v2.1 and show in Fig.~\ref{fig:hive_ip2p_sdv1.5_2.1}(b) that its improvement over SD v1.5 is larger than HIVE in Fig.~\ref{fig:hive_ip2p_sdv1.5_2.1}(a).

\subsection{Model Adaptation}

\begin{figure}
    \centering
    \vspace{-1em}
         \hspace{-0.5em}\includegraphics[width=.35\textwidth]{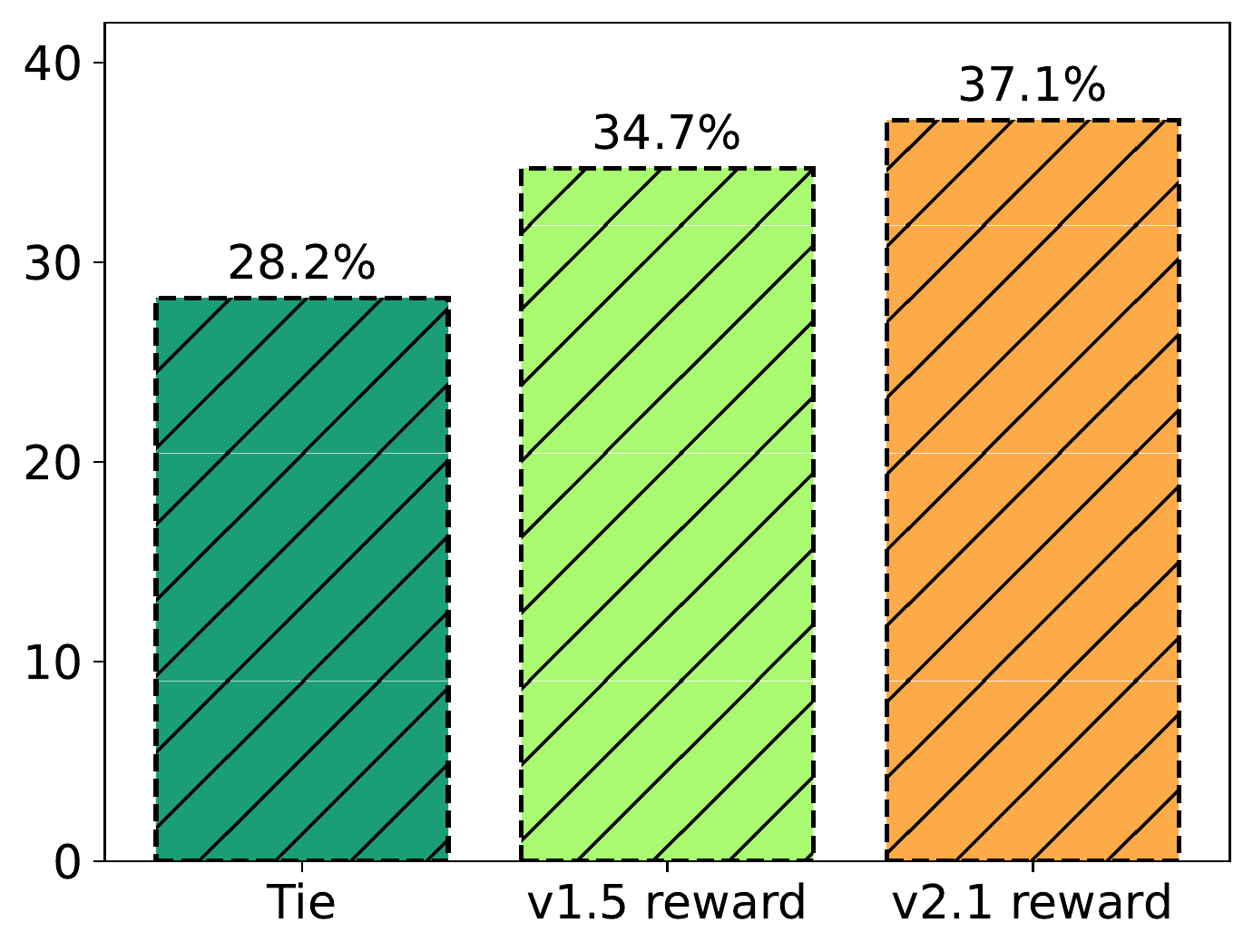}

    \vspace{-1em}
    \caption{SD v1.5 trained vs. SD v2.1 trained reward model}
    \label{fig:user_study_rewards}
\end{figure}

We demonstrate that HIVE is able to adapt the reward model that is trained on a different backbone from the backbone in Step. 3. We use the SD v1.5 generated data to train the reward model, and process the rest steps using SD v2.1. We report user study results in Fig. \ref{fig:user_study_rewards}. It is observed that the users vote similarly between the reward models that are trained on two SD backbones. In other words, the reward model is able to adapt from one backbone to another.

\subsection{Weighted Reward and Conditional Reward Losses}
\label{sec:appendix_rewards_exp}

\begin{figure}
\begin{center}
\includegraphics[ width=0.5\linewidth]{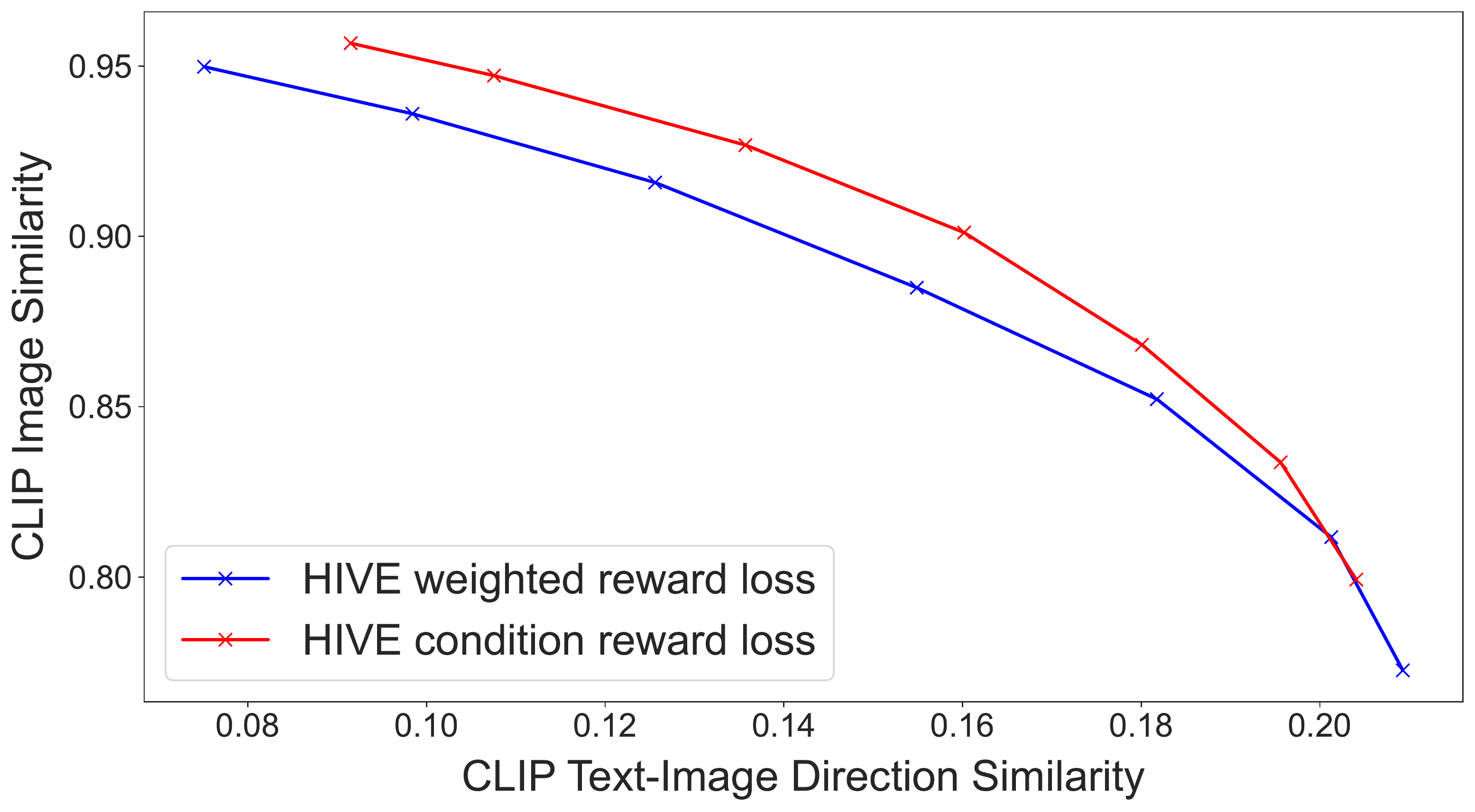}
\end{center}
    \vspace{-1.8em}
   \caption{HIVE with weighted reward loss and conditional reward loss.}
\vspace{-1em}
\label{fig:plot_reward_losses}
\end{figure}

We compare the weighted reward loss and conditional reward loss on the synthetic evaluation dataset. As shown in Fig.~\ref{fig:plot_reward_losses}, the performances of these two losses are close to each other, while the conditional reward loss is slightly better. Therefore we adopt the conditional reward loss in all our experiments.

\subsection{Training with Less Data}

\begin{figure}
\begin{center}
\includegraphics[ width=0.5\linewidth]{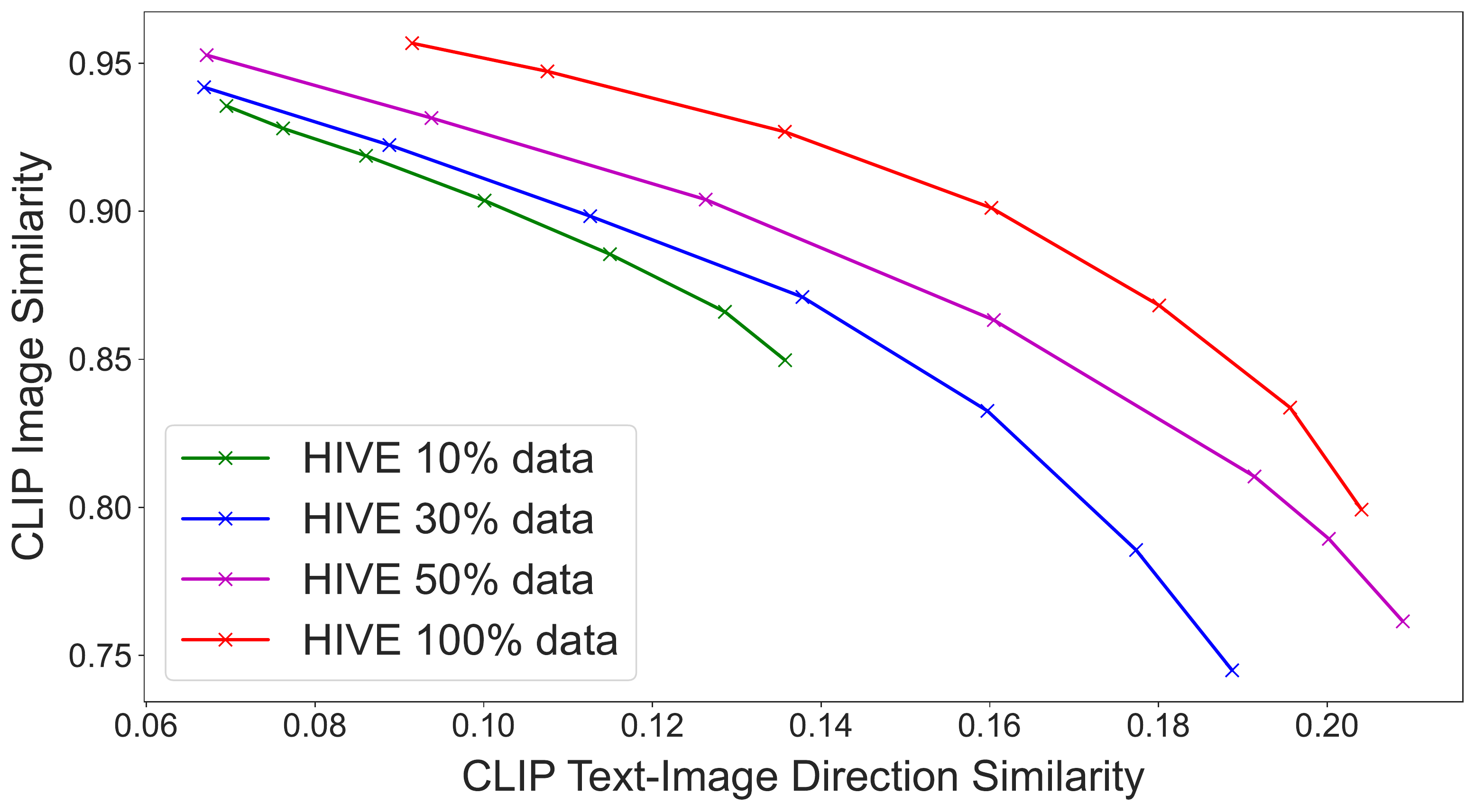}
\end{center}
    \vspace{-1.8em}
   \caption{HIVE with different training data size.}
\vspace{-1em}
\label{fig:plot_v1_different_datasize}
\end{figure}

We analyze the effect of the training data size. We compare HIVE with SD v1.5 at four training dataset size ratios: 100\%, 50\%, 30\% and 10\%. As shown in Fig. \ref{fig:plot_v1_different_datasize}, significantly decreasing the size of the dataset, \eg 10\% data, leads to worse ability to perform large image edits. On the other hand, reasonable decreasing dataset size  can result in a similar yet slightly worse performance \eg 50\% data.

\subsection{Subcategory Analysis}

\begin{figure}[t]
    \centering
    \vspace{-1em}
    \begin{tabular}{ccc}
         \hspace{-0.5em}\includegraphics[width=.16\textwidth]{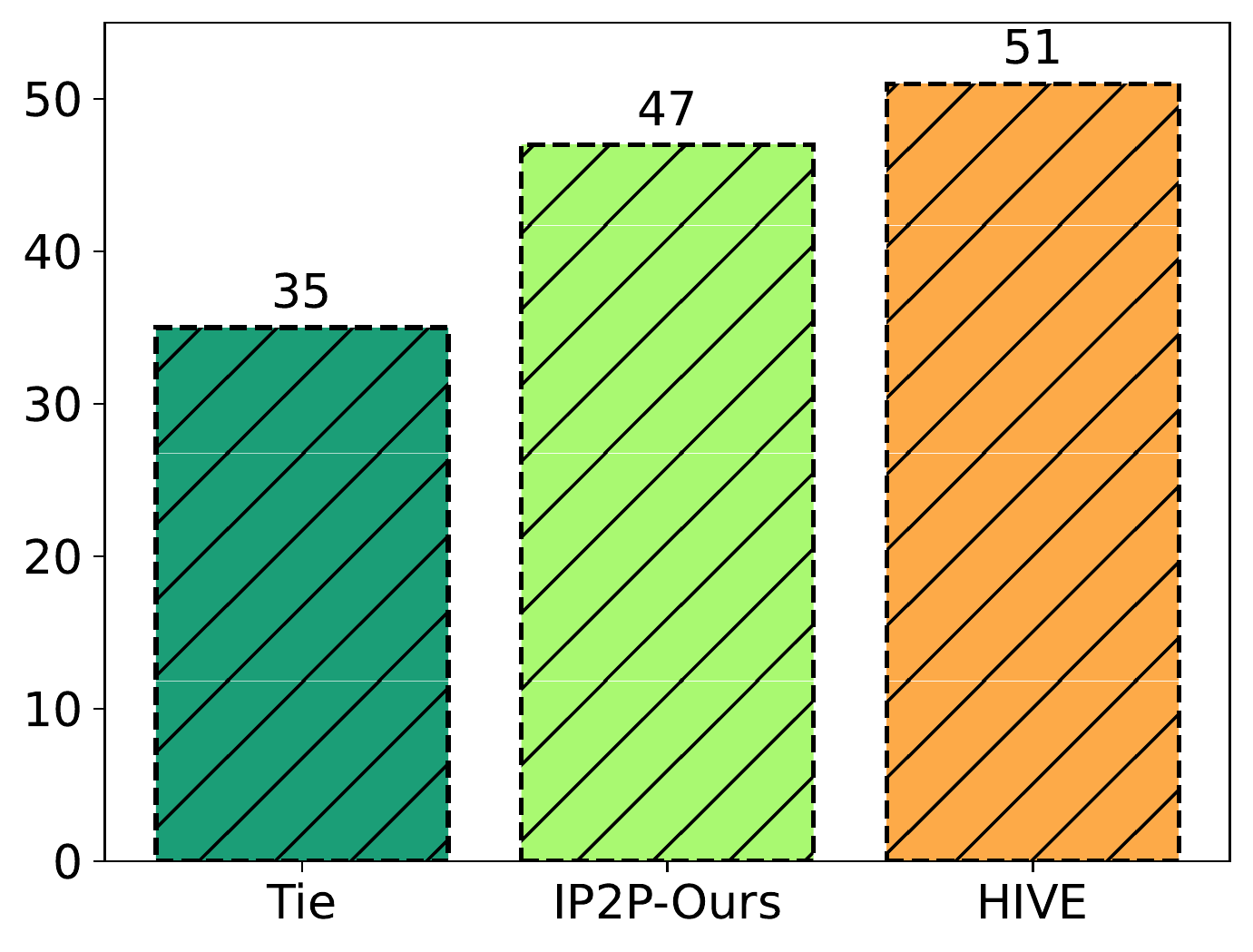}&  
         \hspace{-1em}\includegraphics[width=.16\textwidth]{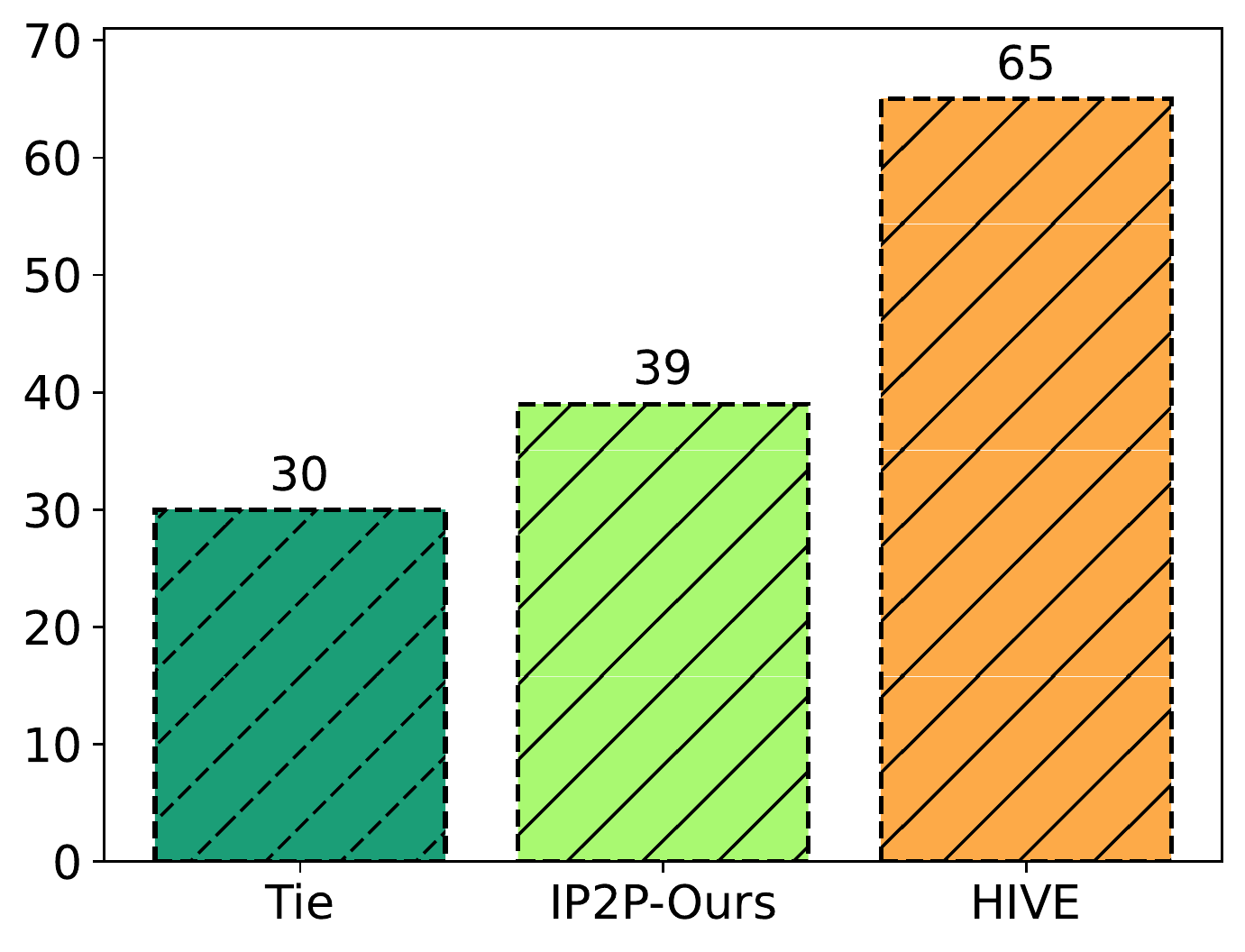} &
         \hspace{-1em}\includegraphics[width=.16\textwidth]{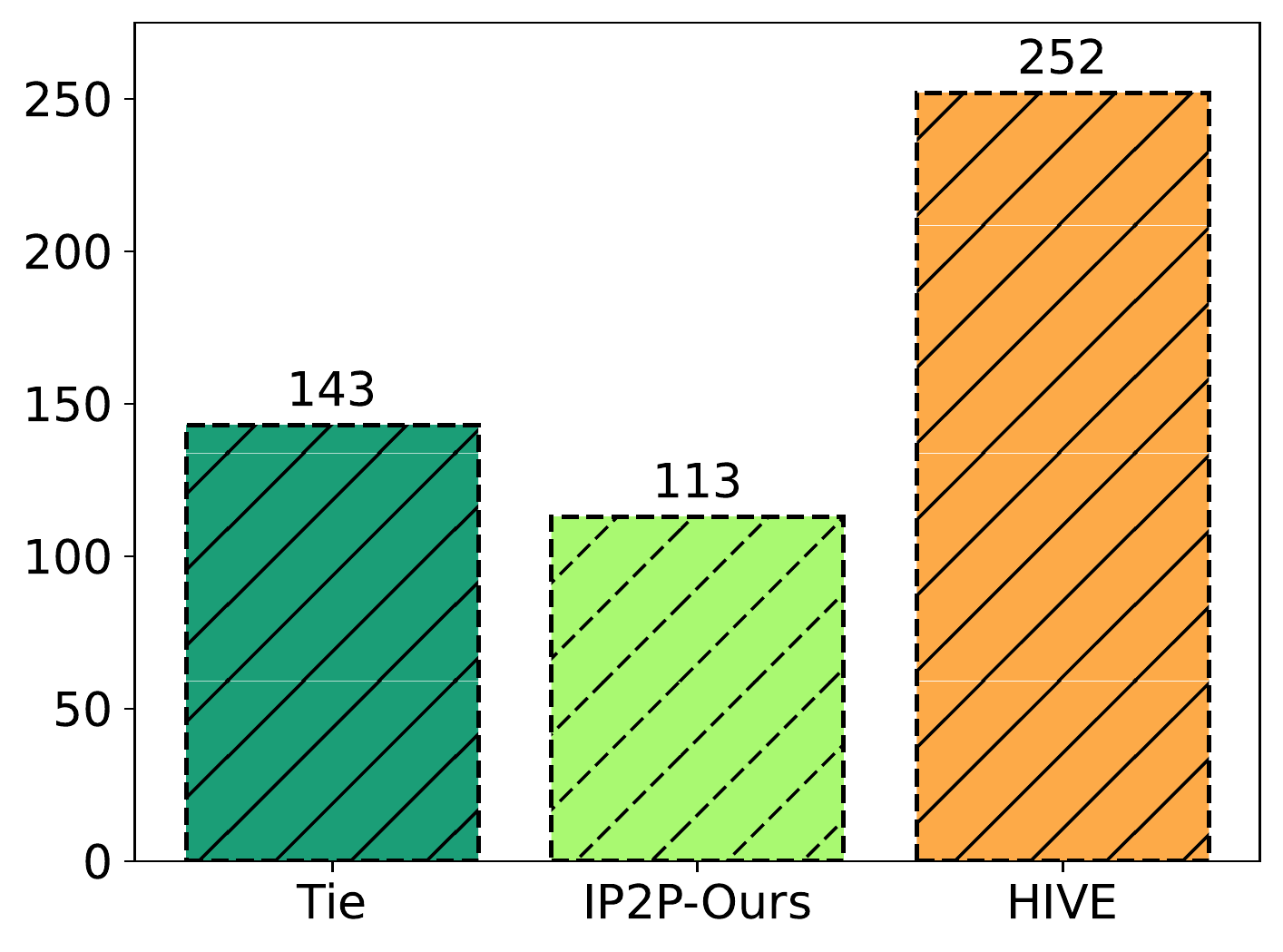}   \\
         \small{Change global style} & \small{Adjust attributes} & \small{Add/remove} \\
       
    \end{tabular}
    \begin{tabular}{ccc}
         \hspace{-0.5em}\includegraphics[width=.16\textwidth]{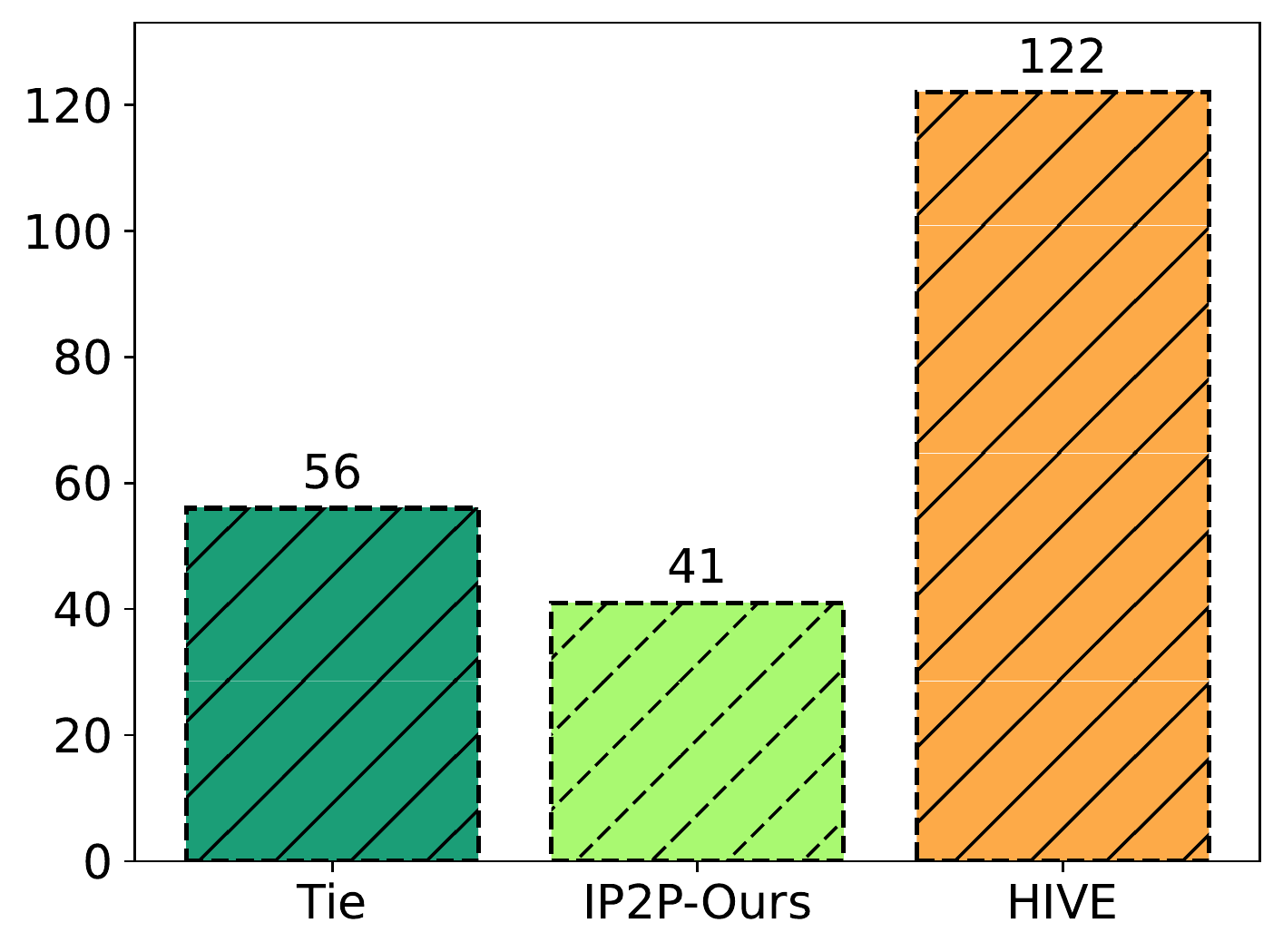}&  
         \hspace{-1em}\includegraphics[width=.16\textwidth]{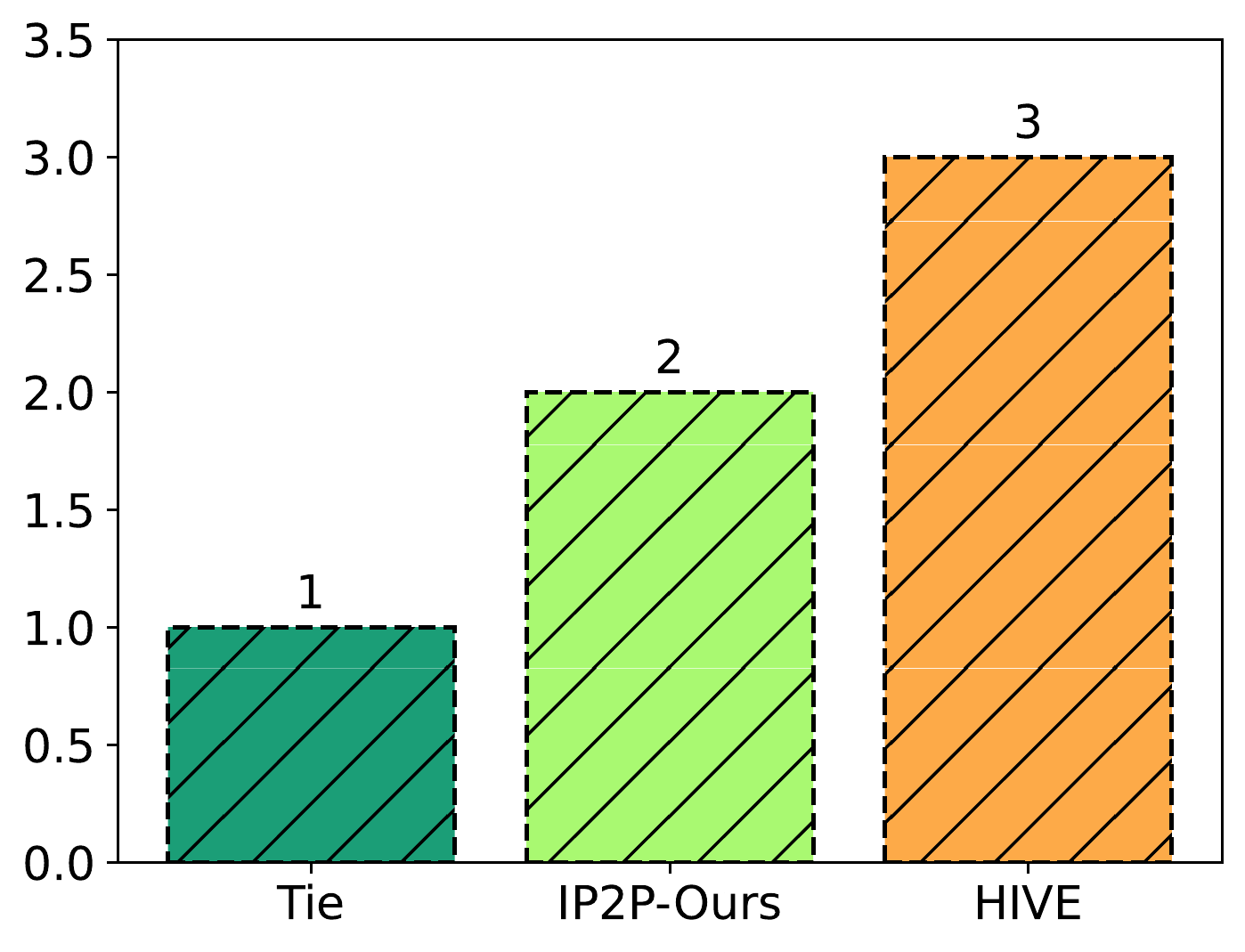}   \\
         \small{Manipulate objects} & \small{Others} &   \\
       
    \end{tabular}
    
    \vspace{-1em}
    \caption{Subcategory analysis between IP2P and HIVE.}
    \label{fig:subcategory}
\end{figure}

We classify the editing into the following sub-categories: changing the global style, adjust attributes for the main object, add/remove objects, manipulate objects, and other challenging cases such as zooming and camera view changes. We use ChatGPT \footnote{https://chat.openai.com/} to determine which sub-category the instruction belongs to. Specifically, the numbers of instructions in each sub-category are as follows: changing global style (133), adjust attributes for the main object (134), add/remove objects (508), manipulate objects (219), and others (6). We analyze user study results for each sub-category. It is shown in Fig. \ref{fig:subcategory} that the most improvement comes from the sub-categories "Add/remove objects" and "Manipulate objects".

\subsection{Additional Visualized Results}

\begin{figure}
\begin{center}
\includegraphics[ width=0.85\linewidth]{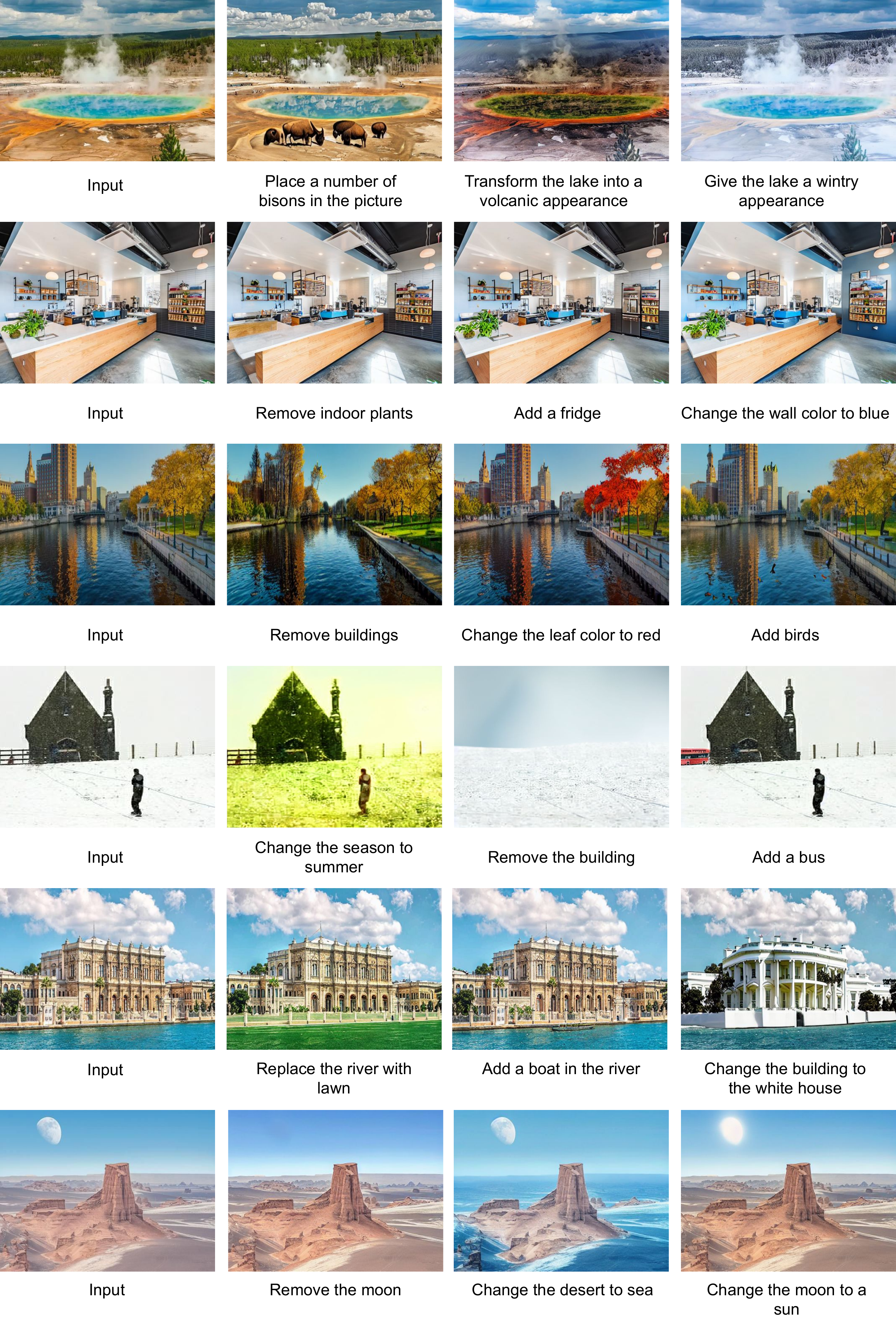}
\end{center}
    \vspace{-2em}
   \caption{Additional editing results.}
\label{fig:res_moreresults_1}
\end{figure}

\begin{figure}
\begin{center}
\includegraphics[ width=0.85\linewidth]{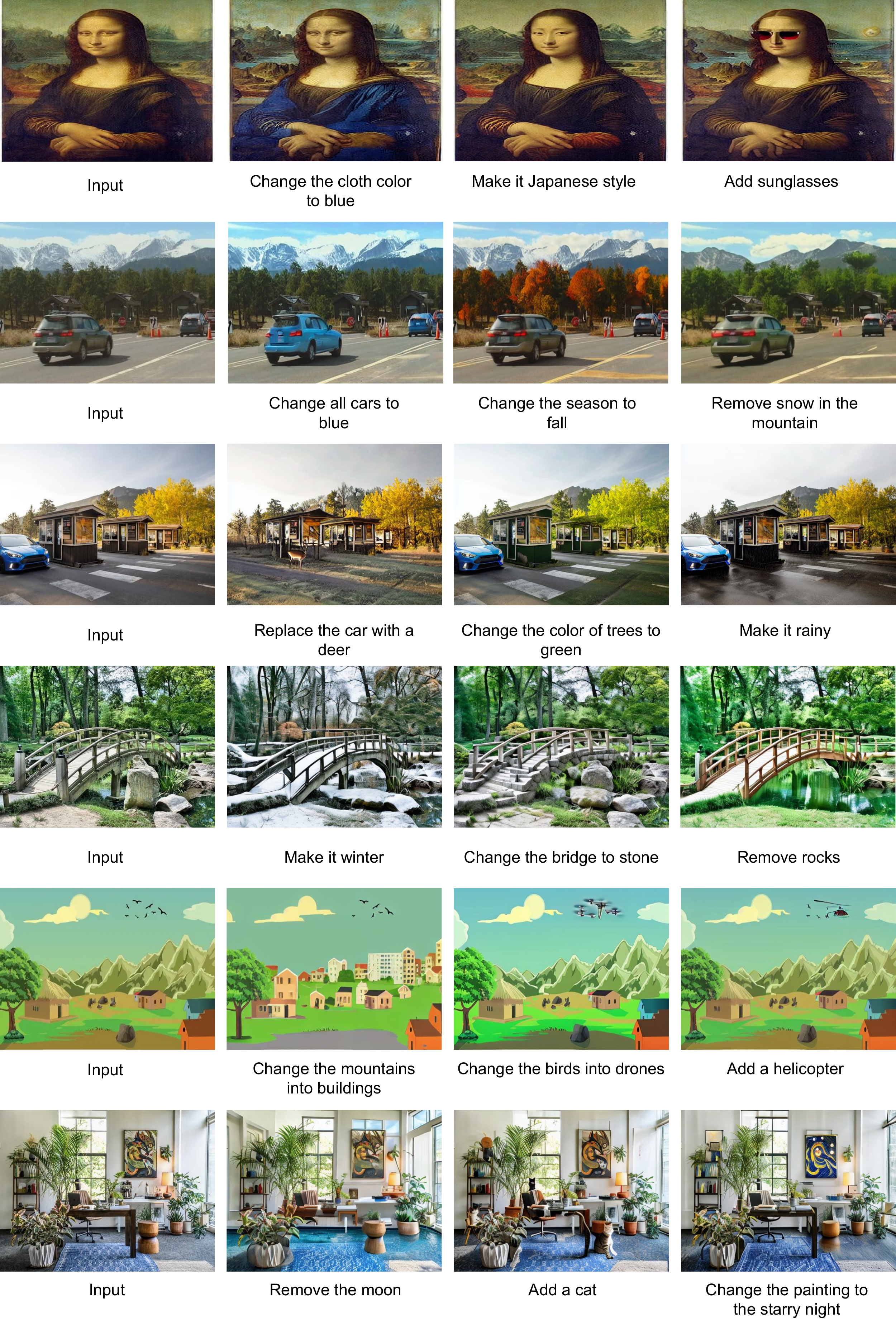}
\end{center}
    \vspace{-2em}
   \caption{Additional editing results.}
\label{fig:res_moreresults_2}
\end{figure}

\begin{figure}
\begin{center}
\includegraphics[ width=0.85\linewidth]{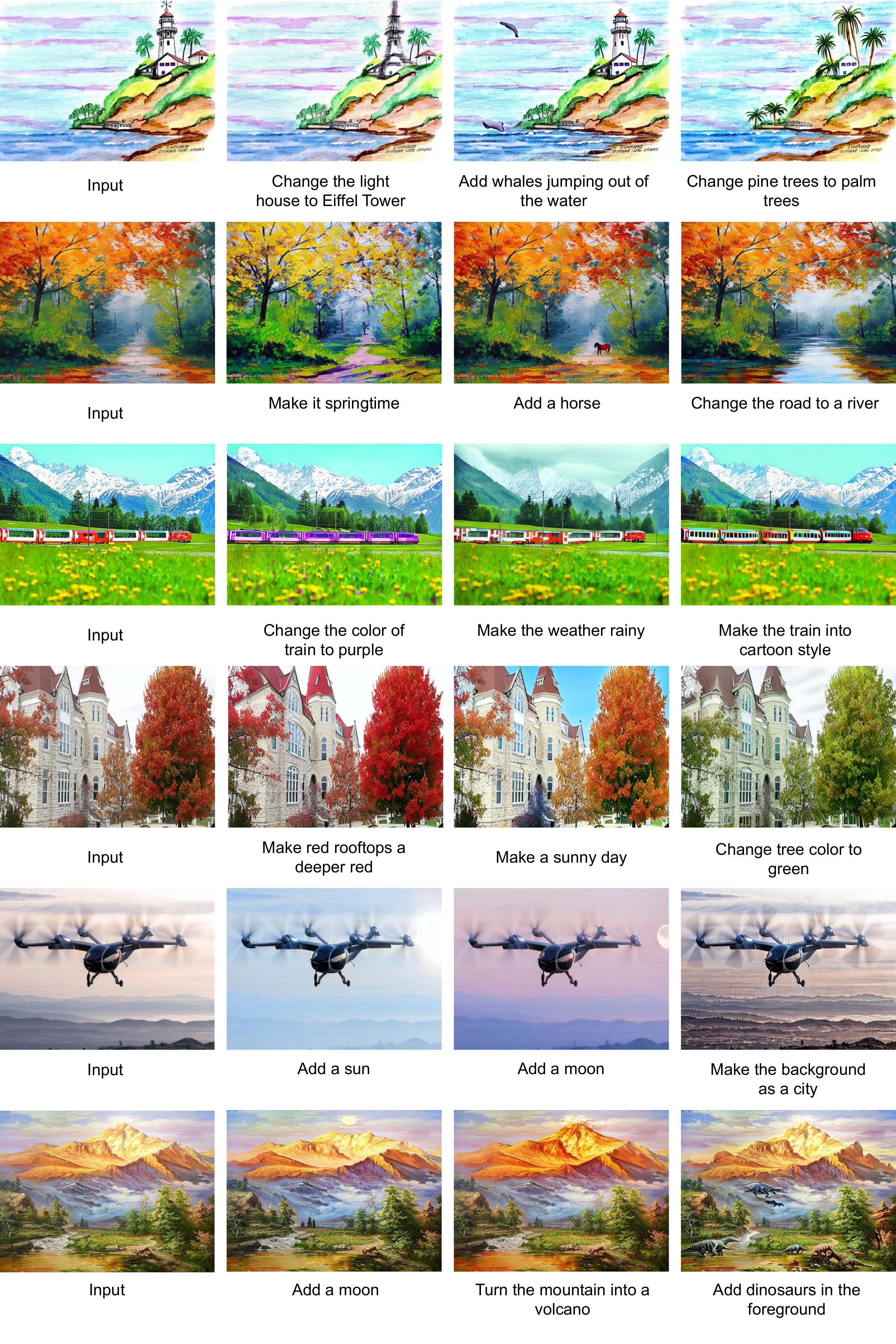}
\end{center}
    \vspace{-2em}
   \caption{Additional editing results.}
\label{fig:res_moreresults_3}
\end{figure}

\begin{figure}
\begin{center}
\includegraphics[ width=0.85\linewidth]{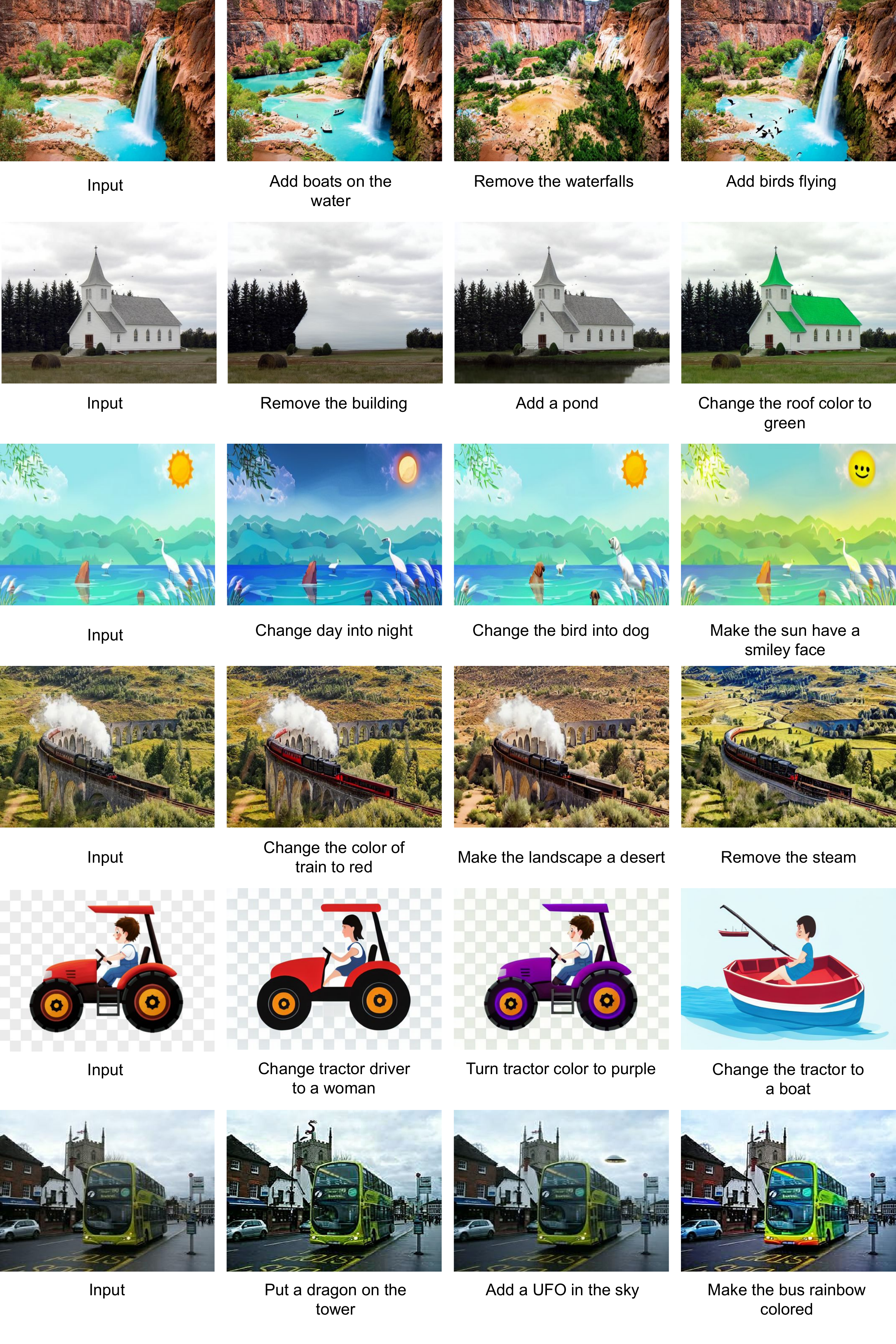}
\end{center}
    \vspace{-2em}
   \caption{Additional editing results.}
\label{fig:res_moreresults_4}
\end{figure}

\begin{figure}
\begin{center}
\includegraphics[ width=0.85\linewidth]{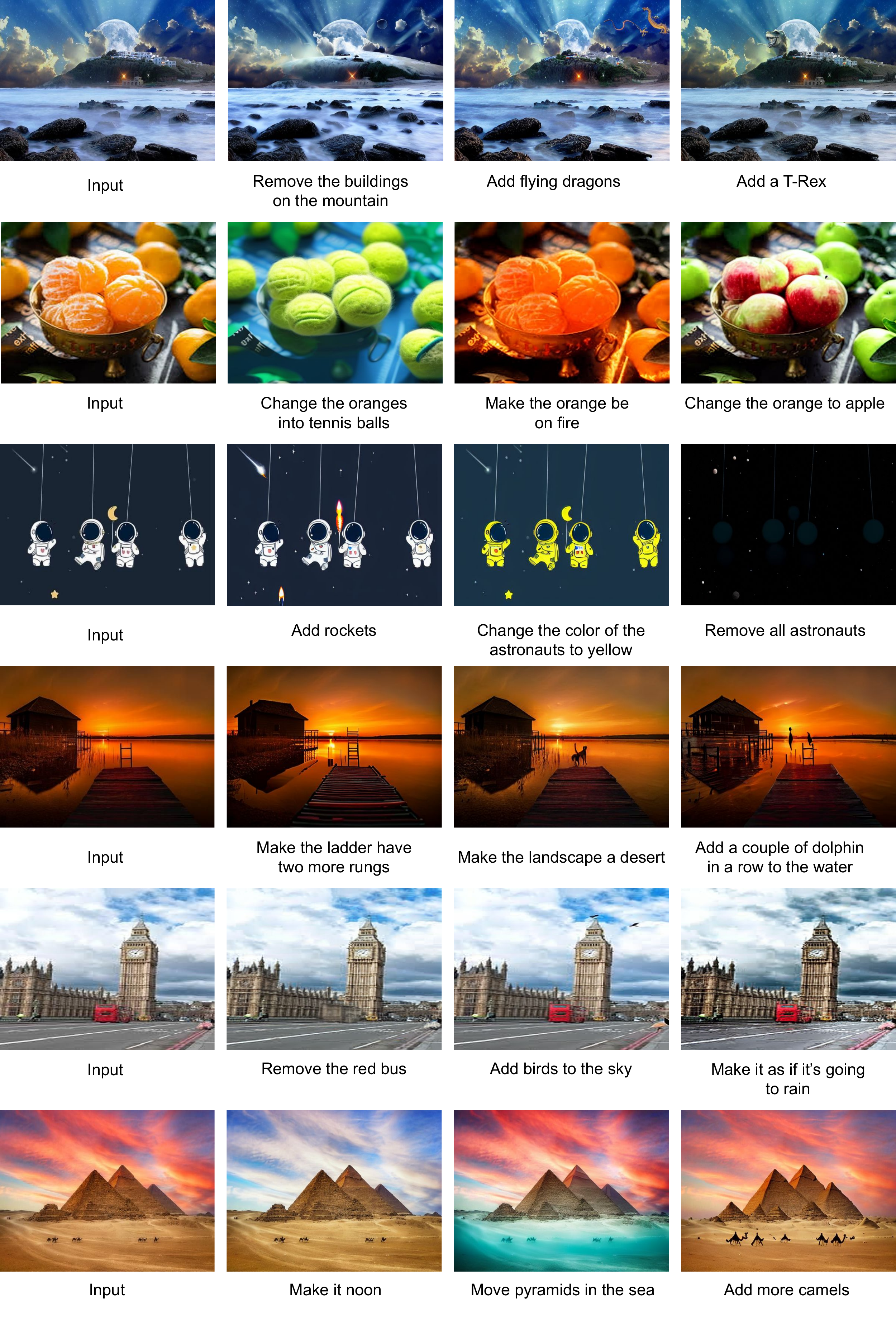}
\end{center}
    \vspace{-2em}
   \caption{Additional editing results.}
\label{fig:res_moreresults_5}
\end{figure}

We illustrate additional visualized results in Fig.~\ref{fig:res_moreresults_1},~\ref{fig:res_moreresults_2},~\ref{fig:res_moreresults_3},~\ref{fig:res_moreresults_4},~\ref{fig:res_moreresults_5}, where each row illustrates three instructional editing examples.

\end{document}